\documentclass[11pt]{article}

\usepackage[margin=1in]{geometry}
\usepackage[T1]{fontenc}
\usepackage{lmodern}
\usepackage{microtype}
\usepackage{amsmath,amssymb,amsthm,physics}
\usepackage{mathtools}
\usepackage{graphicx}
\usepackage{booktabs}
\usepackage{hyperref}
\usepackage[nameinlink]{cleveref}
\usepackage{algorithm}
\usepackage{algpseudocode}
\usepackage{tikz}
\usepackage{enumitem}

\usepackage{subcaption} 
\theoremstyle{plain} 

\newtheorem{hypothesis}{Hypothesis} 
\newtheorem{theorem}{Theorem}
\newtheorem{lemma}[theorem]{Lemma} 
\newtheorem{proposition}[theorem]{Proposition} 
\newtheorem{remark}[theorem]{Remark}

\newtheorem{definition}[theorem]{Definition}

\crefname{hypothesis}{Hypothesis}{Hypotheses}

\usepackage{natbib}
\hypersetup{
  colorlinks=true,
  linkcolor=blue,
  citecolor=blue,
  urlcolor=blue
}
\DeclareMathOperator*{\argmin}{arg\,min}
\title{Large Learning Rates Impair Length Generalization}
\author{Anıl Kamber \\ UC San Diego \\ \texttt{akamber@ucsd.edu} \and Rahul Parhi \\ UC San Diego \\ \texttt{rahul@ucsd.edu}}
\date{}

\begin{document}

\maketitle
\begin{abstract}
   Length generalization is the ability of sequence models to effectively process sequences longer than those seen during training. While the quadratic cost of attention prohibits transformers from processing arbitrarily long sequences, their most promising subquadratic alternatives, such as linear recurrent models, often fail beyond the training horizon. Despite its significance, our general understanding of length generalization remains limited. In this paper, we investigate how the learning rate determines which global minimum gradient descent converges to when these minima differ in their ability to generalize to arbitrarily long sequences. To this end, we study a single-layer linear recurrent neural network whose hidden dimension is large enough to realize any sequence of Markov parameters over a fixed training horizon. Our experiments show that sufficiently large learning rates induce edge-of-stability dynamics, and the resulting global minima exhibit worse length generalization than those reached at smaller learning rates, for which the dynamics remain outside this regime. Among runs that enter the edge-of-stability regime, larger learning rates further degrade length generalization, in contrast to the common view that large learning rates improve generalization. We observe that, among these runs, larger learning rates favor global minima whose recurrent weight matrices have a larger spectral radius restricted to the controllable and observable modes. To understand this phenomenon, we derive a closed-form expression for the maximum Hessian eigenvalue of the training loss at any global minimum.
\end{abstract}
\textbf{Keywords:} recurrent models, dynamical systems, linear dynamical systems, next-token prediction, length generalization, implicit bias, loss-landscape geometry
\begingroup

\renewcommand\thefootnote{}

\footnotetext{Copyright © 2026 by Anıl Kamber and Rahul Parhi}
\endgroup

\section{Introduction}
Given the complexity of the underlying systems, from protein folding \citep{dill2012protein, jumper2021highly} to the mechanisms of human cognition \citep{lake2016building}, pinning down a probabilistic model of the sequences that emerge in nature seems intractable. We therefore rely on the statistical models learned from data. To this end, the next-token prediction task defines a choice of objective for training statistical models that can approximate the Bayes-optimal one-step-ahead predictor\footnote{For a loss function $\ell$, the Bayes-optimal one-step-ahead predictor is a dynamic function $g_k^*$ that satisfies
$g_k^*(Y_1, \ldots, Y_k) \in \argmin_{g}
\mathbb{E}[\ell(Y_{k+1},g(Y_1, \ldots, Y_k)) \mid Y_1, \ldots, Y_k]$ for every $k \in \mathbb{N}$.} of the underlying dynamical system that generates the sequences we train on.

\begin{figure*}[t]
    \centering
    \captionsetup[subfigure]{justification=centering,singlelinecheck=false}

    \begin{subfigure}[t]{0.333\textwidth}
        \centering
        \IfFileExists{figures/mamba_1.png}{%
            \includegraphics[width=\linewidth]{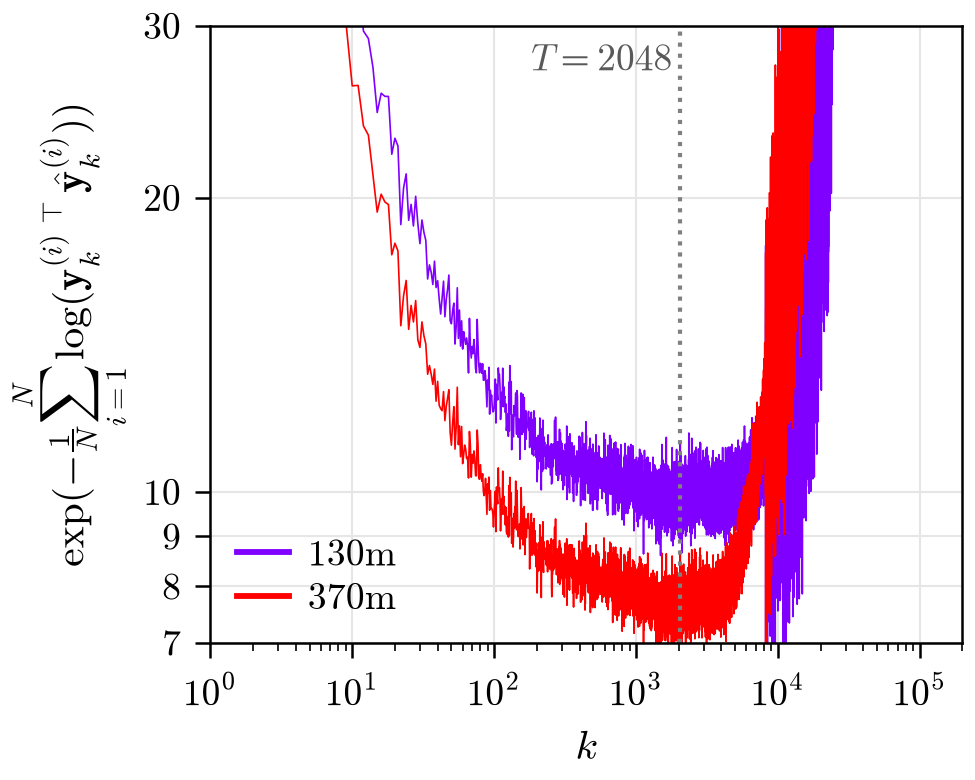}%
        }{%
            \fbox{\parbox[c][0.28\textheight][c]{0.92\linewidth}{\centering\small Missing asset: \\ \texttt{figures/mamba\_1.png}}}%
        }
        \caption{Mamba-1}
        \label{fig:mamba-1}
    \end{subfigure}\hfill
    \begin{subfigure}[t]{0.333\textwidth}
        \centering
        \IfFileExists{figures/mamba_2.png}{%
            \includegraphics[width=\linewidth]{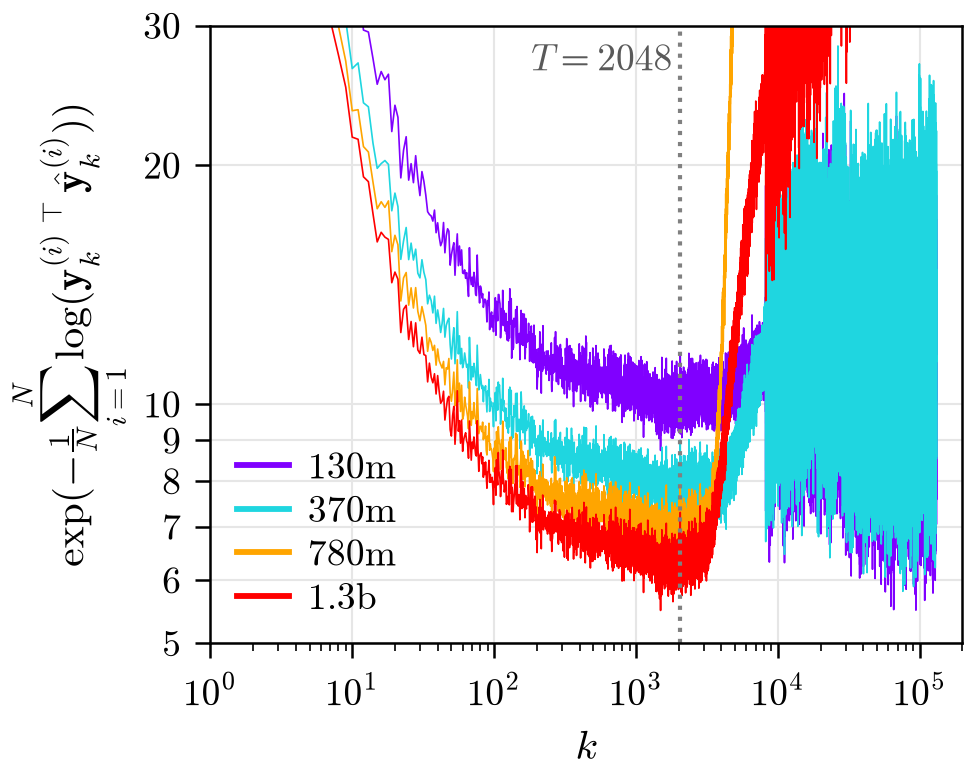}%
        }{%
            \fbox{\parbox[c][0.28\textheight][c]{0.92\linewidth}{\centering\small Missing asset: \\ \texttt{figures/mamba\_2.png}}}%
        }
        \caption{Mamba-2}
        \label{fig:mamba-2}
    \end{subfigure}\hfill
    \begin{subfigure}[t]{0.333\textwidth}
        \centering
        \IfFileExists{figures/mamba3.png}{%
            \includegraphics[width=\linewidth]{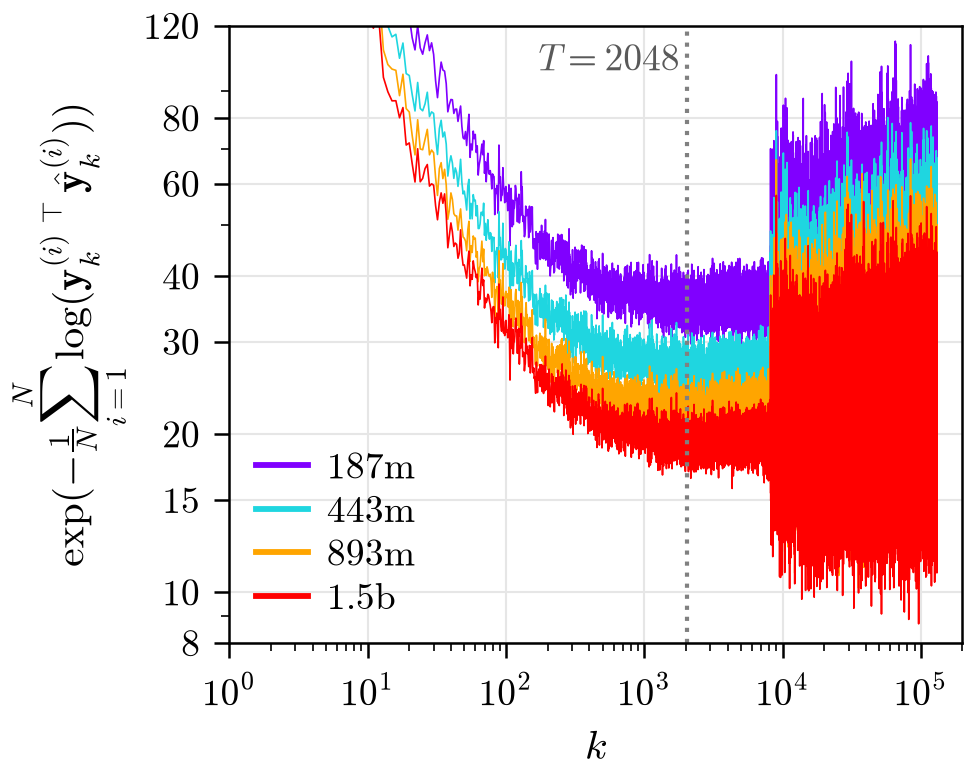}%
        }{%
            \fbox{\parbox[c][0.28\textheight][c]{0.92\linewidth}{\centering\small Missing asset: \\ \texttt{figures/mamba3.png}}}%
        }
        \caption{Mamba-3}
        \label{fig:mamba-3}
    \end{subfigure}

    \caption{The performance of (a) Mamba-1, (b) Mamba-2, and (c) Mamba-3 at different model sizes (shown in different colors) on sequences from the Pile validation set that are longer than those seen during training. The dotted vertical line marks the training horizon $T = 2048$. The evaluation metric is the \emph{position-wise perplexity}, which is the exponential of the per-step prediction error under the cross-entropy loss \citep{ruiz2025understanding}. See \Cref{appendix:mamba-experiment} for the details.} 
    \label{fig:mamba}
\end{figure*}

Designing sequence models that can effectively process arbitrarily long sequences is challenging. Although transformers have dominated sequence modeling, the quadratic cost of attention makes processing arbitrarily long sequences prohibitively expensive \citep{vaswani2017attention}. This has motivated the development of modern recurrent architectures whose per-token inference cost is constant, such as linear attention \citep{katharopoulos2020transformers, choromanski2020rethinking, peng2021random, qin2022devil} and its gated \citep{yang2023gated, sun2023retentive, katsch2023gateloop, qin2023hierarchically, peng2023rwkv, de2024griffin, beck2024xlstm} and delta-rule variants \citep{adaptiveswitchingcircuits, yang2024parallelizing, yang2025gated, iriegoing, siems2026deltaproduct}, state-space models (SSMs) \citep{gu2020hippo, gu2021combining, gu2021efficiently, gupta2022diagonal, smith2022simplified,fu2022hungry,orvieto2023resurrecting, gu2023mamba, dao2024transformers, lahoti2026mamba}, fast weight programmers \citep{hinton1987using,schmidhuber1992learning, ba2016using}, and test-time training layers \citep{sun2020test, sun2024learning, behrouz2024titans, behrouz2026s, zhang2026test}. However, these models often fail to length generalize, which undermines the main advantage that motivated them \citep{ruiz2025understanding, zhong2026rethinking}. For instance, as shown in \Cref{fig:mamba} \citep[see also][Fig.~1]{ruiz2025understanding}, when the context length exceeds the training horizon, the performance of the state-of-the-art SSMs degrades substantially. Unfortunately, our theoretical understanding of why these models fail beyond the training horizon remains limited.

In classical statistical learning theory, \emph{generalization} is defined as the ability of a statistical model to predict outcomes for previously unseen data. Length generalization is a special case of this notion that arises in sequence modeling. In sequence modeling, unlike classical machine learning problems such as classification and regression, the learned map is defined on inputs and outputs of varying length rather than on a fixed-dimensional domain and codomain.\footnote{Although the vocabulary is fixed, the length of the output, which is terminated by an end-of-sequence token in an autoregressive model, varies with the given context.}

Numerous empirical studies have reported that training with large learning rates improves generalization \citep{jastrzkebski2017three, li2019towards, smith2021origin, andriushchenko2023sgd, lobacheva2023large, barsbey2025large}. Most notably, \citet{anil2022exploring} observed that large learning rates lead to better length generalization when training transformer-based models. This effect is commonly attributed to the \emph{implicit regularization} of gradient-based training, which has been investigated in several works \citep{keskar2016large,smith2017don, barrett2020implicit,lewkowycz2020large, sadrtdinov2024large,andriushchenko2023modern,mohtashami2023special,wang2025training, galli2026flatland}. For instance, \citet{wang2022large} showed that, in depth-2 matrix factorization, gradient descent (GD) with a large learning rate balances the Frobenius norms of the factors in the limit, even when these norms are significantly imbalanced at initialization. This balancing effect can be viewed as an \emph{implicit bias} of GD towards certain solutions. Note that an implicit bias mechanism does not necessarily explain or cause an implicit regularization unless the global minima favored by the bias have been shown to be benign or malign solutions.  To the best of our knowledge, the implicit regularization induced by the learning rate has not previously been investigated in the context of length generalization in recurrent architectures.

In this paper, we investigate how the learning rate shapes the implicit bias of GD when it selects among global minimizers that differ in their length-generalization ability. Motivated by the length-generalization failure of linear recurrent models such as Mamba, we study a single-layer linear recurrent neural network (RNN) trained with the squared-error loss on a next-token prediction task in which tokens are drawn from a vector-valued random process. In other words, the network is trained to predict the next token in the minimum-mean-square-error (MMSE) sense.

In recurrent architectures, memory is built incrementally: It consists of a fixed-size hidden state that is updated recurrently over time. Due to its fixed size, the information from the history that is pertinent to predicting the next token can be attenuated or lost entirely as it propagates through the forward pass during training. We therefore hypothesize that training may not effectively teach the recurrent model to generalize to longer sequences. To eliminate this effect in training, we choose a parameterization regime in which the linear RNN is able to store its entire context in its hidden state.

Furthermore, we show in \Cref{sec:main} that, in this regime, if the zero-padded history matrix $\mathbf{Z}$ (see \Cref{remark:representation}) has full rank, all global minimizers realize the same Markov parameters within the training horizon. We further show how global minimizers differ in their length-generalization ability beyond the training horizon, even though they realize the same Markov parameters within the training horizon. This allows us to eliminate other generalization effects and study how the learning rate shapes the implicit bias of GD among minima that differ only in their length-generalization performance.

We present our main results in \Cref{sec:main} and \Cref{sec:Experiments}. In \Cref{sec:Discussion}, we discuss our main results, and we conclude in \Cref{sec:Conclusion}.

\subsection{Contributions}
In this paper, motivated by the length-generalization failure of linear recurrent models, we study how the learning rate determines which global minimizer GD converges to, when these minima differ in their ability to generalize to arbitrarily long sequences, in a single-layer linear RNN trained with squared-error loss on a next-token prediction task. At a glance, our contributions are as follows: 

\begin{itemize}
    \item We identify a parameterization regime where the linear RNN can realize any sequence of Markov parameters within the training horizon. In this regime, for every monic polynomial of degree $d$, there exists a global minimizer whose recurrent weight matrix $\mathbf{W}_1 \in \mathbb{R}^{d \times d}$ has this polynomial as its characteristic polynomial (\Cref{lemma:uniqueness} and \Cref{thm:arbitrary-charpoly}).
    
    \item We empirically demonstrate that single-layer linear RNNs trained with the squared-error loss on a next-token prediction task exhibit \emph{progressive sharpening} and operate at the \emph{edge of stability} (\Cref{fig:sharpness}).

    \item Our experiments show that sufficiently large learning rates induce edge-of-stability dynamics, and the resulting global minima exhibit worse length generalization than those reached at smaller learning rates, for which the dynamics remain outside this regime (\Cref{fig:sharpness} and \Cref{fig:test-error}). This implies that, given that length generalization is a special case of the notion of generalization, a relatively flat minimum need not generalize better.

    \item We show that among runs that enter the edge-of-stability regime, larger learning rates further degrade length generalization, contrary to observations for transformer-based models \citep{anil2022exploring}.

    \item Among these runs, we observe that larger learning rates favor global minima whose recurrent weight matrices have a larger spectral radius restricted to the controllable and observable modes (\Cref{fig:radius}). 

    \item As we observe in \Cref{sec:Experiments} and several works in the literature have shown, larger learning rates converge to relatively flat minima. To understand the relationship between the sharpness of a global minimum and the spectral radius of its recurrent weight matrix restricted to the controllable and observable modes, denoted by $\rho_{\text{co}}$, we derive a closed-form expression for the maximum Hessian eigenvalue of the training loss at any global minimum (\Cref{thm:exact-sharpness}). We show that, for every global minimum, there exists a global minimum with the same $\rho_{\text{co}}$ that is arbitrarily sharp (\Cref{prop:rescaling}) and that $\rho_{\text{co}}(\cdot)$ is unbounded over the set of global minima (\Cref{prop:unbounded-rho}). Therefore, through the lens of dynamical stability, we hypothesize that the sharpness of flat minima among the global minima with a given $\rho_{\text{co}} = r$ is a non-increasing and non-constant function of $r$ (\Cref{hyp:flat}). Under this hypothesis, linear stability excludes only the values of $\rho_{\text{co}}$ below a threshold that does not decrease as the learning rate increases.

    \item \citet{ruiz2025understanding} hypothesized that linear recurrent models fail to length generalize since they are trained only on a subset of the state distributions determined by the training horizon. Our results, however, demonstrate that although all global minima realize the same Markov parameters within the training horizon, their behavior beyond it can differ arbitrarily. Therefore, the training horizon alone does not determine length generalization, and a theory of length generalization should account for the optimization dynamics of gradient-based training.
\end{itemize}

\subsection{Related Work}

\paragraph{Linear RNNs and Modern Recurrent Architectures.}
Most linear recurrent architectures admit the general form
\begin{equation*}
    \mathbf{h}_{k} = \mathbf{A}_k \mathbf{h}_{k-1} + \mathbf{B}_k \mathbf{u}_k, \qquad \mathbf{o}_k = \mathbf{C}_k \mathbf{h}_{k},
\end{equation*}
where $\mathbf{u}_k$ is the input token and $\mathbf{o}_k$ is the output token. These architectures compress the history into a fixed-size recurrent memory that is updated at every time step. Conceptually, they differ in how $\mathbf{A}_k$, $\mathbf{B}_k$, and $\mathbf{C}_k$ are parameterized and how they depend on the input \citep[Section~2.2]{ruiz2025understanding}. For instance, linear attention projects each token onto a key $\mathbf{k}_k = \mathbf{W}_K \mathbf{u}_k \in \mathbb{R}^{d_k}$, a value $\mathbf{v}_k = \mathbf{W}_V \mathbf{u}_k \in \mathbb{R}^{d_v}$, and a query $\mathbf{q}_k = \mathbf{W}_Q \mathbf{u}_k \in \mathbb{R}^{d_k}$, and then stores their associations in a recurrent matrix memory $\mathbf{S}_k = \mathbf{S}_{k-1} + \mathbf{v}_k \mathbf{k}_k^\top$. Subsequently, its recall is simply the matrix-vector product $\mathbf{o}_k = \mathbf{S}_k \mathbf{q}_k = \sum_{i \leq k} (\mathbf{k}_i^\top \mathbf{q}_k)\, \mathbf{v}_i$, which retrieves each stored value weighted by the dot-product similarity between its key and the query. Note that if $\mathbf{h}_k = \operatorname{vec}(\mathbf{S}_k)$, this corresponds to $\mathbf{A}_k = \mathbf{I}$, $\mathbf{B}_k = \mathbf{k}_k \otimes \mathbf{W}_V$ and $\mathbf{C}_k = \mathbf{q}_k^\top \otimes \mathbf{I}_{d_v}$ by using the properties of the Kronecker product. Since vanilla linear attention suffers from a linear memory, its gated \citep{yang2023gated,sun2023retentive,katsch2023gateloop,qin2023hierarchically,peng2023rwkv,de2024griffin} and delta-rule variants \citep{schlag2021linear,yang2024parallelizing,yang2025gated,siems2026deltaproduct} have been proposed. Its gated variants replace $\mathbf{A}_k$ with a data-dependent decay $\gamma_k\mathbf{A}_k$, while its delta-rule variants instead choose $\mathbf{A}_k = (\mathbf{I} - \beta_k \mathbf{k}_k \mathbf{k}_k^\top)\otimes \mathbf{I}_{d_v}$ and $\mathbf{B}_k = \beta_k (\mathbf{k}_k \otimes \mathbf{W}_V)$ with a learning rate $\beta_k$. 

On the other hand, the recurrent memory of an SSM is chosen a priori. HiPPO \citep{gu2020hippo} derives the state and input matrices of a continuous-time linear system so that its state stores the coefficients of the optimal projection of the history onto a basis of orthogonal polynomials. S4 discretizes this continuous-time system into matrices $(\bar{\mathbf{A}}, \bar{\mathbf{B}}, \bar{\mathbf{C}})$, after initializing the continuous-time recurrent matrix $\mathbf{A}$ with HiPPO \citep{gu2021combining,gu2021efficiently}. Since the resulting system is time-invariant, its output is the convolution of the input with the kernel $(\bar{\mathbf{C}}\bar{\mathbf{B}}, \bar{\mathbf{C}}\bar{\mathbf{A}}\bar{\mathbf{B}}, \bar{\mathbf{C}}\bar{\mathbf{A}}^2\bar{\mathbf{B}}, \dots)$. This convolution can be evaluated in parallel by the fast Fourier transform in $\mathcal{O}(T \log T)$ time rather than through $T$ sequential recurrent updates, which makes training substantially faster than that of nonlinear RNNs, whose recurrence cannot be parallelized over time. Subsequent works restrict the state matrix to be diagonal \citep{gupta2022diagonal,smith2022simplified,orvieto2023resurrecting}. In particular, \citet{orvieto2023resurrecting} showed that a linear RNN matches the performance of S4 without discretization or HiPPO, provided its recurrence is diagonal and complex-valued with eigenvalues parameterized to lie inside the unit disk.  

\paragraph{Online Test-Time Regression.} Recently, a unifying perspective for sequence models has been proposed \citep{wang2025testtime}. In this view, during inference, sequence models form a memory from the key--value pairs computed from each token processed so far, approximately solve a regression problem that maps these keys to their corresponding values, and then evaluate the fitted regressor at a query to produce its output. The regression is \emph{online} because the memory is updated as each token arrives and it is performed at \emph{test time}. From this point of view, softmax attention, linear attention and its gated and delta-rule variants, state-space models, fast weight programmers, and test-time training layers differ in how the associations are weighted, whether the regressor is parametric or non-parametric, and the algorithm that solves the regression \citep{katharopoulos2020transformers,schlag2021linear,yang2024parallelizing,sun2024learning,liu2024longhorn,behrouz2024titans,wang2025testtime,vonoswald2025mesanet,mcdermott2026lifelong}.

\paragraph{Edge of Stability and Progressive Sharpening.}
\citet{cohen2021gradient} reported that neural networks trained with GD generally enter a regime called the \emph{edge of stability}, in which the maximum eigenvalue of the Hessian of the training loss hovers around $2/\eta$, where $\eta$ is the learning rate, while the training loss is decreasing non-monotonically. They also reported that this regime is preceded by a phase called \emph{progressive sharpening}, during which the maximum Hessian eigenvalue steadily increases until it reaches $2/\eta$ \citep{jastrzebski2020break}. Furthermore, a decrease in the maximum Hessian eigenvalue, followed by a temporary plateau, has been observed early in training and called \emph{sharpness reduction} \citep{kalra2023phase, kalra2025universal}. This has raised the question of whether these phenomena occur universally in neural network training under certain conditions. To this end, \citet{wang2022analyzing} and \citet{zhu2022understanding} analyzed two-layer linear networks. \citet{cohen2021gradient} reported that progressive sharpening is weak for shallow networks trained with gradient flow or, at a fixed depth, for small datasets. Moreover, \citet{yoo2025understanding} reported that, in stochastic gradient descent (SGD) with a small batch size, progressive sharpening is weaker for large learning rates. Recently, \citet{galli2026flatland} showed that GD with large adaptive step sizes operates at the edge of stability from the start of training. 

\paragraph{Flat Minima and Generalization.}
One of the mechanisms that give rise to the implicit bias of GD has been characterized in terms of \emph{dynamical stability}. Conceptually, dynamical stability describes the ability of GD to converge stably to a minimum, and it is closely related to the sharpness of the loss landscape near that minimum. We recall that \emph{an implicit bias mechanism does not necessarily explain an implicit regularization unless the global minima favored by an algorithm have been shown to be benign or malign solutions relative to others}. For instance, it has been shown that sharp minima can be dynamically unstable for gradient-based methods; therefore, gradient iterates avoid sharp minima and converge to flat minima \citep{Wustab,mulayoff2021implicit,nacson2022implicit}. However, the relationship between flatness and generalization remains unclear: In a large-scale empirical analysis, \citet{Jiang2020Fantastic} evaluated a set of complexity measures for deep networks and observed that sharpness-based measures were most strongly correlated with generalization. However, this correlation does not establish a causal relationship between flatness and generalization. Nonetheless, there exists theoretical evidence for this phenomenon in low-rank matrix recovery~\citep{ding2024flat} and two-layer neural networks~\citep{qiao2024stable,liang2026stable,liang2026generalization}. On the other hand, \citet{sharpminimacangeneralize} demonstrated that even good minima in deep neural networks can be arbitrarily sharp, and \citet{xu2026doessgdseekflatness} showed that SGD does not seek flat minima. Therefore, the implicit bias of gradient-based methods toward flat minima has not yet been characterized as an implicit regularization mechanism, although the two terms are used interchangeably in the literature.

\section{Notation, Problem Setup, and Preliminaries} 

We use the following notation throughout the paper. We denote the \emph{Kronecker product} by $\otimes$, the $\ell^2$ norm by $\|\cdot\|_2$, the \emph{spectral norm} by $\sigma_{\max}(\cdot)$, and the \emph{Frobenius norm} by $\|\cdot\|_F$. We write $[L] := \{1,2,\dots,L\}$ for the set of natural numbers up to $L$ and $[L]_0 := \{0,1,2,\dots,L\}$ for the set of natural numbers up to $L$ including $0$. We denote by $\mathbf{1} := [1 \quad 1 \quad \cdots \quad 1]^\top \in \mathbb{R}^n$ the vector whose entries are all equal to $1$ and by $\mathbb{R}_+^n$ the non-negative orthant of $\mathbb{R}^n$.  Furthermore, we denote a scalar by $a \in \mathbb{R}$, an $n$-dimensional vector by $\mathbf{a} \in \mathbb{R}^n$, and an $m \times n$ matrix by $\mathbf{A} \in \mathbb{R}^{m \times n}$. We denote the range (column space) of a matrix $\mathbf{A}$ by $\operatorname{ran}(\mathbf{A})$ and its kernel (null space) by $\operatorname{ker}(\mathbf{A})$. We denote an $n$-by-$n$ identity matrix by $\mathbf{I}_n \in \mathbb{R}^{n\times n}$ and denote an $m$-by-$n$ zero matrix by $\mathbf{0}_{m\times n} \in \mathbb{R}^{m \times n}$. Moreover, we denote random variables by capital letters, e.g., $Y \in \mathbb{R}$, and $n$-dimensional random vectors by bold and calligraphic letters, e.g., $\boldsymbol{\mathcal{Y}} \in \mathbb{R}^n$. Finally, we denote a sequence by $\{x_i\}_{i \geq 0}$ and a sequence indexed by $[t]_0$ by either $\{x_i\}_{i \leq t}$ or $x^t$. 

\subsection{Problem Setup}
\label{sec:ProblemSetup}
We consider a partially observed linear dynamical system governed by white Gaussian noise sequences,
\begin{align}
\boldsymbol{\mathcal{X}}_{k+1} &= \mathbf{A}\boldsymbol{\mathcal{X}}_k + \boldsymbol{\mathcal{W}}_k, \nonumber \\
\boldsymbol{\mathcal{Y}}_k &= \mathbf{C}\boldsymbol{\mathcal{X}}_k + \boldsymbol{\mathcal{V}}_k, \label{LDS}
\end{align}
where $\boldsymbol{\mathcal{X}}_k \in \mathbb{R}^n$ is the \emph{hidden state}, $\boldsymbol{\mathcal{Y}}_k \in \mathbb{R}^m$ is the \emph{observation}, $\mathbf{A} \in \mathbb{R}^{n \times n}$ is the \emph{state transition} matrix, and $\mathbf{C} \in \mathbb{R}^{m \times n}$ is the \emph{observation matrix}. The \emph{process noise} $\{\boldsymbol{\mathcal{W}}_k\}_{k \geq 0}$ and the \emph{measurement noise} $\{\boldsymbol{\mathcal{V}}_k\}_{k \geq 0}$ are mutually independent, vector-valued random sequences with
$\boldsymbol{\mathcal{W}}_k \overset{\mathrm{i.i.d.}}{\sim} \mathcal{N}(\mathbf{0}, \mathbf{Q})$ and
$\boldsymbol{\mathcal{V}}_k \overset{\mathrm{i.i.d.}}{\sim} \mathcal{N}(\mathbf{0}, \mathbf{R})$, where $\mathbf{R}$ is assumed to be positive definite. We further assume that $\boldsymbol{\mathcal{X}}_0 \sim \mathcal{N}(\bar{\mathbf{x}}_0, \mathbf{P}_0)$ and that $\boldsymbol{\mathcal{X}}_0$, $\{\boldsymbol{\mathcal{W}}_k\}_{k \geq 0}$, and $\{\boldsymbol{\mathcal{V}}_k\}_{k \geq 0}$ are mutually independent.

After having observed the system up to time step $k \in \mathbb{N}$, we wish to make a prediction for $\boldsymbol{\mathcal{Y}}_{k+1} \in \mathbb{R}^m$ in the MMSE sense.\footnote{Human cognitive processes expressed through language can be modeled as a stochastic dynamical system in which the observations are discrete probability distributions over a finite vocabulary, i.e., $
\mathbf{y}_k \in \Delta_m := \left\{ \mathbf{p} \in \mathbb{R}_+^m : \mathbf{p}^\top \mathbf{1} = 1 \right\},$
for which the Kullback--Leibler divergence is a more suitable discrepancy measure than the mean-squared error.} Accordingly, we fix a training horizon $T \in \mathbb{N}$, observe $N \in \mathbb{N}$ i.i.d.\ trajectories of the system, denoted by $\{(\mathbf{y}^{(i)}_0, \mathbf{y}^{(i)}_1, \ldots, \mathbf{y}^{(i)}_T)\}_{i=1}^N$, and train a single-layer linear RNN of the form
\begin{align}
    \mathbf{h}_{k+1} &= \mathbf{W}_1 \mathbf{h}_{k} + \mathbf{W}_2 \mathbf{y}_{k}, \nonumber \\
    \hat{\mathbf{y}}_{k+1} &= \mathbf{W}_3 \mathbf{h}_{k+1},
    \label{RNN}
\end{align}
where $\mathbf{h}_k \in \mathbb{R}^{d}$ is the \emph{hidden state}, 
$\mathbf{W}_1 \in \mathbb{R}^{d \times d}$ is the \emph{recurrent weight matrix}, $\mathbf{W}_2 \in \mathbb{R}^{d \times m}$ is the \emph{input matrix}, 
and $\mathbf{W}_3 \in \mathbb{R}^{m \times d}$ is the \emph{output matrix}. We denote by $\hat{\mathbf{y}}_{k+1}$ our prediction for $\mathbf{y}_{k+1}$ for every $k \in \mathbb{N}$. Then, we train the linear RNN by minimizing the averaged squared-error loss $\hat{\mathcal{L}}\colon \Theta \to \mathbb{R}_+$, defined as
\begin{equation}
\label{oop}
    \hat{\mathcal{L}}(\boldsymbol{\theta}) := \frac{1}{NT} \sum_{i=1}^{N} \sum_{k=0}^{T-1} \big\| \mathbf{y}^{(i)}_{k+1} - \hat{\mathbf{y}}^{(i)}_{k+1}(\boldsymbol{\theta}) \big\|_2^2,
\end{equation}
where $\boldsymbol{\theta} := (\mathbf{W}_1, \mathbf{W}_2, \mathbf{W}_3) \in \Theta := \mathbb{R}^{d \times d} \times \mathbb{R}^{d \times m} \times \mathbb{R}^{m \times d}$. Without loss of generality, we initialize the linear RNN with the zero hidden state, i.e., $\mathbf{h}_0=\mathbf{0}$, which is a common initialization in modern recurrent models \citep{ruiz2025understanding}. Furthermore, we denote the set of global minimizers of $\hat{\mathcal{L}}$ by
\begin{equation}
    \Omega := \argmin_{\boldsymbol{\theta} \in \Theta} \hat{\mathcal{L}}(\boldsymbol{\theta}).
\end{equation}

\begin{lemma}
\label[lemma]{remark:representation}
Consider the training objective in \eqref{oop}. Define the history matrix
by $\mathbf{Y}_k := [\mathbf{y}_k^{(1)} \ \cdots \ \mathbf{y}_k^{(N)}]
\in \mathbb{R}^{m \times N}$. Then, under zero initialization, i.e., $\mathbf{h}_0 = \mathbf{0}$, we can rewrite the optimization objective equivalently as 
\begin{equation}
    \hat{\mathcal{L}}(\boldsymbol{\theta})
    = \frac{1}{NT}
    \left\|
        \operatorname{vec}(\mathbf{Y}_+)
        - (\mathbf{Z}^{\top} \otimes \mathbf{I}_m)
        \operatorname{vec}(\mathbf{\Phi}_{T}(\boldsymbol{\theta}))
    \right\|_2^2,
\end{equation}
where 
\begin{equation}
    \mathbf{Z}
    := \begin{bmatrix}
        \mathbf{Y}_0 & \mathbf{Y}_1 & \cdots & \mathbf{Y}_{T-1} \\
        \mathbf{0}   & \mathbf{Y}_0 & \cdots & \mathbf{Y}_{T-2} \\
        \vdots       & \ddots       & \ddots & \vdots \\
        \mathbf{0}   & \cdots       & \mathbf{0} & \mathbf{Y}_0
    \end{bmatrix}
    \in \mathbb{R}^{mT \times NT}\quad \text{and} \quad
    \mathbf{Y}_+
    := [\mathbf{Y}_1 \ \cdots \ \mathbf{Y}_T]
    \in \mathbb{R}^{m \times NT}
\end{equation}
are the zero-padded history matrix and the corresponding target
matrix, respectively.
\end{lemma}
\begin{proof}
With zero initialization, $\hat{\mathbf{y}}_{k+1}$ can be written as a discrete
convolution of the input sequence with its \emph{Markov parameters}
\begin{equation}
    \hat{\mathbf{y}}_{k+1}
    = \sum_{i=0}^{k} \mathbf{M}_i(\boldsymbol{\theta}) \mathbf{y}_{k-i},
\end{equation}
where $\mathbf{M}_i(\boldsymbol{\theta}) := \mathbf{W}_3 \mathbf{W}_1^i \mathbf{W}_2$. Furthermore, we can define a dynamic, matrix-valued map
$\mathbf{\Phi}_k(\boldsymbol{\theta})
:= [\mathbf{M}_0(\boldsymbol{\theta}) \ \cdots \ \mathbf{M}_{k-1}(\boldsymbol{\theta})]
\in \mathbb{R}^{m \times mk}$ to express $\hat{\mathbf{y}}_{k+1}$ as a linear regression on the history as 

\begin{equation}
    \hat{\mathbf{y}}_{k+1} = \mathbf{\Phi}_{k+1}(\boldsymbol{\theta})\begin{bmatrix}
        \mathbf{y}_{k}^\top & \mathbf{y}_{k-1}^\top & \ldots & \mathbf{y}_{0}^\top
    \end{bmatrix}^\top
\end{equation}
for every $k \geq 0$. Consequently, 
\begin{equation}
    \hat{\mathcal{L}}(\boldsymbol{\theta})
    = \frac{1}{NT}
    \left\|
        \mathbf{Y}_+
        - \mathbf{\Phi}_{T}(\boldsymbol{\theta})\mathbf{Z}
    \right\|_F^2 
    = \frac{1}{NT}
    \left\|
        \operatorname{vec}(\mathbf{Y}_+)
        - (\mathbf{Z}^{\top} \otimes \mathbf{I}_m)
        \operatorname{vec}(\mathbf{\Phi}_{T}(\boldsymbol{\theta}))
    \right\|_2^2,
    \label{equivalentrepresentation}
\end{equation}
by using the identity
$\operatorname{vec}(\mathbf{A}\mathbf{X}\mathbf{B})
= (\mathbf{B}^\top \otimes \mathbf{A})\operatorname{vec}(\mathbf{X})$. Thus, $\hat{\mathcal{L}}$ is a convex quadratic function of the Markov parameters restricted to the range of the map
$\boldsymbol{\theta} \mapsto \mathbf{\Phi}_{T}(\boldsymbol{\theta})$.
\end{proof}
\Cref{remark:representation} provides a key insight into how the linear RNN predicts the next token: It performs a regression on its history. This regression is \emph{online} because $\mathbf{\Phi}_k$ is updated at each time step $k$ by a function of the parameters learned from training. Note that if the range of the map $\boldsymbol{\theta} \mapsto \mathbf{\Phi}_{T}$ does not contain the optimal Markov parameters for the least-squares regression of the history to predict the next token, the training is suboptimal. This is due to the limited memory capacity of the linear RNN as we mention in \Cref{sec:main}. Moreover, the optimization landscape is non-convex; hence, it may contain spurious minima and non-strict saddle points. Therefore, even with a linear RNN that possesses sufficient memory capacity to realize optimal Markov parameters, we might not be able to reach them since we rely on gradient-based methods for the training, and they can be trapped by these points.

\subsection{Preliminaries}
\label{sec:Preliminaries}

We note that the linear RNN in \eqref{RNN} is itself a linear dynamical system with input $\mathbf{y}_k$, state $\mathbf{h}_k$, and output $\hat{\mathbf{y}}_{k+1}$. Therefore, before delving into the main results, we review the fundamental concepts of linear system theory. Throughout this section, we assume that $\mathbf{h}_0 = \mathbf{0}$. Furthermore, we fix $\boldsymbol{\theta} = (\mathbf{W}_1, \mathbf{W}_2, \mathbf{W}_3) \in \Theta$ and recall that $\mathbf{M}_i(\boldsymbol{\theta}) = \mathbf{W}_3 \mathbf{W}_1^i \mathbf{W}_2$ for every $i \geq 0$. We denote by $\mathrm{GL}_d(\mathbb{R})$ the group of invertible  $d \times d$ real matrices. We define the spectral radius as follows: 

\begin{definition}[{\citealp[Definition 1.2.9]{horn2012matrix}}]
    Suppose $\mathbf{A} \in \mathbb{R}^{n \times n}$ is an arbitrary square matrix, and let
    $\operatorname{spec}(\mathbf{A}) := \{\lambda \in \mathbb{C} : \mathbf{A}\mathbf{v} = \lambda \mathbf{v} \text{ for some } \mathbf{v} \in \mathbb{C}^n \setminus \{\mathbf{0}\}\}$
    denote its spectrum. Then, we define the spectral radius of $\mathbf{A}$ by
    \begin{equation}
        \rho(\mathbf{A}) := \max \{{|\lambda| \in \mathbb{R}: \lambda \in \operatorname{spec}(\mathbf{A})}\}.
    \end{equation}
\end{definition}

By \Cref{remark:representation}, the prediction of the linear RNN, and hence $\hat{\mathcal{L}}$, depends on $\boldsymbol{\theta}$ only through its Markov parameters. This motivates the following definition.
\begin{definition}[{\citealp[Theorem 4.1]{chen1984linear}}]
\label{def:realization}
Let $\{\mathbf{M}_i\}_{i \geq 0} \subset \mathbb{R}^{m \times m}$. A parameterization $\boldsymbol{\theta} = (\mathbf{W}_1, \mathbf{W}_2, \mathbf{W}_3) \in \mathbb{R}^{d \times d} \times \mathbb{R}^{d \times m} \times \mathbb{R}^{m \times d}$ is a \emph{realization of order $d$} of $\{\mathbf{M}_i\}_{i \geq 0}$ if $\mathbf{M}_i = \mathbf{W}_3 \mathbf{W}_1^i \mathbf{W}_2$ for every $i \geq 0$. Two parameterizations, possibly of different orders, are \emph{input-output equivalent} if they realize the same sequence of Markov parameters. 
\end{definition}

\begin{proposition}
\label[proposition]{prop:similarity}
For every $\mathbf{S} \in \mathrm{GL}_d(\mathbb{R})$, the
parameterization $
\boldsymbol{\theta}_{\mathbf{S}}
:=
(\mathbf{S}\mathbf{W}_1\mathbf{S}^{-1},
\mathbf{S}\mathbf{W}_2,
\mathbf{W}_3\mathbf{S}^{-1})
$ satisfies
$\mathbf{M}_i(\boldsymbol{\theta}_{\mathbf{S}})
= \mathbf{M}_i(\boldsymbol{\theta})$
for every $i \geq 0$.
Consequently,
$\hat{\mathcal{L}}(\boldsymbol{\theta}_{\mathbf{S}})
= \hat{\mathcal{L}}(\boldsymbol{\theta})$,
and $\Omega$ is invariant under
$\mathrm{GL}_d(\mathbb{R})$. Furthermore, the orbit
$\{\boldsymbol{\theta}_{\mathbf{S}} :
\mathbf{S} \in \mathrm{GL}_d(\mathbb{R})\}$
is a singleton if and only if
$\mathbf{W}_2 = \mathbf{0}$,
$\mathbf{W}_3 = \mathbf{0}$, and
$\mathbf{W}_1 = \lambda \mathbf{I}_d$
for some $\lambda \in \mathbb{R}$.
\end{proposition}

We defer the proof of \Cref{prop:similarity} to \Cref{appendix:proofofproposition}. This implies that, apart from this degenerate case, no global minimizer is isolated, and its parameterization $\boldsymbol{\theta}^*$ cannot be identified from the training data. We now introduce two important notions from linear systems and control theory, which have direct interpretations in terms of recurrent memory in linear RNNs. 
\begin{definition}[{\citealp[Theorems 6.1 and 6.O1]{chen1984linear}}]
\label{def:ctrb_obsv}
For every $k \geq 1$, the \emph{controllability matrix} and the \emph{observability matrix} of $\boldsymbol{\theta}$ are
\begin{equation}
    \mathcal{C}_k(\boldsymbol{\theta}) := \begin{bmatrix} \mathbf{W}_2 & \mathbf{W}_1\mathbf{W}_2 & \cdots & \mathbf{W}_1^{k-1}\mathbf{W}_2 \end{bmatrix} \in \mathbb{R}^{d \times mk}
    \quad \text{and} \quad
    \mathcal{O}_k(\boldsymbol{\theta}) := \begin{bmatrix} \mathbf{W}_3 \\ \mathbf{W}_3\mathbf{W}_1 \\ \vdots \\ \mathbf{W}_3\mathbf{W}_1^{k-1} \end{bmatrix} \in \mathbb{R}^{mk \times d},
\end{equation}
respectively. The pair $(\mathbf{W}_1, \mathbf{W}_2)$ is \emph{controllable} if $\operatorname{rank}(\mathcal{C}_d(\boldsymbol{\theta})) = d$, and the pair $(\mathbf{W}_3, \mathbf{W}_1)$ is \emph{observable} if $\operatorname{rank}(\mathcal{O}_d(\boldsymbol{\theta})) = d$.
\end{definition}

These notions have a direct interpretation in terms of the operations a recurrent unit performs at each time step: reading the previous hidden state $\mathbf{h}_k$ and generating $\mathbf{h}_{k+1}$. Unrolling \eqref{RNN} from $\mathbf{h}_0 = \mathbf{0}$ gives
\begin{equation}
    \mathbf{h}_k = \sum_{j=0}^{k-1} \mathbf{W}_1^{k-1-j}\mathbf{W}_2\mathbf{y}_j
    = \mathcal{C}_k(\boldsymbol{\theta})
    \begin{bmatrix} \mathbf{y}_{k-1}^\top & \mathbf{y}_{k-2}^\top & \cdots & \mathbf{y}_0^\top \end{bmatrix}^\top
    \quad \text{and} \quad
    \mathbf{\Phi}_k(\boldsymbol{\theta}) = \mathbf{W}_3\,\mathcal{C}_k(\boldsymbol{\theta})
    \label{eq:hidden_state_unrolled}
\end{equation}
for every $k \geq 1$. Thus, the hidden state is a linear encoding of the $mk$-dimensional history into $d$ dimensions, and it always lies in $\operatorname{ran}(\mathcal{C}_k(\boldsymbol{\theta}))$. If there exists a direction $\mathbf{u} \in \mathbb{R}^d$ such that $\mathbf{u}^\top \mathcal{C}_k(\boldsymbol{\theta}) = \mathbf{0}$, then the hidden state cannot be excited in the direction of $\mathbf{u}$ within $k$ steps for any input sequence. Therefore, controllability means that the hidden state $\mathbf{h}_k$ can be excited in every direction in $\mathbb{R}^d$ by some input sequence. On the other hand, observability means that, for a fixed sequence of inputs, every nonzero perturbation $\mathbf{z} \in \mathbb{R}^d$ of the hidden state $\mathbf{h}_{k_0}$ changes the prediction within $k$ steps. To see this, suppose that the hidden state is perturbed at time step $k_0 \in \mathbb{N}$ in the direction of $\mathbf{z} \in \mathbb{R}^d$. Then
\begin{align}
    \mathbf{h}_{k_0+1} &= \mathbf{W}_1 \mathbf{h}_{k_0} + \mathbf{W}_2 \mathbf{y}_{k_0}, \\
    \mathbf{h}^z_{k_0+1} &= \mathbf{W}_1 \mathbf{h}^z_{k_0} + \mathbf{W}_2 \mathbf{y}_{k_0},
\end{align}
where $\mathbf{h}^z_{k_0} = \mathbf{h}_{k_0} + \mathbf{z}$. This means that $\hat{\mathbf{y}}_{k_0}^z - \hat{\mathbf{y}}_{k_0} = \mathbf{W}_3 \mathbf{z}$ and $\hat{\mathbf{y}}_{k_0+1}^z - \hat{\mathbf{y}}_{k_0+1} = \mathbf{W}_3 \mathbf{W}_1 \mathbf{z}$. By unrolling the recursion, we see that $\hat{\mathbf{y}}_{k_0+j}^z - \hat{\mathbf{y}}_{k_0+j} = \mathbf{W}_3 \mathbf{W}_1^j \mathbf{z}$ for all $j \in \mathbb{N}$. This implies that if $\operatorname{ker}(\mathcal{O}_k(\boldsymbol{\theta})) = \{\mathbf{0}\}$, then for any nonzero $\mathbf{z}$, there exists $i \in [k-1]_0$ such that $\mathbf{W}_3\mathbf{W}_1^i\mathbf{z} \neq \mathbf{0}$. 

We now show that if the hidden state cannot be excited in a direction
$\mathbf{u} \in \mathbb{R}^d$ within the first $d$ steps, then it cannot be
excited in that direction at any later time. Similarly, for a fixed input
sequence, if a perturbation of the hidden state leads to no change in the
predictions within the first $d$ steps after the perturbation, then it cannot
be observed at any later time. This shows that the behavior of the linear RNN
beyond the training horizon is determined by the spectrum of the recurrent weight matrix of the parameterization obtained from training.

\begin{lemma}
\label[lemma]{lemma:cayley_hamilton}
Let $\chi(s) := \det(s\mathbf{I}_d - \mathbf{W}_1) = s^d + \sum_{j=0}^{d-1} a_j s^j$ be the characteristic polynomial of $\mathbf{W}_1$. Then,
\begin{enumerate}
    \item[(i)] $\mathbf{M}_{i+d}(\boldsymbol{\theta}) = -\sum_{j=0}^{d-1} a_j \mathbf{M}_{i+j}(\boldsymbol{\theta})$ for every $i \geq 0$.
    \item[(ii)] $\operatorname{ran}(\mathcal{C}_k(\boldsymbol{\theta})) = \operatorname{ran}(\mathcal{C}_d(\boldsymbol{\theta}))$ and $\ker(\mathcal{O}_k(\boldsymbol{\theta}))= \ker(\mathcal{O}_d(\boldsymbol{\theta}))$ for every $k \geq d$.
\end{enumerate}
\end{lemma}
\begin{proof}
By the Cayley--Hamilton theorem \citep{frobenius1878lineare}, $\mathbf{W}_1^d = -\sum_{j=0}^{d-1} a_j \mathbf{W}_1^j$. Multiplying from the left by $\mathbf{W}_3\mathbf{W}_1^i$ and from the right by $\mathbf{W}_2$ gives (i). By induction, we can show that $\mathbf{W}_1^k \in \operatorname{span}\{\mathbf{I}_d, \mathbf{W}_1, \ldots, \mathbf{W}_1^{d-1}\}$ for every $k \geq 0$. Hence, every column of $\mathbf{W}_1^k\mathbf{W}_2$ lies in $\operatorname{ran}(\mathcal{C}_d(\boldsymbol{\theta}))$, and this means that $\operatorname{ran}(\mathcal{C}_k(\boldsymbol{\theta})) \subseteq \operatorname{ran}(\mathcal{C}_d(\boldsymbol{\theta}))$. Since $\mathcal{C}_d(\boldsymbol{\theta})$ consists of the first $md$ columns of $\mathcal{C}_k(\boldsymbol{\theta})$, we have $ \operatorname{ran}(\mathcal{C}_d(\boldsymbol{\theta})) \subseteq \operatorname{ran}(\mathcal{C}_k(\boldsymbol{\theta}))$. Therefore, $\operatorname{ran}(\mathcal{C}_k(\boldsymbol{\theta})) = \operatorname{ran}(\mathcal{C}_d(\boldsymbol{\theta}))$. The claim for $\ker(\mathcal{O}_k(\boldsymbol{\theta}))$ follows by the same argument applied to the rows of $\mathcal{O}_k(\boldsymbol{\theta})$.
\end{proof}

\begin{remark}
\label[remark]{remark:coefficients}
    This means that the Markov parameters of a linear RNN with hidden dimension $d$ obey a linear recurrence of order $d$. Hence, the entire sequence $\{\mathbf{M}_i(\boldsymbol{\theta})\}_{i \geq 0}$, including its behavior beyond the training horizon $T$, is determined by $\{\mathbf{M}_i(\boldsymbol{\theta})\}_{i \leq d-1}$ and the coefficients of the characteristic polynomial of $\mathbf{W}_1$. 
\end{remark}
Using the Hautus lemma \citep{hautus1969controllability}, also known as the Popov--Belevitch--Hautus test, we define the spectral radius restricted to controllable and observable modes as follows: 
\begin{definition}
\label[definition]{def:rho-co}
    We denote the spectral radius restricted to controllable and observable modes (eigenvalues) of $\mathbf{W}_1$ as
    \begin{equation}
    \rho_{\text{co}}{(\boldsymbol{\theta})} := \max\left\{ |\lambda| : \lambda \in \operatorname{spec}(\mathbf{W}_1),\;
    \operatorname{rank}\left(\begin{bmatrix} \lambda \mathbf{I}_d - \mathbf{W}_1 & \mathbf{W}_2 \end{bmatrix}\right) = d,\;
    \operatorname{rank}\left(\begin{bmatrix} \lambda \mathbf{I}_d - \mathbf{W}_1 \\ \mathbf{W}_3 \end{bmatrix}\right) = d \right\},
\end{equation}
where we set $\rho_{\text{co}}(\boldsymbol{\theta}) := 0$ if this set is empty.
\end{definition}

\section{Within Training Horizon}
\label{sec:main}
To isolate length generalization from other generalization effects, we consider a setting in which all global minimizers realize the same Markov parameters within the training horizon. This allows us to study how the learning rate shapes the implicit bias of GD among minimizers that differ only in their length-generalization ability. To this end, the following lemma provides a sufficient condition for the single-layer linear RNN that is independent of the sequential data.

\begin{lemma}
\label[lemma]{lemma:uniqueness}
Consider the zero-initialized single-layer linear RNN defined in \Cref{sec:ProblemSetup}. If $d \geq mT$, then the optimal training loss $\hat{\mathcal{L}}(\boldsymbol{\theta}^*)$ can be computed by solving\footnote{If $\operatorname{ran}(\mathbf{Y}_+^\top) \nsubseteq \operatorname{ran}(\mathbf{Z}^\top)$, the solution is instead obtained via least squares (see \Cref{columnspace}).} the equation $\mathbf{Y}_+ = \mathbf{\Phi}_{T}(\boldsymbol{\theta})\mathbf{Z}$ for $\mathbf{\Phi}_{T}(\boldsymbol{\theta})$. Furthermore, if $\mathbf{Z}\mathbf{Z}^\top$ is nonsingular, then every pair $\boldsymbol{\theta}^*_1, \boldsymbol{\theta}^*_2 \in \Omega$ satisfies
\begin{equation}
\label{eq:normalequation}
\mathbf{\Phi}_T(\boldsymbol{\theta}^*_1)
    = \mathbf{\Phi}_T(\boldsymbol{\theta}^*_2) = \mathbf{Y}_+\mathbf{Z}^\top(\mathbf{Z}\mathbf{Z}^\top)^{-1}.
\end{equation}
\end{lemma}
We defer the full proof to \Cref{appendix:proof-uniqueness}. The proof relies on the fact that $\boldsymbol{\theta} \mapsto \mathbf{\Phi}_{T}(\boldsymbol{\theta})$ is \emph{surjective} when $d \geq mT$. Therefore, to establish the uniqueness of the optimal Markov parameters, it suffices to have sequences such that $\mathbf{Z}\mathbf{Z}^{\top}$ is nonsingular. Note that, in our setting, $\boldsymbol{\mathcal{Y}}_k$ is a Gaussian random vector with a \emph{positive-definite} covariance matrix. Consequently, if $N \geq m$, then $\mathbf{Z}\mathbf{Z}^{\top}$ is nonsingular almost surely. For the proof, see \Cref{almostsurely}.

In this regime, the linear RNN can stack its context in its hidden state by updating it recursively like a shift register. Furthermore, since we can compute the optimal loss $\hat{\mathcal{L}}(\boldsymbol{\theta}^*)$ when $d \geq mT$ as shown in \Cref{lemma:uniqueness}, we can train the linear RNN until its training loss gap $\hat{\mathcal{L}}(\boldsymbol{\theta}_k) - \hat{\mathcal{L}}(\boldsymbol{\theta}^*)$ reaches machine precision to obtain a global minimizer.

As we showed in \Cref{lemma:cayley_hamilton}, the Markov parameters beyond the training horizon are determined by $\{\mathbf{M}_i(\boldsymbol{\theta})\}_{i \leq d-1}$ and the coefficients of the characteristic polynomial of $\mathbf{W}_1$. In the following, we show that when $d \geq mT$, there exists a global minimum whose recurrent weight matrix has a characteristic polynomial with arbitrary coefficients.

\begin{theorem}
\label{thm:arbitrary-charpoly}
Consider the single-layer linear RNN defined in \Cref{sec:ProblemSetup}. If $d \geq mT$, then for every $(a_0, a_1, \ldots, a_{d-1}) \in \mathbb{R}^d$, there exists a global minimizer $\boldsymbol{\theta}^* = (\mathbf{W}_1^*, \mathbf{W}_2^*, \mathbf{W}_3^*) \in \Omega$ such that
\begin{equation}
    \det(s\mathbf{I}_d - \mathbf{W}_1^*) = s^d + \sum_{j=0}^{d-1} a_j s^j.
\end{equation}
\end{theorem}
We defer the proof to \Cref{appendix:proof-arbitrary-charpoly}. The proof combines the shift-register realization used in the proof of \Cref{lemma:uniqueness} with a \emph{companion matrix} of a monic polynomial with arbitrary coefficients (\Cref{companionmatrix}).

\section{Experiments: Beyond Training Horizon}
\label{sec:Experiments}
\begin{figure*}[t]
    \centering
    \captionsetup[subfigure]{justification=centering,singlelinecheck=false}

    \begin{subfigure}[t]{\textwidth}
        \centering
        \includegraphics[width=0.333\linewidth]{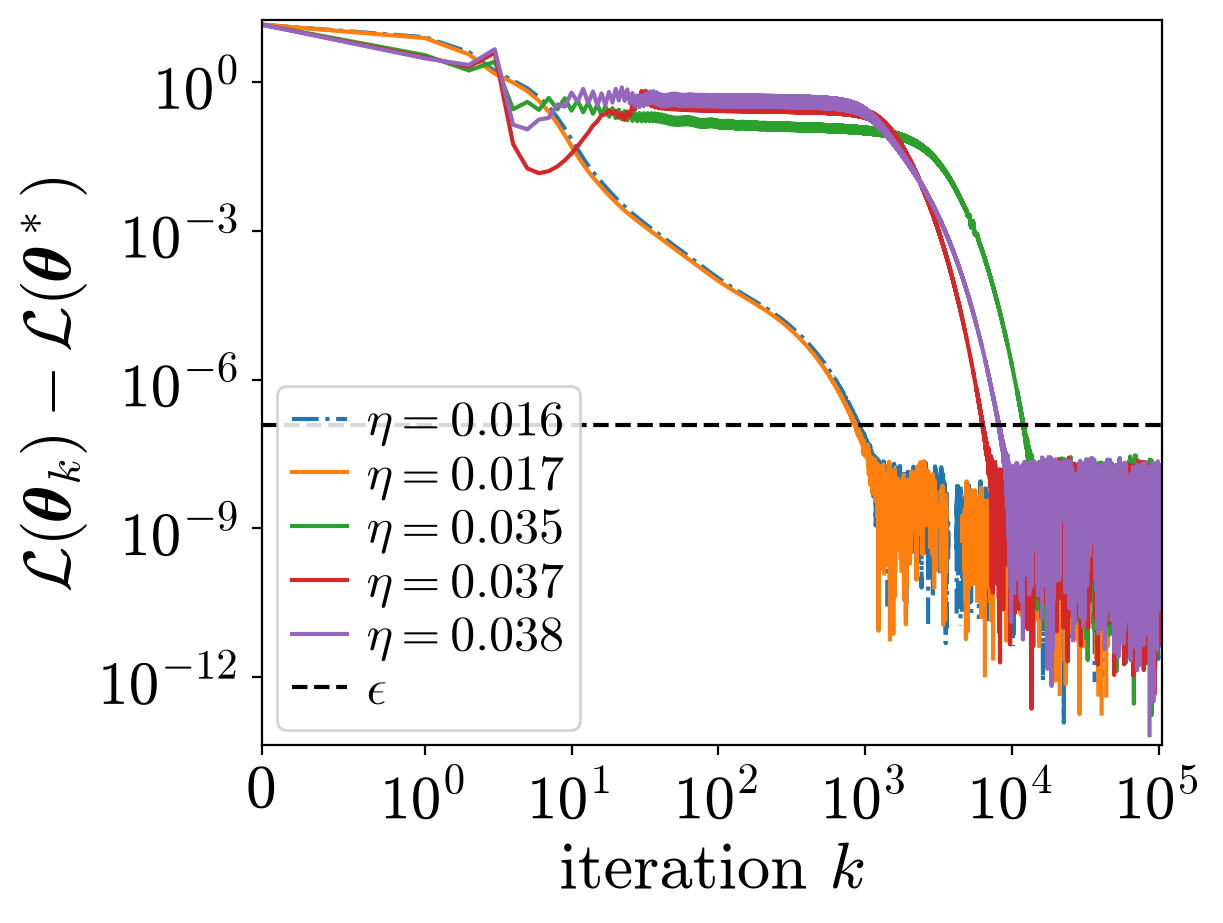}\hfill
        \includegraphics[width=0.333\linewidth]{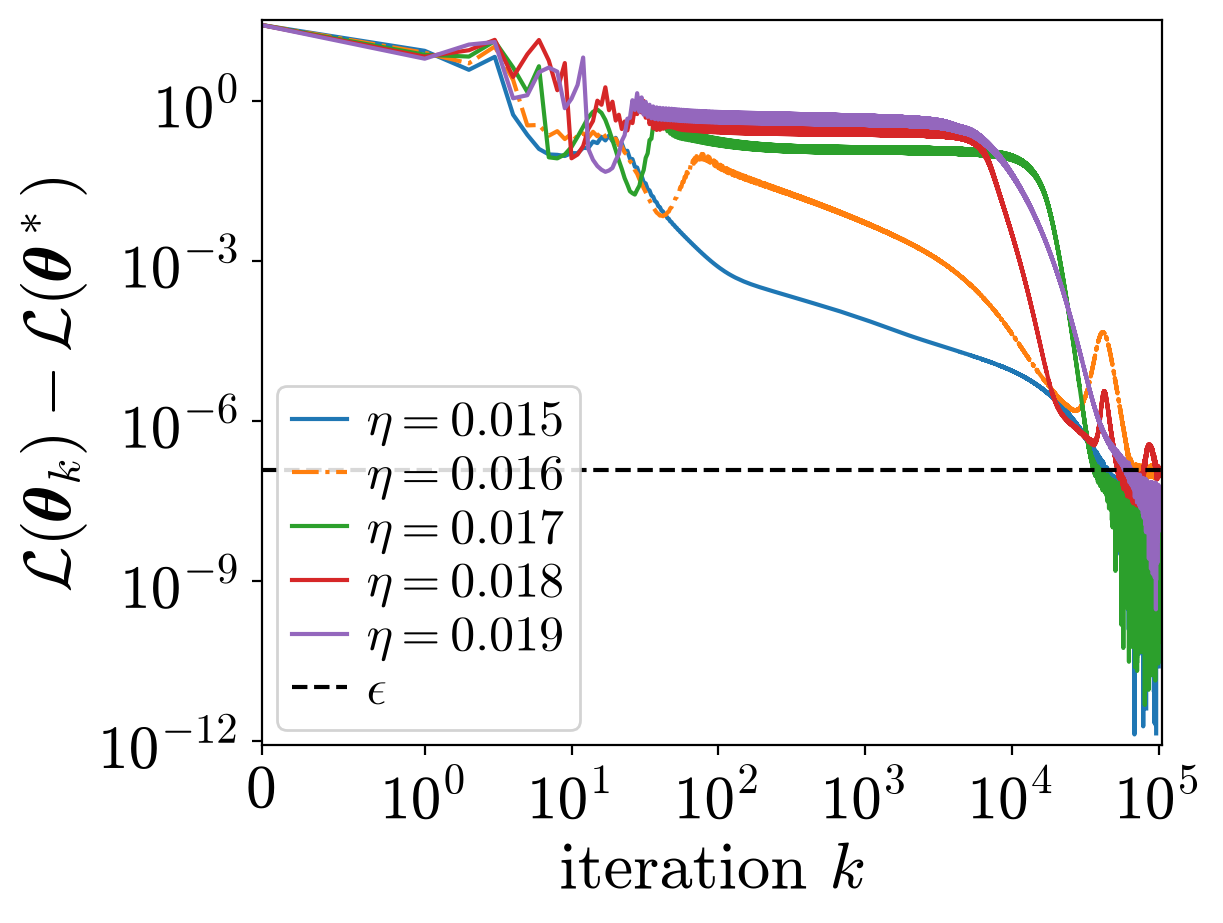}\hfill
        \includegraphics[width=0.333\linewidth]{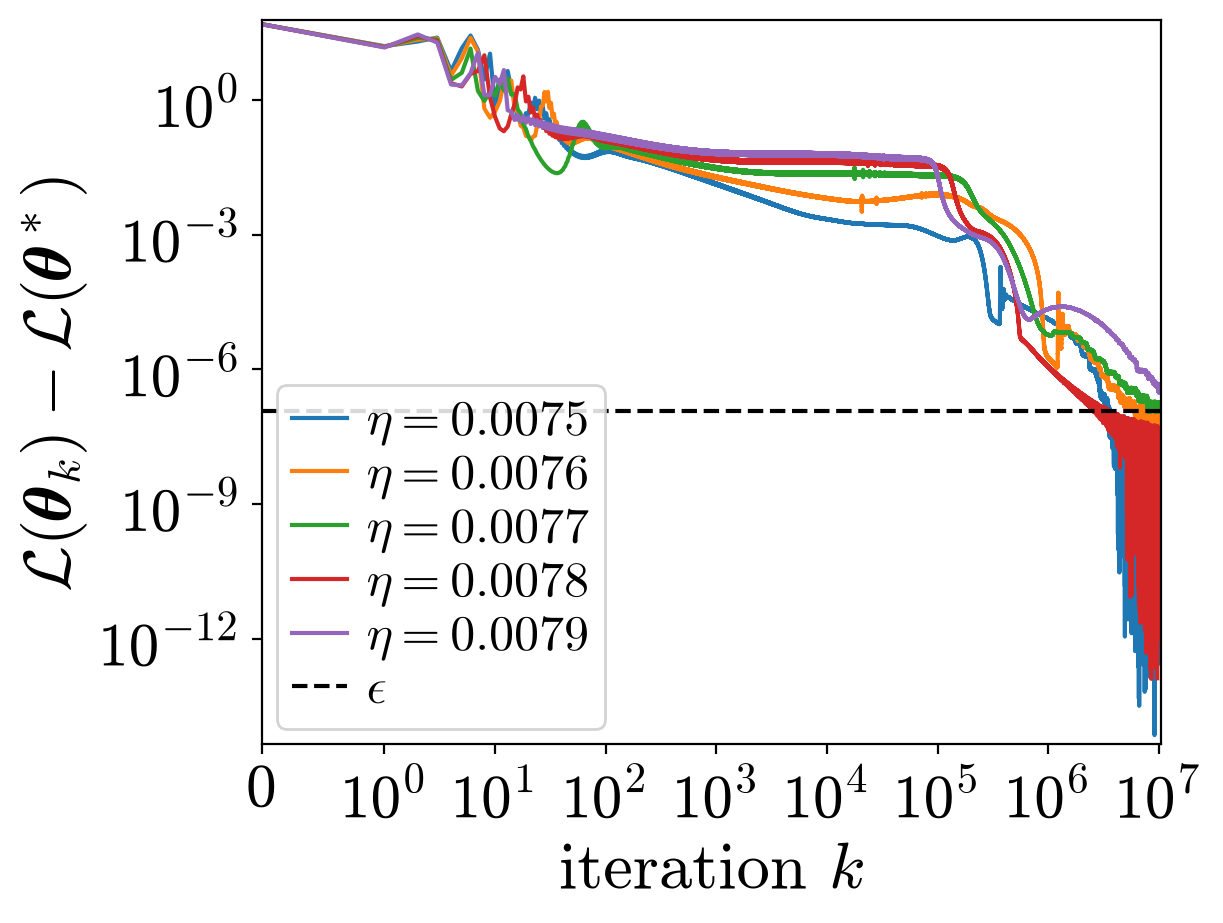}
        \caption{Training loss gap across iterations where $\hat{\mathcal{L}}(\boldsymbol{\theta}^*)$ is computed by \Cref{lemma:uniqueness}. $T \in \{4, 8, 16\}$ (left to right).}
        \label{fig:loss}
    \end{subfigure}

    \begin{subfigure}[t]{\textwidth}
        \centering
        \includegraphics[width=0.333\linewidth]{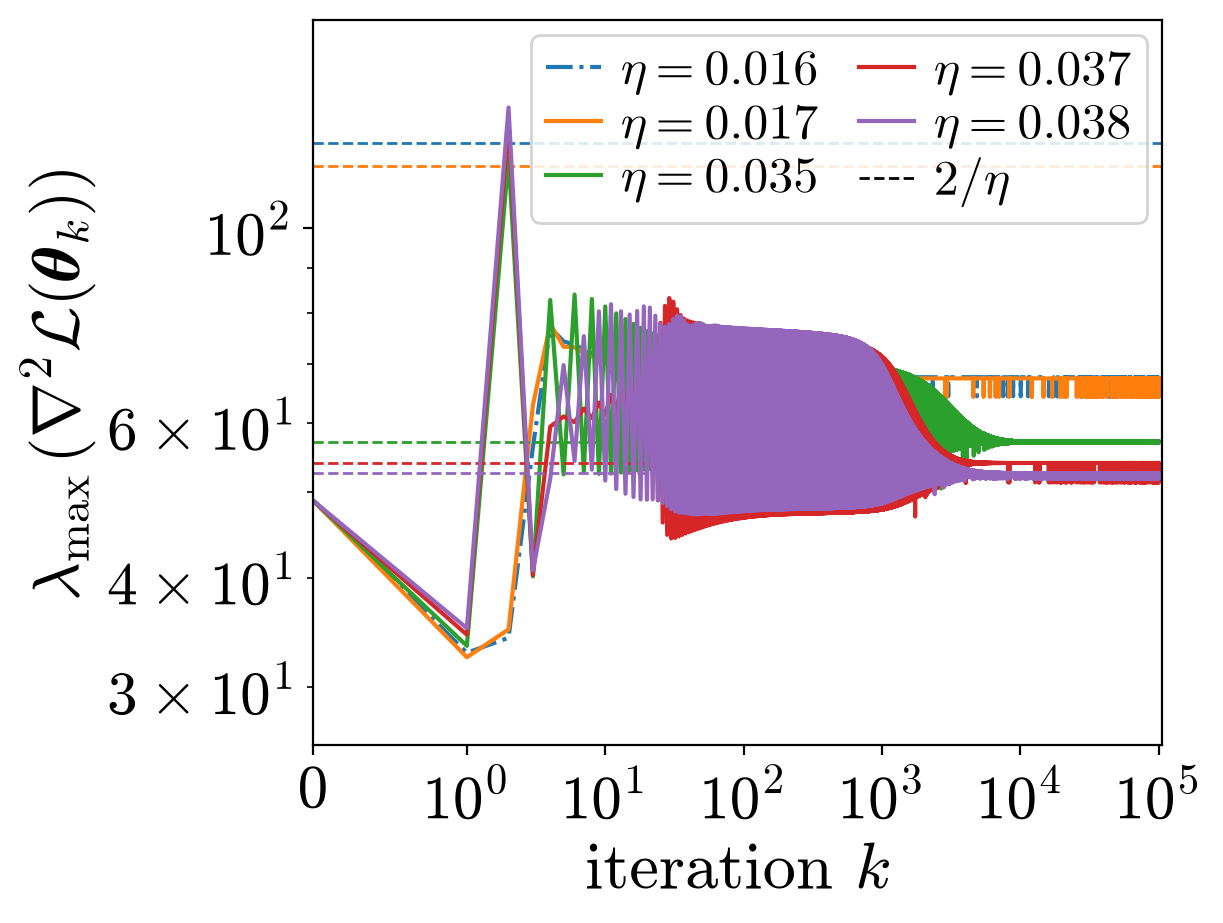}\hfill
        \includegraphics[width=0.333\linewidth]{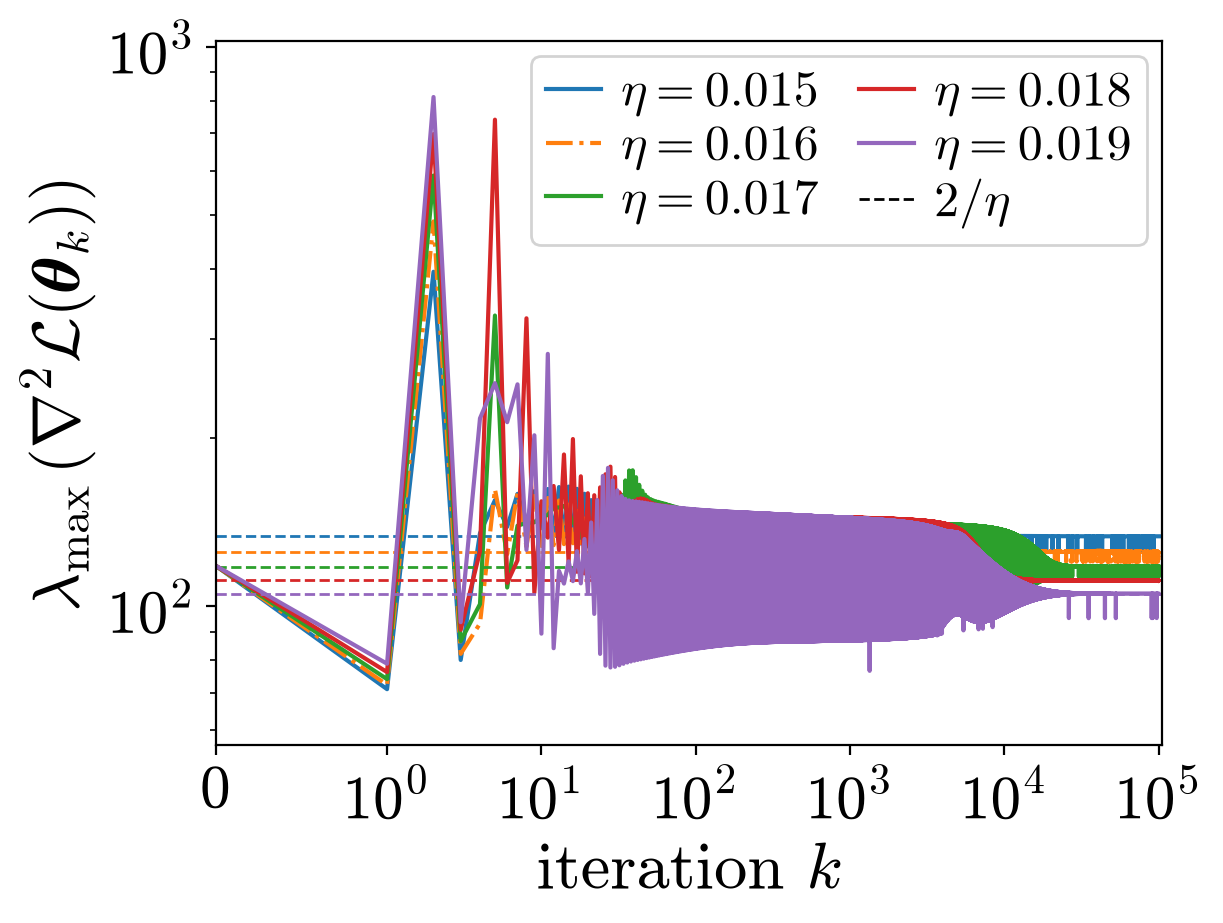}\hfill
        \includegraphics[width=0.333\linewidth]{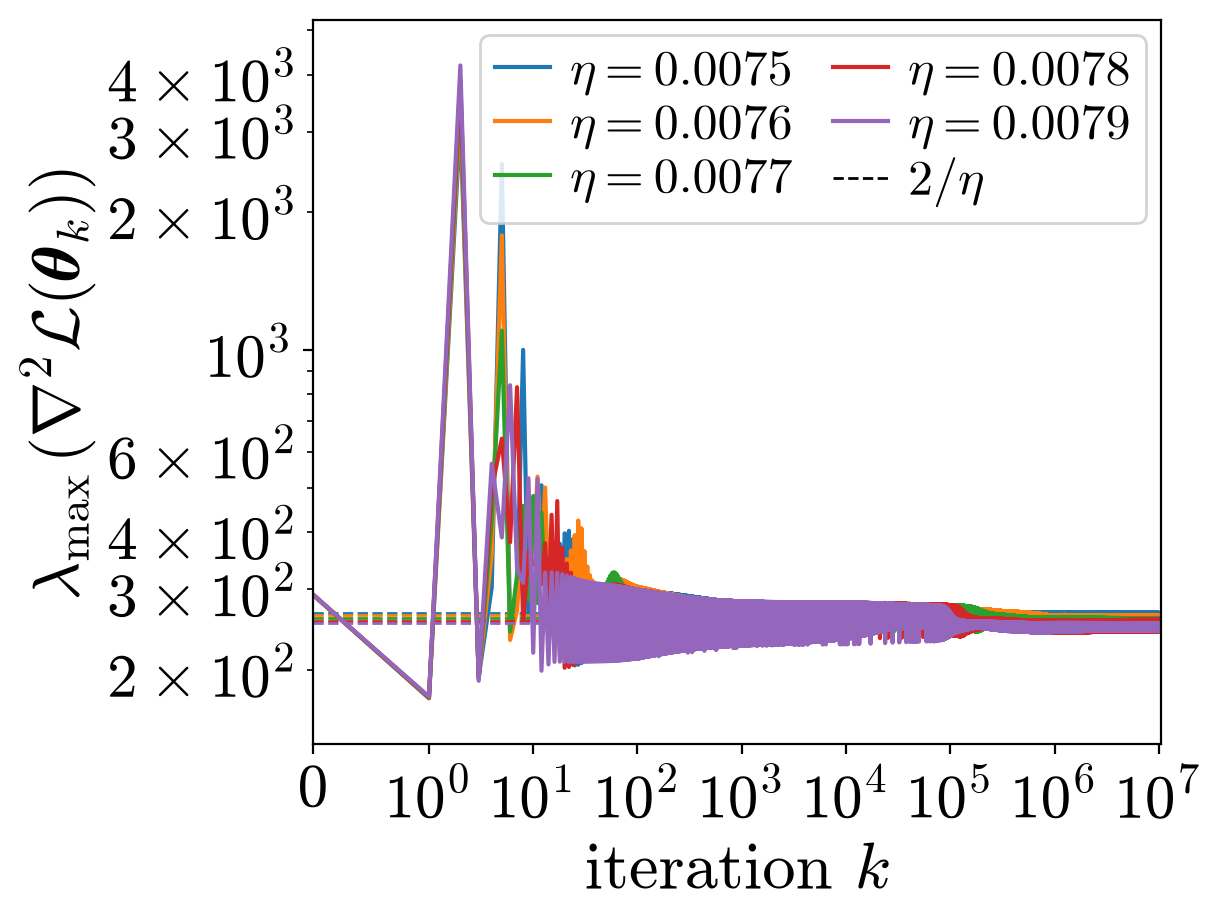}
        \caption{The maximum eigenvalue of the Hessian of the training loss across iterations. $T \in \{4, 8, 16\}$ (left to right).}
        \label{fig:sharpness}
    \end{subfigure}

    \caption{Training of the single-layer linear RNN with $d = 128$ at different learning rates (shown in different colors) for the training horizons $T \in \{4, 8, 16\}$. (a) The training loss gap $\hat{\mathcal{L}}(\boldsymbol{\theta}_k) - \hat{\mathcal{L}}(\boldsymbol{\theta}^*)$, where the optimal loss is computed by \Cref{lemma:uniqueness}. The dashed horizontal line marks the machine precision $\epsilon$. (b) The maximum eigenvalue of the Hessian of the training loss. The dashed horizontal lines mark $2/\eta$ for the learning rate shown in the same color.}
    \label{fig:training}
\end{figure*}

\begin{figure*}[!t]
    \centering
    \captionsetup[subfigure]{justification=centering,singlelinecheck=false}

    \begin{subfigure}[t]{\textwidth}
        \centering
        \includegraphics[width=0.333\linewidth]{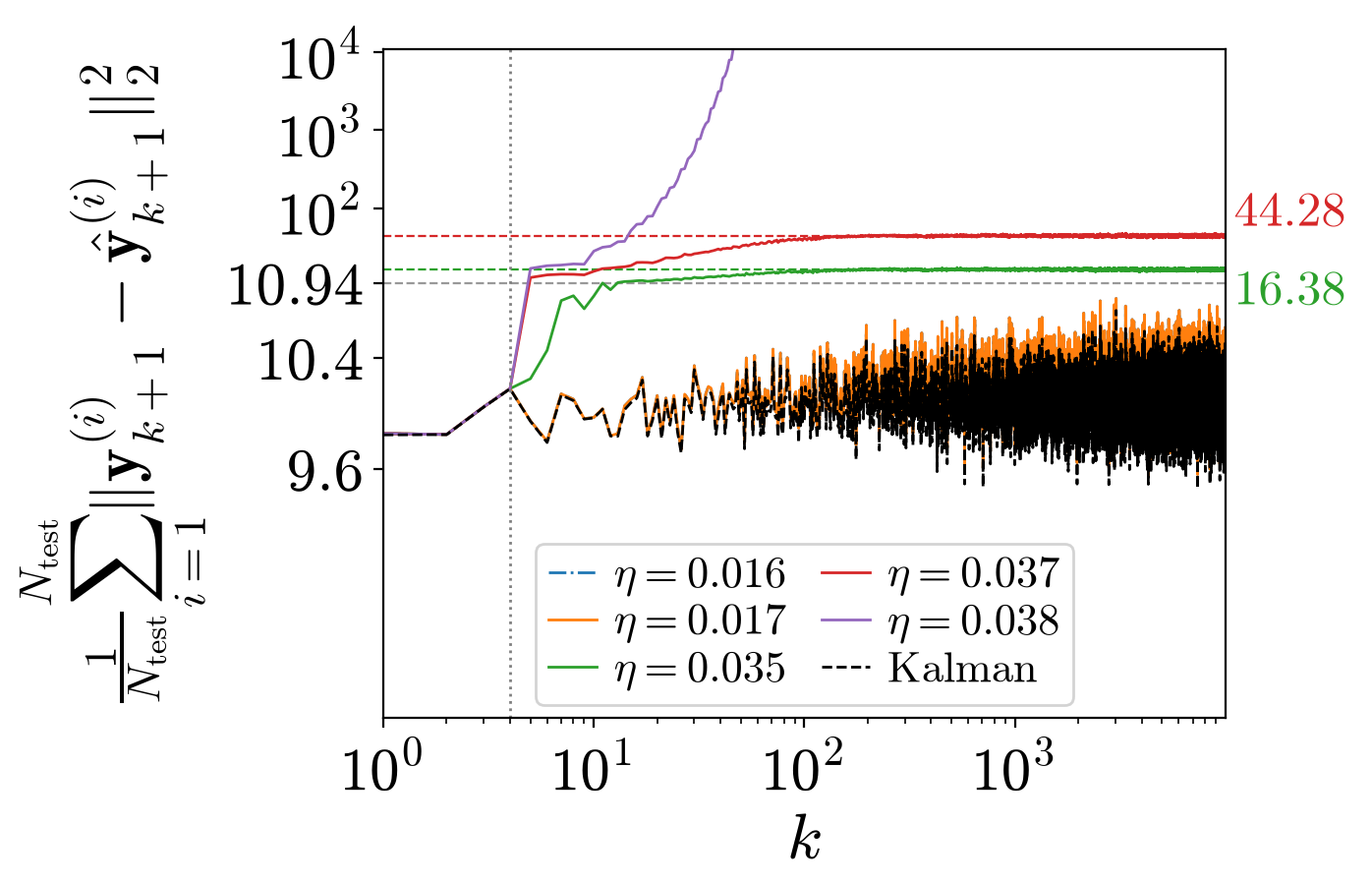}\hfill
        \includegraphics[width=0.333\linewidth]{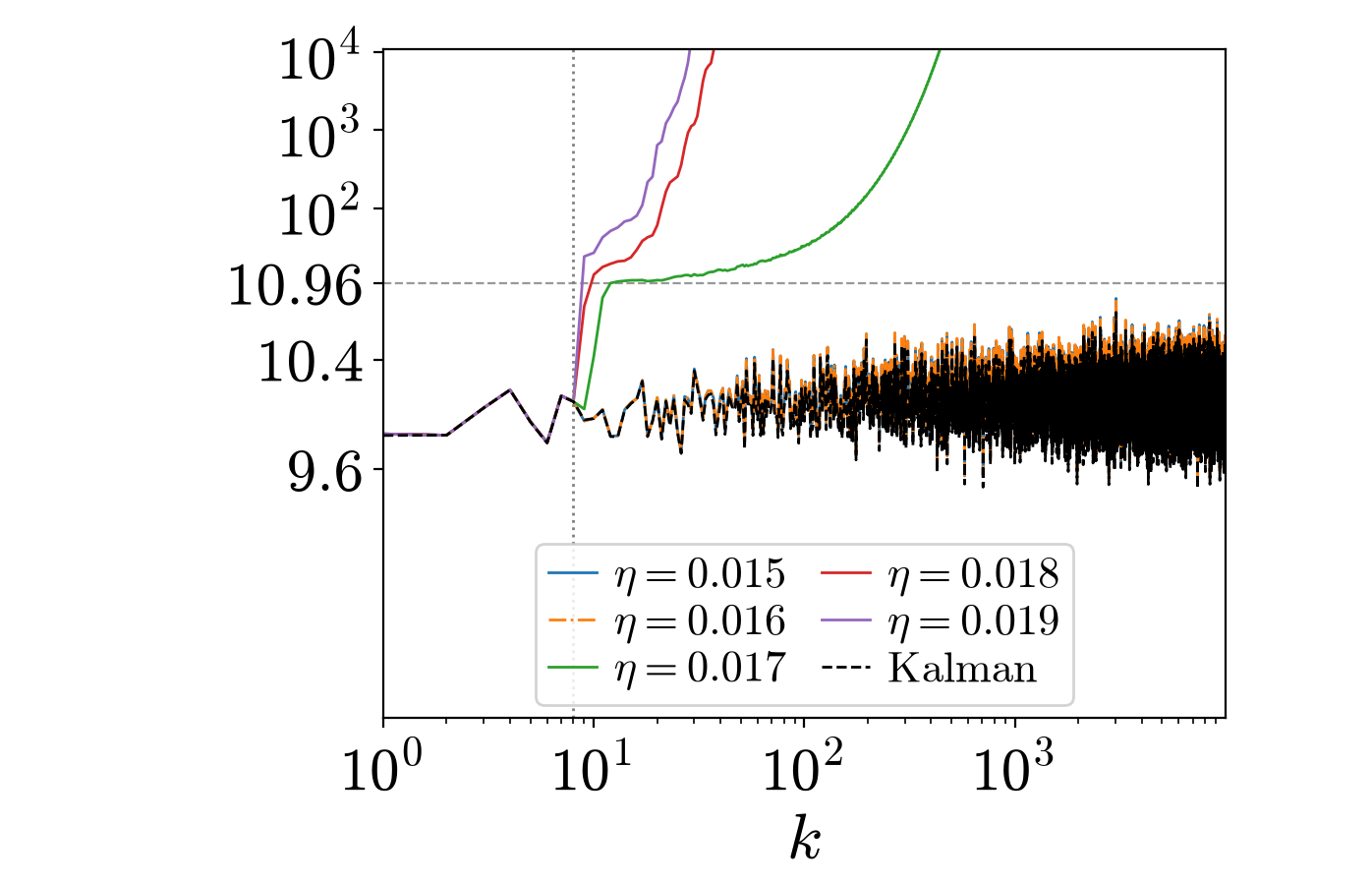}\hfill
        \includegraphics[width=0.333\linewidth]{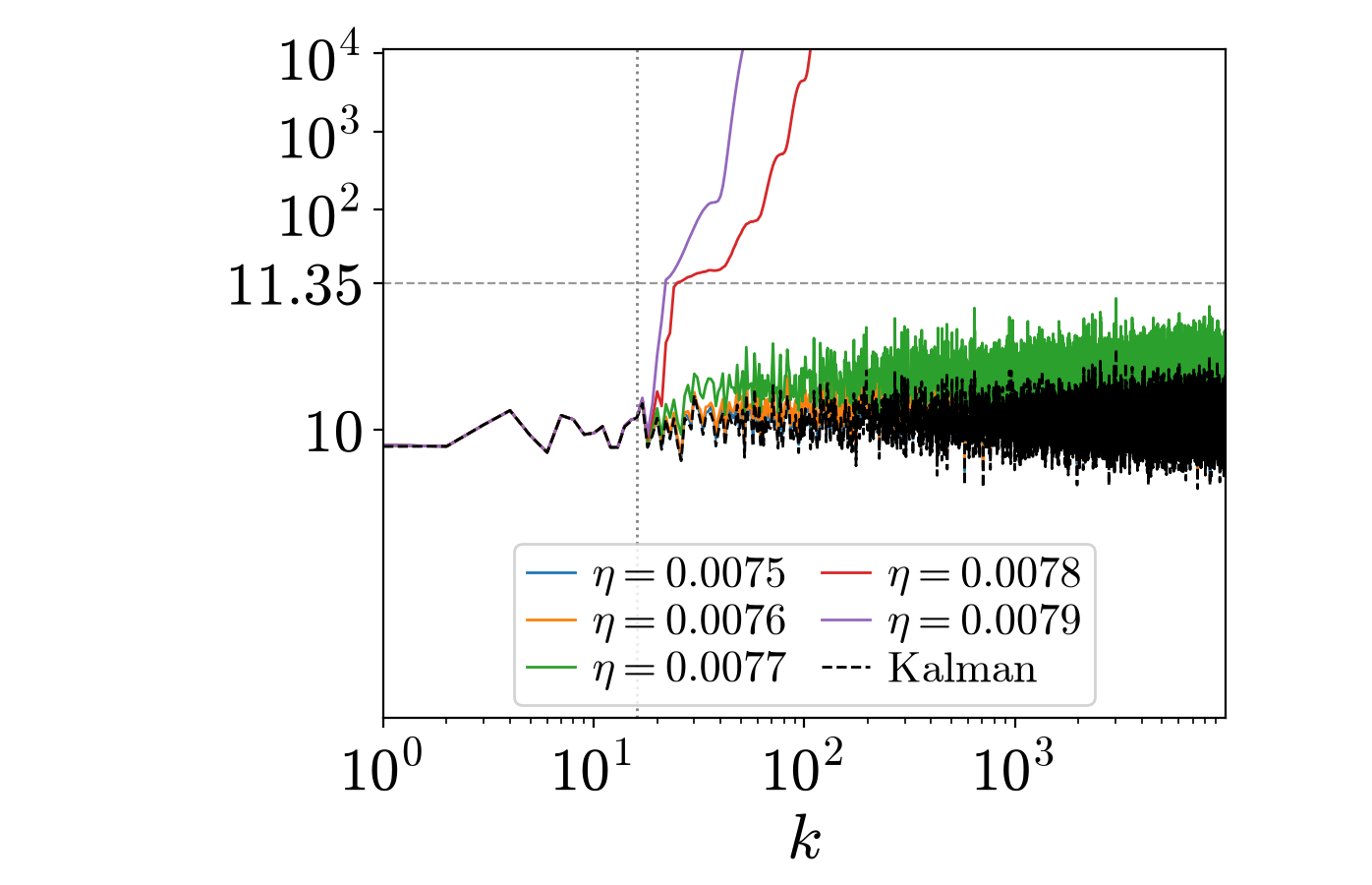}
        \caption{The per-step prediction error computed by $1/N_{\text{test}} \sum_{i=1}^{N_{\text{test}}} \|\hat{\mathbf{y}}^{(i)}_k - \mathbf{y}^{(i)}_k\|_2^2$. $T \in \{4, 8, 16\}$ (left to right).}
        \label{fig:test-error}
    \end{subfigure}

    \begin{subfigure}[t]{\textwidth}
        \centering
        \includegraphics[width=0.333\linewidth]{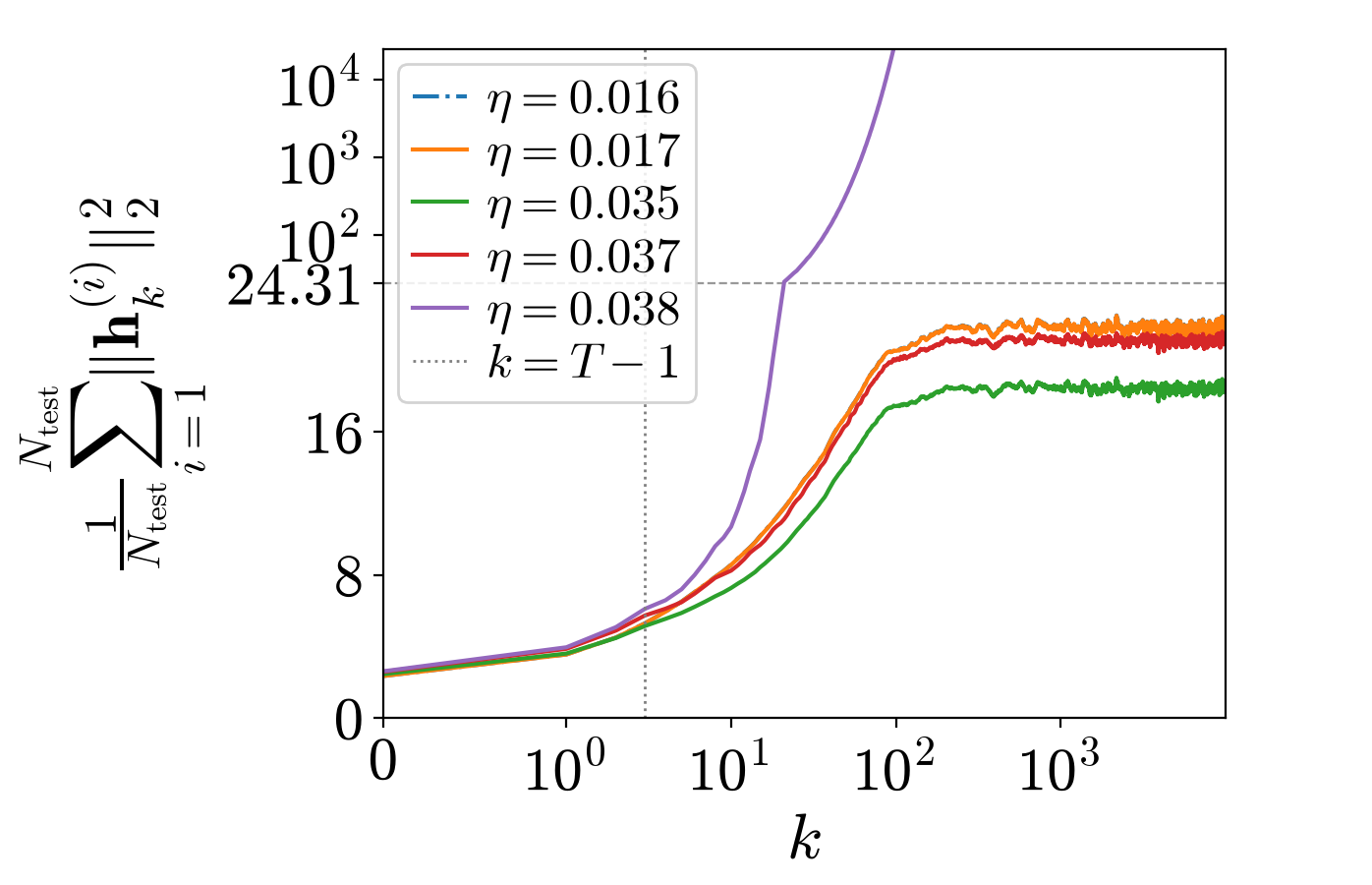}\hfill
        \includegraphics[width=0.333\linewidth]{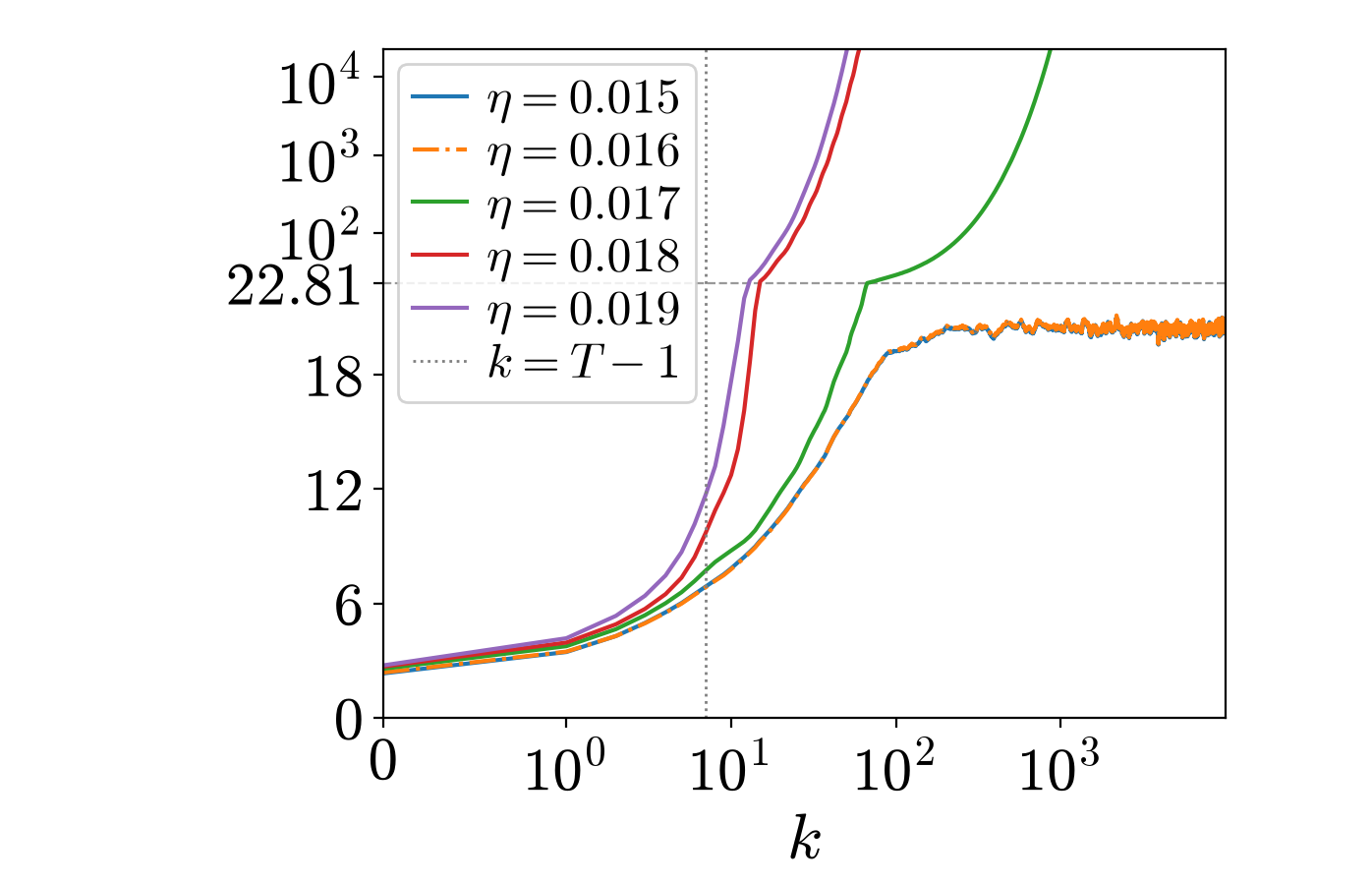}\hfill
        \includegraphics[width=0.333\linewidth]{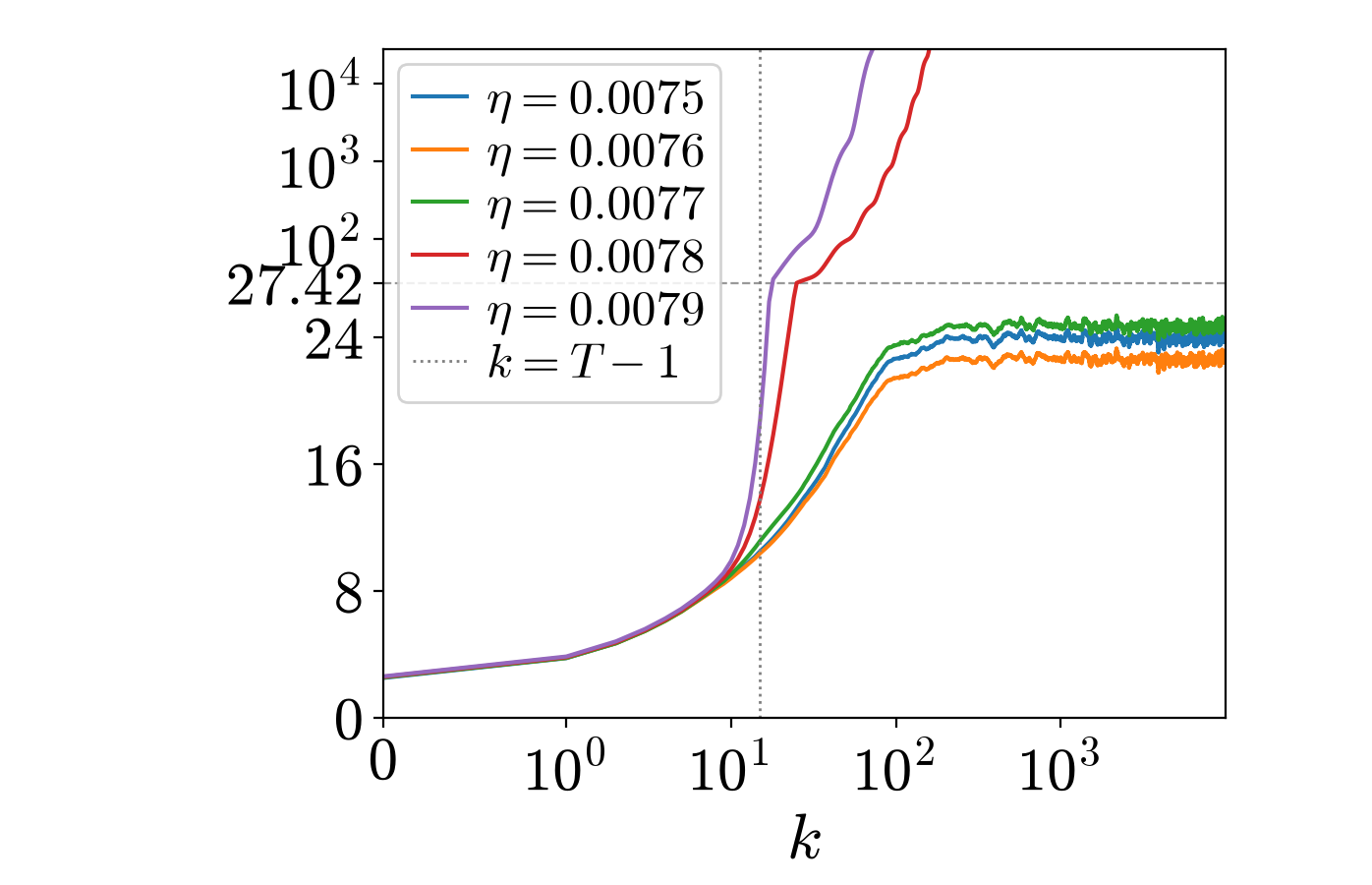}
        \caption{The average $\ell^2$-norm of the hidden state $\mathbf{h}_k$ during inference. $T \in \{4, 8, 16\}$ (left to right).}
        \label{fig:hidden}
    \end{subfigure}

    \begin{subfigure}[t]{\textwidth}
        \centering
        \includegraphics[width=0.333\linewidth]{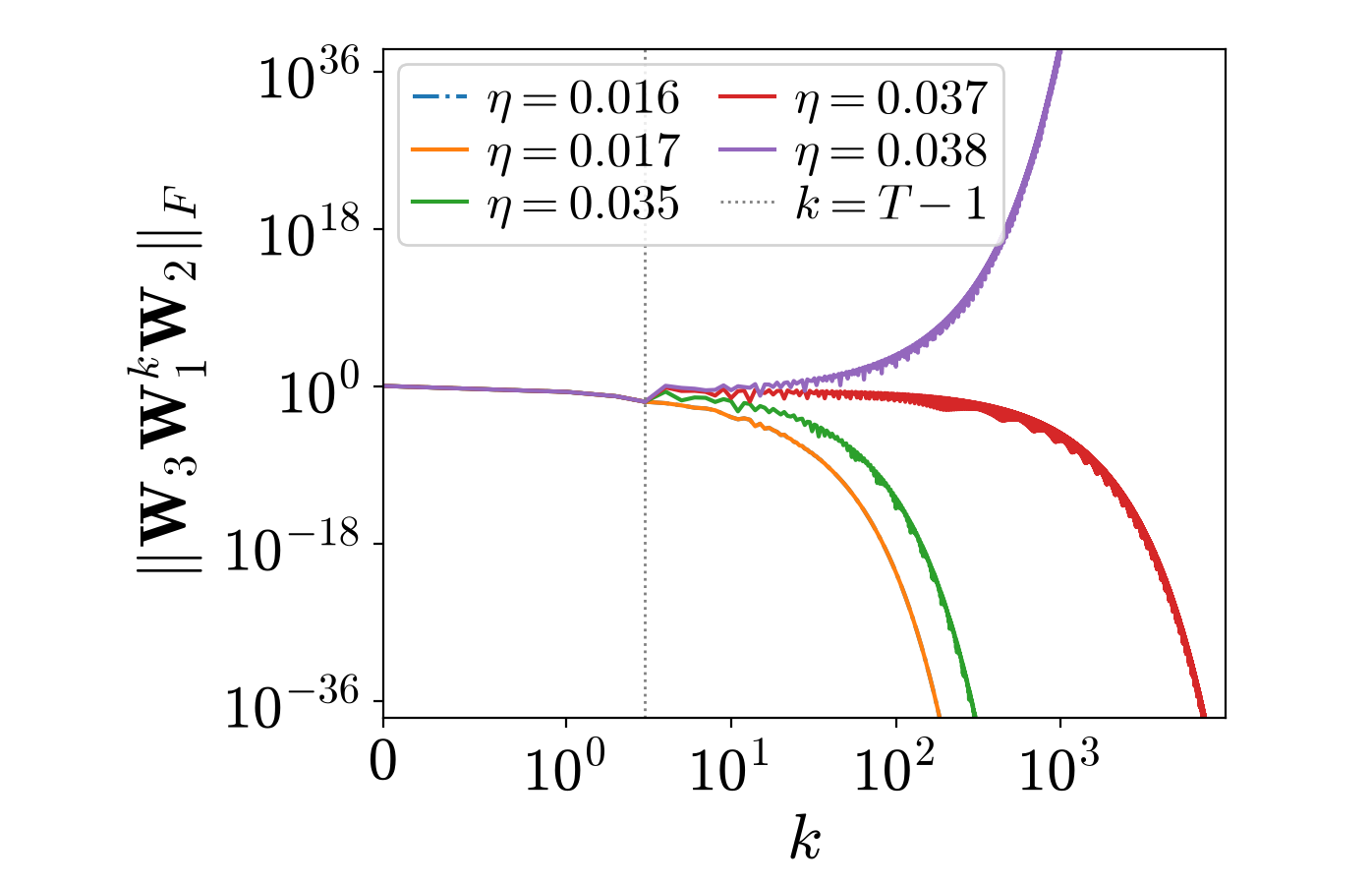}\hfill
        \includegraphics[width=0.333\linewidth]{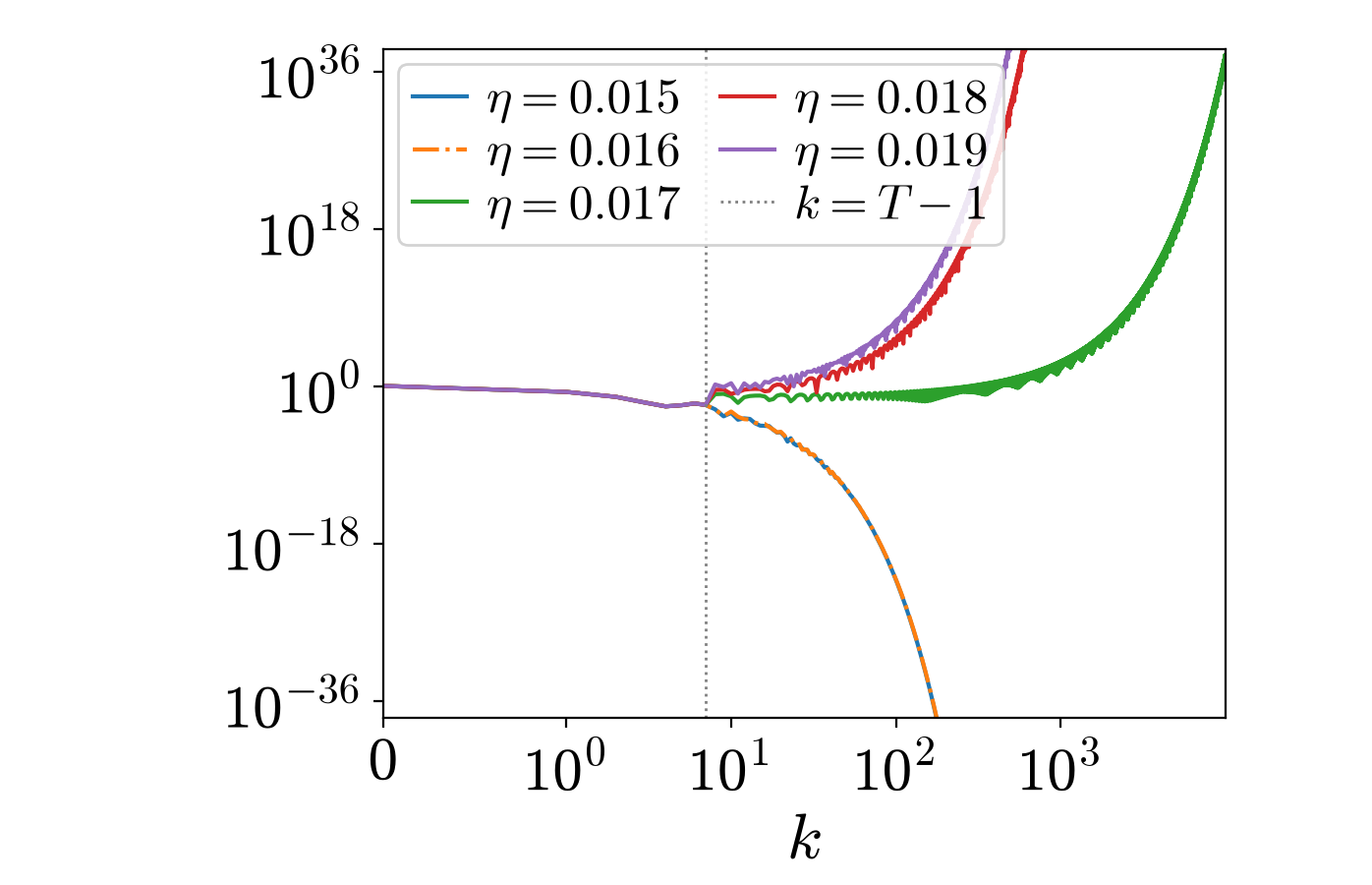}\hfill
        \includegraphics[width=0.333\linewidth]{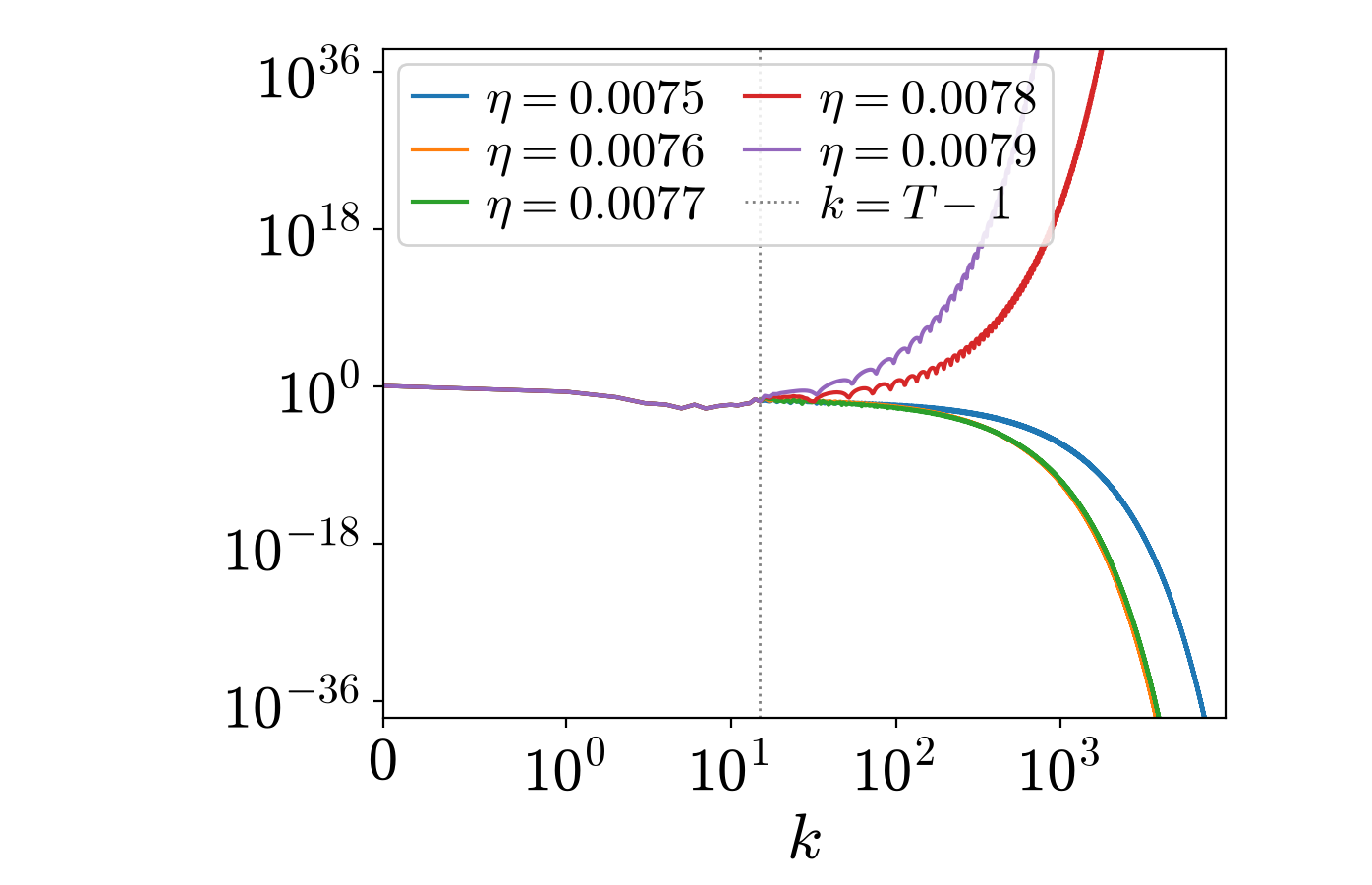}
        \caption{The Frobenius norm of the Markov parameter at time step $k$, i.e., $\|\mathbf{M}_k(\boldsymbol{\theta}_K)\|_F$, where $\boldsymbol{\theta}_K$ is the last iterate in \Cref{fig:training}. $T \in \{4, 8, 16\}$ (left to right).}
        \label{fig:markov}
    \end{subfigure}

    \caption{The performance of the linear RNNs trained in \Cref{fig:training} on test sequences that are longer than those seen during training. The dotted vertical line marks the training horizon $T$. (a) The dashed black curve is the Bayes-optimal error. The vertical axis is linear below the dashed horizontal line and logarithmic above it. The colored dashed horizontal lines mark the level at which the per-step prediction error settles.}
    \label{fig:test}
\end{figure*}

\begin{figure}[!t]
    \centering
    \captionsetup[subfigure]{justification=centering,singlelinecheck=false}

    \begin{subfigure}[t]{0.5\textwidth}
        \centering
        \includegraphics[width=0.90\linewidth]{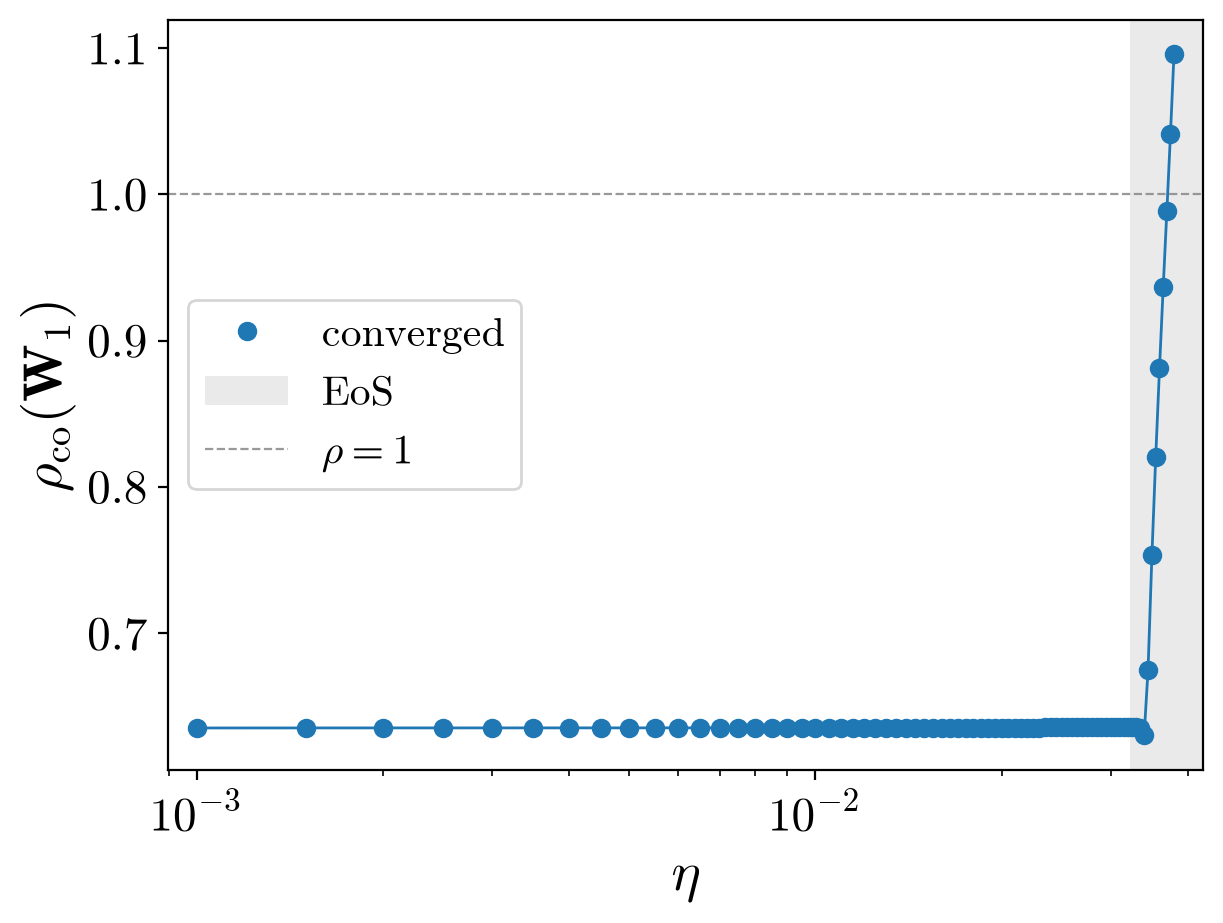}
        \caption{$T = 4$}
        \label{fig:radius-T4}
    \end{subfigure}\hfill
    \begin{subfigure}[t]{0.5\textwidth}
        \centering
        \includegraphics[width=0.90\linewidth]{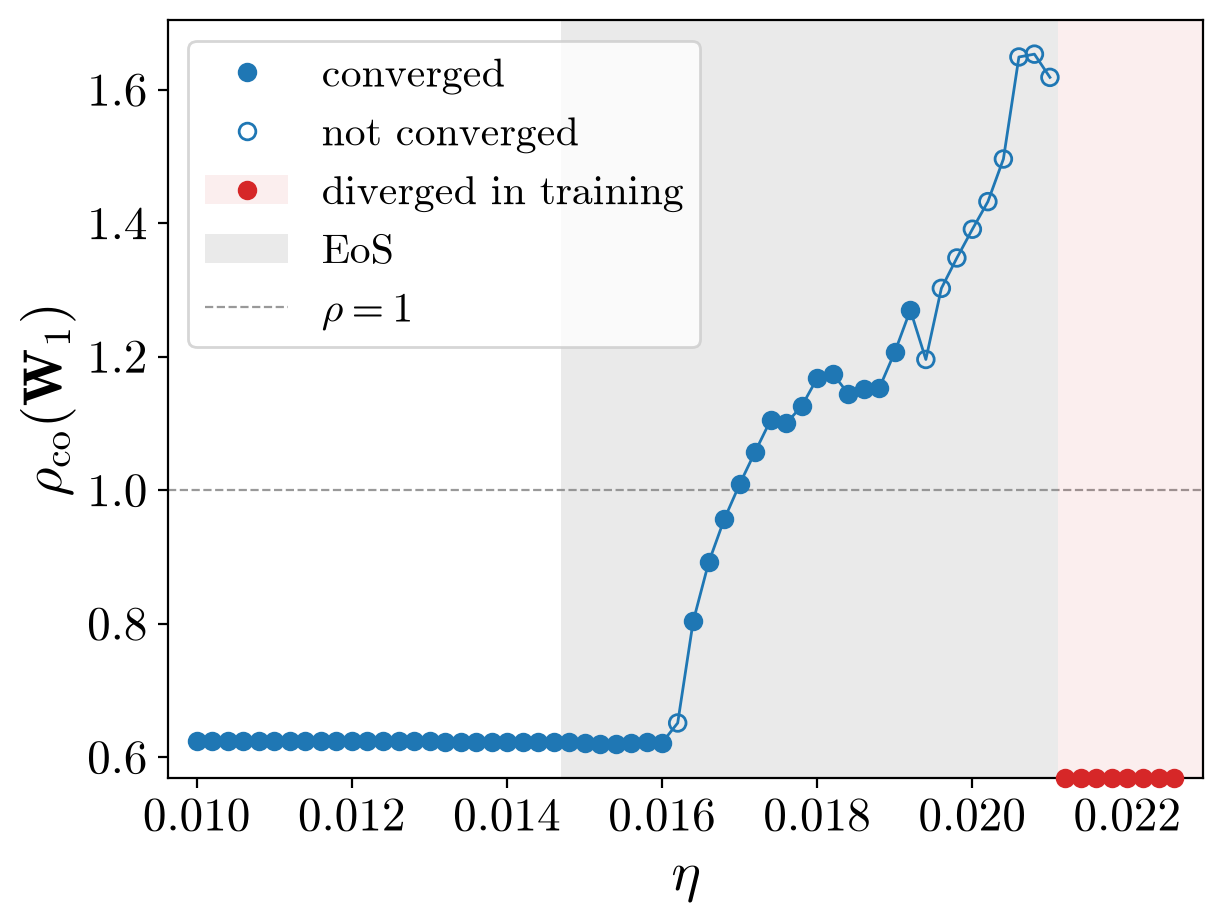}
        \caption{$T = 8$}
        \label{fig:radius-T8}
    \end{subfigure}

    \caption{The spectral radius restricted to controllable and observable modes of the recurrent weight matrix obtained from training, $\rho_{\text{co}}$, as a function of the learning rate $\eta$ for (a) $T = 4$ and (b) $T = 8$. Filled circles are the runs that converged to a global minimum, and hollow circles are the runs that did not converge in finite time. The gray region marks the learning rates whose runs end training at the edge of stability, and the red region marks the learning rates for which GD diverges.}
    \label{fig:radius}
\end{figure}

In this section, we empirically investigate how the learning rate determines which global minimum GD converges to and how these minima perform beyond the training horizon. Recall that, by \Cref{lemma:uniqueness}, we can compute the optimal training loss $\hat{\mathcal{L}}(\boldsymbol{\theta}^*)$ by solving the normal equation in \Cref{lemma:uniqueness} when $d \geq mT$. Therefore, we can train the linear RNN until its training loss gap $\hat{\mathcal{L}}(\boldsymbol{\theta}_k) - \hat{\mathcal{L}}(\boldsymbol{\theta}^*)$ reaches the machine precision $\epsilon$, where $\boldsymbol{\theta}_k$ denotes the $k$th iterate of GD. 

For the experiments in \Cref{fig:training,fig:radius,fig:test}, we generate $N_\text{tot} = 20480$ trajectories of length $10000$ i.i.d.\ from the linear dynamical system in \eqref{LDS} with $n = 4$ and $m = 2$, where the eigenvalues of $\mathbf{A}$ are $0.96$, $0.97$, $0.98$, and $0.99$. Moreover, the entries of $\mathbf{C}$ are drawn i.i.d.\ from $\mathcal{N}(0,1)$, $\mathbf{Q} = \mathbf{I}_n$, $\mathbf{R} = \mathbf{I}_m$, $\bar{\mathbf{x}}_0 = \mathbf{0}$, and $\mathbf{P}_0 = \mathbf{I}_n$. We use the first $T + 1$ observations of $N = 16384$ trajectories for training and the remaining $N_{\text{test}} = 4096$ trajectories for testing. We fix the hidden dimension to $d = 128$ and consider the training horizons $T \in \{4, 8, 16\}$. We initialize the entries of $\mathbf{W}_1$, $\mathbf{W}_2$, and $\mathbf{W}_3$ i.i.d.\ from the uniform distribution on $[-1/\sqrt{d}, 1/\sqrt{d}]$ and use the same initialization for every learning rate. We train the linear RNN using full-batch GD with a constant learning rate $\eta$ in single-precision arithmetic, whose machine precision is $\epsilon \approx 1.19 \times 10^{-7}$, until the training loss gap reaches $\epsilon$. Furthermore, we compute the maximum eigenvalue of the Hessian of the training loss  $\lambda_{\max}(\nabla^2\hat{\mathcal{L}}(\boldsymbol{\theta}_k))$ at every iteration through the Lanczos iteration \citep{lanczos1950iteration}.

As shown in \Cref{fig:sharpness}, we observe progressive sharpening during training. Following this regime, the training dynamics enter the edge of stability, i.e., the maximum Hessian eigenvalue hovers around $2/\eta$ while the training loss decreases non-monotonically. This demonstrates that even single-layer linear RNNs trained with the squared-error loss on a next-token prediction task operate at the edge of stability. To the best of our knowledge, this has not been observed before.

During inference, we run the trained linear RNNs as one-step-ahead predictors rather than autoregressively, which eliminates the exposure bias that can be catastrophic for a next-token prediction task on a continuous vocabulary. We test them on sequences of length $10^4$. As in training, the initial hidden state is set to zero. 

To investigate the length-generalization capability of the trained models, we compare the per-step prediction error of the linear RNNs with that of the Kalman predictor, which is the Bayes-optimal one-step-ahead predictor in the MMSE sense for the underlying linear dynamical system in \eqref{LDS}. Furthermore, we define length-generalization failure as the case in which the per-step prediction error either settles at a higher value than the Bayes-optimal error or diverges. When we compare two global minima in terms of their length-generalization capabilities, we use the following conceptual binary relation: For two global minima whose per-step prediction errors settle to finite values, if the per-step prediction error of one settles to a lower level than that of the other, we say that the first length generalizes better than the second. Furthermore, for two global minima whose per-step prediction errors diverge, if the per-step prediction error of one grows faster than that of the other, we say that the latter length generalizes better than the former even though they are both unstable systems.

Under this relation, we observe that sufficiently large learning rates induce edge-of-stability dynamics, and the resulting global minima exhibit worse length generalization than those reached at smaller learning rates, for which the dynamics remain outside this regime. To determine whether a run is at the edge of stability, we check the median of the maximum Hessian eigenvalue over the last $10\%$ of training. If it is at least $0.99 \cdot 2/\eta$, we consider the run to be at the edge of stability. As shown in \Cref{fig:test-error}, the runs with $\eta = 0.016$ and $\eta = 0.017$, which remain outside the edge-of-stability regime, perform close to the Kalman predictor. In contrast, every run that enters the edge-of-stability regime fails to generalize beyond the training horizon. Note that the runs at the edge of stability converge to relatively flat minima as shown in \Cref{fig:training}. Given that length generalization is a special case of the notion of generalization, this means a relatively flat minimum need not generalize better than a sharper one.

We observe that for the runs whose per-step prediction error diverges, the Markov parameters also diverge exponentially, and the rate of divergence is higher for larger learning rates. To understand this behavior, we investigate the effect of the learning rate on the controllable and observable modes of the recurrent weight matrix $\mathbf{W}_1$ at the converged global minimum, as shown in \Cref{fig:radius}. We observe that among the runs that enter the edge of stability, the larger learning rates favor global minima whose recurrent weight matrices have a larger spectral radius restricted to the controllable and observable modes. To the best of our knowledge, this implicit bias mechanism has not been characterized in the literature for linear recurrent models.

\section{Discussion}
\label{sec:Discussion}

\paragraph{Sharpness and Spectrum of $\mathbf{W}_1$.} We observe in \Cref{fig:radius} that, among the runs that enter the edge-of-stability regime, larger learning rates favor global minima whose recurrent weight matrices have a larger $\rho_{\text{co}}$. Since the maximum Hessian eigenvalue at the end of these runs hovers around $2/\eta$ as shown in \Cref{fig:sharpness}, larger learning rates also converge to relatively flat minima. To understand the relationship between the sharpness of a global minimum and $\rho_{\text{co}}$, we derive a closed-form expression for the maximum Hessian eigenvalue of the training loss at any global minimum.
\begin{theorem}
\label{thm:exact-sharpness}
Consider the single-layer linear RNN in \eqref{RNN} with $d \geq mT$, and define the sample autocorrelation matrix of the zero-padded history by
$\widehat{\boldsymbol{\Sigma}}_T := \frac{1}{NT}\mathbf{Z}\mathbf{Z}^\top$. Then, for any $\boldsymbol{\theta}^* \in \Omega$, we have
\begin{equation}
    \lambda_{\max}(\nabla^2\hat{\mathcal{L}}(\boldsymbol{\theta}^{*})) =
2
\sigma_{\max}^2\left (
    \left(
        \widehat{\boldsymbol{\Sigma}}_T^{1/2}
        \otimes
        \mathbf{I}_m
    \right)
    \mathbf{K}^*
\right),
\end{equation}
where $\mathbf{K}^*$ is defined by
\begin{equation}
    \mathbf{K}^*
:=
\frac{
    \partial\operatorname{vec}
    \left(
        \mathbf{\Phi}_T(\boldsymbol{\theta})
    \right)
}{
    \partial\operatorname{vec}(\boldsymbol{\theta})
}\Bigg |_{\boldsymbol{\theta}= \boldsymbol{\theta}^*}
=
\begin{bmatrix}
    \mathbf{K}^*_{1,0} & \mathbf{K}^*_{2,0} & \mathbf{K}^*_{3,0}\\
    \mathbf{K}^*_{1,1} & \mathbf{K}^*_{2,1} & \mathbf{K}^*_{3,1}\\
    \vdots & \vdots & \vdots\\
    \mathbf{K}^*_{1,T-1} & \mathbf{K}^*_{2,T-1} & \mathbf{K}^*_{3,T-1}
\end{bmatrix},
\end{equation}
with
\begin{align}
    \mathbf{K}^*_{1,k} &= \sum_{\ell=0}^{k-1}
    \left[
        \left(
            \mathbf{W}_1^{*\,k-1-\ell}\mathbf{W}_2^*
        \right)^{\top}
        \otimes
        \mathbf{W}_3^*\mathbf{W}_1^{*\,\ell}
    \right], \\
    \mathbf{K}^*_{2,k} &=
    \mathbf{I}_m\otimes\mathbf{W}_3^*\mathbf{W}_1^{*\,k},
    \qquad
    \mathbf{K}^*_{3,k} = \left(
        \mathbf{W}_1^{*\,k}\mathbf{W}_2^*
    \right)^{\top}
    \otimes\mathbf{I}_m
\end{align}
for all $k \in [T-1]_0$, and $\mathbf{K}^*_{1,0} = \mathbf{0}$.

\end{theorem}
The proof relies on an explicit expression for the second directional derivative of $\hat{\mathcal{L}}$ that simplifies at an arbitrary global minimizer due to the surjectivity of $\boldsymbol{\theta} \mapsto \mathbf{\Phi}_{T}(\boldsymbol{\theta})$ and the convexity of $\hat{\mathcal{L}}$ with respect to $\mathbf{\Phi}_T(\boldsymbol{\theta})$. We defer the relevant technical preliminaries to \Cref{appendix:dir_derivatives} and the full proof to \Cref{appendix:proof-exact-sharpness}. Our analysis is inspired by \citet{mulayoff, kamber2025sharpness, kamber2026loss}, and \citet{xu2026doessgdseekflatness} who study the worst-case sharpness of the loss landscape in the vicinity of a minimizer in deep linear models.

Note that, by \Cref{thm:exact-sharpness}, the sharpness of a global minimum depends on its parameterization through $\mathbf{K}^*$, which is a function over the set of global minima. Similar to the observation of \citet{sharpminimacangeneralize} that the sharpness of a deep neural network can be changed arbitrarily by a reparameterization that does not change the end-to-end function it represents, the following result shows that the global minima with the same recurrent weight matrices can be arbitrarily sharp.

\begin{proposition}
\label[proposition]{prop:rescaling}
Consider the single-layer linear RNN given in \eqref{RNN} with $d \geq mT$. Suppose that $\widehat{\boldsymbol{\Sigma}}_T$ has full rank. For every $\boldsymbol{\theta}^* = (\mathbf{W}_1^*, \mathbf{W}_2^*, \mathbf{W}_3^*) \in \Omega$ and $c \in \mathbb{R}\setminus\{0\}$, define $\boldsymbol{\theta}^*_c := (\mathbf{W}_1^*, c\mathbf{W}_2^*, c^{-1}\mathbf{W}_3^*)$. Then, $\boldsymbol{\theta}^*_c \in \Omega$, $\mathbf{M}_i(\boldsymbol{\theta}^*_c) = \mathbf{M}_i(\boldsymbol{\theta}^*)$ for every $i \geq 0$, $\rho_{\text{co}}(\boldsymbol{\theta}^*_c) = \rho_{\text{co}}(\boldsymbol{\theta}^*)$, and
\begin{equation}
    \lambda_{\max}(\nabla^2\hat{\mathcal{L}}(\boldsymbol{\theta}^*_c))
    \geq
    2\lambda_{\min}(\widehat{\boldsymbol{\Sigma}}_T)
    \max\left\{
        c^2\sigma_{\max}^2(\mathcal{C}_T(\boldsymbol{\theta}^*)),\,
        c^{-2}\sigma_{\max}^2(\mathcal{O}_T(\boldsymbol{\theta}^*))
    \right\}.
    \label{eq:rescaling-lower-bound}
\end{equation}
\end{proposition}
We defer the proof to \Cref{appendix:proof-rescaling}. This means that when $\mathbf{\Phi}^*  \neq \mathbf{0}$, the right-hand side of \eqref{eq:rescaling-lower-bound} tends to infinity as $c \to \infty$. Therefore, a condition on $\rho_{\text{co}}$ alone cannot imply that a global minimum is flat. Furthermore, as we show in the following proposition, $\rho_{\text{co}}$ is unbounded on $\Omega$.

\begin{proposition}
\label[proposition]{prop:unbounded-rho}
Consider the single-layer linear RNN given in \eqref{RNN} with $d \geq mT$. Suppose that $\widehat{\boldsymbol{\Sigma}}_T$ has full rank and $\mathbf{\Phi}^* \neq \mathbf{0}$. Then, for every $\lambda \in \mathbb{R}\setminus\{0\}$, there exists $\boldsymbol{\theta}^* \in \Omega$ such that $\rho_{\text{co}}(\boldsymbol{\theta}^*) = |\lambda|$.
\end{proposition}
We defer the proof to \Cref{appendix:proof-unbounded-rho}. This means that $\sup_{\boldsymbol{\theta} \in \Omega}\rho_{\text{co}}(\boldsymbol{\theta}) = \infty$. Therefore, we investigate the flat minima that attain a given value of $\rho_{\text{co}}$. For every $r \in \rho_{\text{co}}(\Omega) := \{\rho_{\text{co}}(\boldsymbol{\theta}^*) : \boldsymbol{\theta}^* \in \Omega\}$, define $s: \rho_{\text{co}}(\Omega) \to \mathbb{R}_+$ by
\begin{equation}
    s(r) := \inf\left\{
        \lambda_{\max}(\nabla^2\hat{\mathcal{L}}(\boldsymbol{\theta}^*)) :
        \boldsymbol{\theta}^* \in \Omega,\ \rho_{\text{co}}(\boldsymbol{\theta}^*) = r
    \right\}.
    \label{eq:min-sharpness}
\end{equation}
We showed in \Cref{prop:rescaling} that the supremum of the same set is infinite for every $r \in \rho_{\text{co}}(\Omega)$. Through the lens of dynamical stability, we hypothesize the following.
\begin{hypothesis}
\label{hyp:flat}
Suppose that $d \geq mT$, $T \geq 2$, and $\widehat{\boldsymbol{\Sigma}}_T$ has full rank. Then, $s$ is non-increasing and non-constant on $\rho_{\text{co}}(\Omega)$.
\end{hypothesis}

\Cref{hyp:flat} relates $\rho_{\text{co}}(\boldsymbol{\theta}^*)$ to the sharpness $\lambda_{\max}(\nabla^2\hat{\mathcal{L}}(\boldsymbol{\theta}^*))$ and, in turn, to the learning rate as follows: A global minimum $\boldsymbol{\theta}^* \in \Omega$ is \emph{linearly stable} for GD with learning rate $\eta$ if and only if $\lambda_{\max}(\nabla^2\hat{\mathcal{L}}(\boldsymbol{\theta}^*)) \leq 2/\eta$ \citep{Wustab}. Let us define for every $\eta > 0$,
\begin{equation}
    \mathcal{A}_\eta := \left\{r \in \rho_{\text{co}}(\Omega) : s(r) \leq 2/\eta\right\}
    \quad \text{and} \quad
    r_\eta := \inf \mathcal{A}_\eta,
\end{equation}
where $\inf \emptyset := \infty$. If $\boldsymbol{\theta}^*$ is linearly stable for GD with learning rate $\eta$, then $s(\rho_{\text{co}}(\boldsymbol{\theta}^*)) \leq \lambda_{\max}(\nabla^2\hat{\mathcal{L}}(\boldsymbol{\theta}^*)) \leq 2/\eta$ by \eqref{eq:min-sharpness}. Hence, $\rho_{\text{co}}(\boldsymbol{\theta}^*) \in \mathcal{A}_\eta$, and thus $\rho_{\text{co}}(\boldsymbol{\theta}^*) \geq r_\eta$. Moreover, if $\eta_1 < \eta_2$, then $\mathcal{A}_{\eta_2} \subseteq \mathcal{A}_{\eta_1}$; therefore, $r_{\eta_1} \leq r_{\eta_2}$. Under \Cref{hyp:flat}, if $r \in \mathcal{A}_\eta$, then $s(r') \leq s(r) \leq 2/\eta$ for every $r' \in \rho_{\text{co}}(\Omega)$ with $r' \geq r$. Hence, $\mathcal{A}_\eta$ contains every $r \in \rho_{\text{co}}(\Omega)$ with $r > r_\eta$, i.e., linear stability excludes only the values of $\rho_{\text{co}}$ below $r_\eta$, and this threshold does not decrease as the learning rate increases. Furthermore, since $s$ is non-constant, there exist $r_1, r_2 \in \rho_{\text{co}}(\Omega)$ with $r_1 < r_2$ and $s(r_1) > s(r_2)$. Then, for every $\eta$ with $s(r_2) \leq 2/\eta < s(r_1)$, every $r \in \rho_{\text{co}}(\Omega)$ with $r \leq r_1$ is excluded, i.e., the threshold is nontrivial. This agrees with the phenomenon we observe in \Cref{fig:radius}. We leave open the question of whether $s$ is non-increasing and non-constant over $\rho_{\text{co}}(\Omega)$.

\paragraph{Unexplored State Hypothesis.}\citet{ruiz2025understanding} argued that recurrent models fail to length generalize when applied to long sequences whose hidden-state distributions were not encountered during training. As we observe empirically in \Cref{fig:test-error}, the implicit bias of GD toward certain global minima can also lead to length-generalization failure when the models are trained within the same training horizon. Therefore, to understand length generalization, we need to understand the optimization dynamics of the gradient-based algorithms. 

\section{Conclusion}
In this paper, we empirically demonstrated that single-layer linear RNNs trained with squared-error loss on the next-token prediction task operate at the edge of stability following a progressive sharpening regime. Our experiments demonstrated that sufficiently large learning rates induce the edge-of-stability dynamics, and the resulting global minima exhibit worse performance than those reached at smaller learning rates, which did not enter the edge of stability. This implies that a relatively flat minimum need not generalize better than a sharper one. Furthermore, we demonstrated that among the runs that enter the edge-of-stability regime, larger learning rates further degrade length generalization. We empirically characterized an implicit bias of GD in the training of linear RNNs: Larger learning rates favor global minima whose recurrent weight matrices have a larger spectral radius restricted to the controllable and observable modes. To understand this implicit bias, we provided a closed-form expression for the maximum Hessian eigenvalue of the training loss at any global minimizer, and we showed that the global minima with the same $\rho_{\text{co}}$ can be arbitrarily sharp and that this spectral radius is unbounded on the set of global minima. Therefore, through the lens of dynamical stability, we hypothesized that the sharpness of the flat minima among the global minima with a given $\rho_{\text{co}}$ is a non-increasing and non-constant function of this spectral radius. We leave this as an open question.
\label{sec:Conclusion}

\section*{Acknowledgments}
The experiment shown in \Cref{fig:mamba} was designed and executed by Claude Opus 5.5. The parameterization regime $d \ge mT$, in which $\mathbf{\Phi}_T(\cdot)$ is surjective, was proposed by Claude Fable 5.1. \Cref{thm:arbitrary-charpoly} and \Cref{prop:unbounded-rho} were formulated and proved by Claude Opus 5.5. The characterization of $\mathbf{K}^*$ via a multivariate Taylor expansion and the Jacobian in the proof of \Cref{thm:exact-sharpness} was suggested by the authors, and Claude Opus 5.5 provided a complete proof. Furthermore, \Cref{hyp:flat} was formalized with the assistance of Claude Opus 5.5. The authors take full responsibility for the correctness of all results. Beyond this, GPT-5.6 and Claude Opus 5.5 were used as spell checkers, grammar-suggestion tools, editing aids, and code assistants.

\bibliographystyle{plainnat}  
\bibliography{refs}

@inproceedings{
Jiang2020Fantastic,
title={Fantastic Generalization Measures and Where to Find Them},
author={Yiding Jiang and Behnam Neyshabur and Hossein Mobahi and Dilip Krishnan and Samy Bengio},
booktitle={International Conference on Learning Representations},
year={2020}
}

@inproceedings{sharpminimacangeneralize,
  title={Sharp minima can generalize for deep nets},
  author={Dinh, Laurent and Pascanu, Razvan and Bengio, Samy and Bengio, Yoshua},
  booktitle={International Conference on Machine Learning},
  pages={1019--1028},
  year={2017},
  organization={PMLR}
}

@article{ding2024flat,
  title={Flat minima generalize for low-rank matrix recovery},
  author={Ding, Lijun and Drusvyatskiy, Dmitriy and Fazel, Maryam and Harchaoui, Zaid},
  journal={Information and Inference: A Journal of the IMA},
  volume={13},
  number={2},
  pages={iaae009},
  year={2024},
  publisher={Oxford University Press}
}

@article{xu2026doessgdseekflatness,
  title={Does SGD Seek Flatness or Sharpness? An Exactly Solvable Model},
  author={Xu, Yizhou and Beneventano, Pierfrancesco and Chuang, Isaac and Ziyin, Liu},
  journal={arXiv preprint arXiv:2602.05065},
  year={2026}
}

@article{kamber2025sharpness,
  title={Sharpness of Minima in Deep Matrix Factorization},
  author={Kamber, Anil and Parhi, Rahul},
  journal={arXiv preprint arXiv:2509.25783},
  year={2025}
}

@inproceedings{
wang2022large,
title={Large Learning Rate Tames Homogeneity: Convergence and Balancing Effect},
author={Yuqing Wang and Minshuo Chen and Tuo Zhao and Molei Tao},
booktitle={International Conference on Learning Representations},
year={2022},
}

@book{horn2012matrix,
  title={Matrix Analysis},
  author={Horn, R.A. and Johnson, C.R.},
  year={2012},
  publisher={Cambridge University Press}
}

@article{Wustab,
  title={How sgd selects the global minima in over-parameterized learning: A dynamical stability perspective},
  author={Wu, Lei and Ma, Chao and others},
  journal={Advances in Neural Information Processing Systems},
  volume={31},
  year={2018}
}

@article{liang2026stable,
  title={Stable minima of ReLU neural networks suffer from the curse of dimensionality: The neural shattering phenomenon},
  author={Liang, Tongtong and Qiao, Dan and Wang, Yu-Xiang and Parhi, Rahul},
  journal={Advances in Neural Information Processing Systems},
  volume={38},
  pages={24686--24750},
  year={2025}
}

@article{vaswani2017attention,
  title={Attention is all you need},
  author={Vaswani, Ashish and Shazeer, Noam and Parmar, Niki and Uszkoreit, Jakob and Jones, Llion and Gomez, Aidan N and Kaiser, {\L}ukasz and Polosukhin, Illia},
  journal={Advances in neural information processing systems},
  volume={30},
  year={2017}
}

@inproceedings{mulayoff,
  title={Unique properties of flat minima in deep networks},
  author={Mulayoff, Rotem and Michaeli, Tomer},
  booktitle={International conference on machine learning},
  pages={7108--7118},
  year={2020},
  organization={PMLR}
}

@article{kamber2026loss,
  title={On the Loss Landscape Geometry of Regularized Deep Matrix Factorization: Uniqueness and Sharpness},
  author={Kamber, Anil and Parhi, Rahul},
  journal={arXiv preprint arXiv:2603.27072},
  year={2026}
}

@article{keskar2016large,
  title={On large-batch training for deep learning: Generalization gap and sharp minima},
  author={Keskar, Nitish Shirish and Mudigere, Dheevatsa and Nocedal, Jorge and Smelyanskiy, Mikhail and Tang, Ping Tak Peter},
  journal={arXiv preprint arXiv:1609.04836},
  year={2016}
}

@article{mulayoff2021implicit,
  title={The implicit bias of minima stability: A view from function space},
  author={Mulayoff, Rotem and Michaeli, Tomer and Soudry, Daniel},
  journal={Advances in Neural Information Processing Systems},
  volume={34},
  pages={17749--17761},
  year={2021}
}

@inproceedings{nacson2022implicit,
  title={The Implicit Bias of Minima Stability in Multivariate Shallow {ReLU} Networks},
  author={Nacson, Mor Shpigel and Mulayoff, Rotem and Ongie, Greg and Michaeli, Tomer and Soudry, Daniel},
  booktitle={International Conference on Learning Representations},
  year={2023}
}

@inproceedings{
qiao2024stable,
title={Stable Minima Cannot Overfit in Univariate {ReLU} Networks: Generalization by Large Step Sizes},
author={Dan Qiao and Kaiqi Zhang and Esha Singh and Daniel Soudry and Yu-Xiang Wang},
booktitle={Advances in Neural Information Processing Systems},
	pages = {94163--94208},
	volume = {37},
year={2024},
}

@inproceedings{liang2026generalization,
  title = {Generalization Below the Edge of Stability: The Role of Data Geometry},
  author = {Liang, Tongtong and Cloninger, Alexander and Parhi, Rahul and Wang, Yu-Xiang},
  year = {2026},
  booktitle = {International Conference on Learning Representations (ICLR)},
}

@article{ruiz2025understanding,
  title={Understanding and improving length generalization in recurrent models},
  author={Ruiz, Ricardo Buitrago and Gu, Albert},
  journal={arXiv preprint arXiv:2507.02782},
  year={2025}
}

@article{wang2025testtime,
  title={Test-time regression: a unifying framework for designing sequence models with associative memory},
  author={Wang, Ke Alexander and Shi, Jiaxin and Fox, Emily B.},
  journal={arXiv preprint arXiv:2501.12352},
  year={2025}
}

@article{mcdermott2026lifelong,
  title={Position: Lifelong In-Context Learning with Transformers Requires Parametric Forms of Attention},
  author={McDermott, Luke and Heath Jr., Robert W. and Parhi, Rahul},
  journal={arXiv preprint arXiv:2606.25342},
  year={2026}
}

@inproceedings{schlag2021linear,
  title={Linear transformers are secretly fast weight programmers},
  author={Schlag, Imanol and Irie, Kazuki and Schmidhuber, J{\"u}rgen},
  booktitle={International Conference on Machine Learning},
  pages={9355--9366},
  year={2021},
  organization={PMLR}
}

@inproceedings{katharopoulos2020transformers,
  title={Transformers are {RNN}s: Fast autoregressive transformers with linear attention},
  author={Katharopoulos, Angelos and Vyas, Apoorv and Pappas, Nikolaos and Fleuret, Fran{\c{c}}ois},
  booktitle={International Conference on Machine Learning},
  pages={5156--5165},
  year={2020},
  organization={PMLR}
}

@inproceedings{yang2024parallelizing,
  title={Parallelizing linear transformers with the delta rule over sequence length},
  author={Yang, Songlin and Wang, Bailin and Zhang, Yu and Shen, Yikang and Kim, Yoon},
  booktitle={Advances in Neural Information Processing Systems},
  volume={37},
  year={2024}
}

@article{sun2024learning,
  title={Learning to (learn at test time): {RNN}s with expressive hidden states},
  author={Sun, Yu and Li, Xinhao and Dalal, Karan and Xu, Jiarui and Vikram, Arjun and Zhang, Genghan and Dubois, Yann and Chen, Xinlei and Wang, Xiaolong and Koyejo, Sanmi and Hashimoto, Tatsunori and Guestrin, Carlos},
  journal={arXiv preprint arXiv:2407.04620},
  year={2024}
}

@article{liu2024longhorn,
  title={Longhorn: State space models are amortized online learners},
  author={Liu, Bo and Wang, Rui and Wu, Lemeng and Feng, Yihao and Stone, Peter and Liu, Qiang},
  journal={arXiv preprint arXiv:2407.14207},
  year={2024}
}

@article{behrouz2024titans,
  title={Titans: Learning to memorize at test time},
  author={Behrouz, Ali and Zhong, Peilin and Mirrokni, Vahab},
  journal={arXiv preprint arXiv:2501.00663},
  year={2024}
}

@article{vonoswald2025mesanet,
  title={{MesaNet}: Sequence modeling by locally optimal test-time training},
  author={von Oswald, Johannes and others},
  journal={arXiv preprint arXiv:2506.05233},
  year={2025}
}

@book{kailath1980linear,
  author    = {Kailath, Thomas},
  title     = {Linear Systems},
  publisher = {Prentice-Hall},
  year      = {1980}
}

@article{choromanski2020rethinking,
  title={Rethinking attention with performers},
  author={Choromanski, Krzysztof and Likhosherstov, Valerii and Dohan, David and Song, Xingyou and Gane, Andreea and Sarlos, Tamas and Hawkins, Peter and Davis, Jared and Mohiuddin, Afroz and Kaiser, Lukasz and others},
  journal={arXiv preprint arXiv:2009.14794},
  year={2020}
}

@article{peng2021random,
  title={Random feature attention},
  author={Peng, Hao and Pappas, Nikolaos and Yogatama, Dani and Schwartz, Roy and Smith, Noah A and Kong, Lingpeng},
  journal={arXiv preprint arXiv:2103.02143},
  year={2021}
}

@inproceedings{qin2022devil,
  title={The devil in linear transformer},
  author={Qin, Zhen and Han, Xiaodong and Sun, Weixuan and Li, Dongxu and Kong, Lingpeng and Barnes, Nick and Zhong, Yiran},
  booktitle={Proceedings of the 2022 Conference on Empirical Methods in Natural Language Processing},
  pages={7025--7041},
  year={2022}
}

@article{yang2023gated,
  title={Gated linear attention transformers with hardware-efficient training},
  author={Yang, Songlin and Wang, Bailin and Shen, Yikang and Panda, Rameswar and Kim, Yoon},
  journal={arXiv preprint arXiv:2312.06635},
  year={2023}
}

@article{sun2023retentive,
  title={Retentive network: A successor to transformer for large language models},
  author={Sun, Yutao and Dong, Li and Huang, Shaohan and Ma, Shuming and Xia, Yuqing and Xue, Jilong and Wang, Jianyong and Wei, Furu},
  journal={arXiv preprint arXiv:2307.08621},
  year={2023}
}

@article{katsch2023gateloop,
  title={Gateloop: Fully data-controlled linear recurrence for sequence modeling},
  author={Katsch, Tobias},
  journal={arXiv preprint arXiv:2311.01927},
  year={2023}
}

@article{qin2023hierarchically,
  title={Hierarchically gated recurrent neural network for sequence modeling},
  author={Qin, Zhen and Yang, Songlin and Zhong, Yiran},
  journal={Advances in Neural Information Processing Systems},
  volume={36},
  pages={33202--33221},
  year={2023}
}

@inproceedings{peng2023rwkv,
  title={Rwkv: Reinventing rnns for the transformer era},
  author={Peng, Bo and Alcaide, Eric and Anthony, Quentin and Albalak, Alon and Arcadinho, Samuel and Biderman, Stella and Cao, Huanqi and Cheng, Xin and Chung, Michael and Derczynski, Leon and others},
  booktitle={Findings of the association for computational linguistics: EMNLP 2023},
  pages={14048--14077},
  year={2023}
}

@article{de2024griffin,
  title={Griffin: Mixing gated linear recurrences with local attention for efficient language models},
  author={De, Soham and Smith, Samuel L and Fernando, Anushan and Botev, Aleksandar and Cristian-Muraru, George and Gu, Albert and Haroun, Ruba and Berrada, Leonard and Chen, Yutian and Srinivasan, Srivatsan and others},
  journal={arXiv preprint arXiv:2402.19427},
  year={2024}
}

@article{beck2024xlstm,
  title={xlstm: Extended long short-term memory},
  author={Beck, Maximilian and P{\"o}ppel, Korbinian and Spanring, Markus and Auer, Andreas and Prudnikova, Oleksandra and Kopp, Michael and Klambauer, G{\"u}nter and Brandstetter, Johannes and Hochreiter, Sepp},
  journal={Advances in Neural Information Processing Systems},
  volume={37},
  pages={107547--107603},
  year={2024}
}

@inproceedings{yang2025gated,
  title={Gated delta networks: Improving mamba2 with delta rule},
  author={Yang, Songlin and Kautz, Jan and Hatamizadeh, Ali},
  booktitle={International Conference on Learning Representations},
  volume={2025},
  pages={29687--29707},
  year={2025}
}

@article{iriegoing,
  title={Going Beyond Linear Transformers with Recurrent Fast Weight Programmers. NeurIPS 2021},
  author={Irie, K and Schlag, I and Csordas, R and Schmidhuber, J},
  journal={Preprint}
}

@inbook{adaptiveswitchingcircuits, author = {Widrow, Bernard and Hoff, Marcian E.}, title = {Adaptive switching circuits}, year = {1988}, isbn = {0262010976}, publisher = {MIT Press}, address = {Cambridge, MA, USA}, booktitle = {Neurocomputing: Foundations of Research}, pages = {123–134}, numpages = {12} }

@article{gu2020hippo,
  title={Hippo: Recurrent memory with optimal polynomial projections},
  author={Gu, Albert and Dao, Tri and Ermon, Stefano and Rudra, Atri and R{\'e}, Christopher},
  journal={Advances in neural information processing systems},
  volume={33},
  pages={1474--1487},
  year={2020}
}

@article{siems2026deltaproduct,
  title={Deltaproduct: Improving state-tracking in linear rnns via householder products},
  author={Siems, Julien and Carstensen, Timur and Zela, Arber and Hutter, Frank and Pontil, Massimiliano and Grazzi, Riccardo},
  journal={Advances in Neural Information Processing Systems},
  volume={38},
  pages={153738--153782},
  year={2026}
}

@article{gu2021combining,
  title={Combining recurrent, convolutional, and continuous-time models with linear state space layers},
  author={Gu, Albert and Johnson, Isys and Goel, Karan and Saab, Khaled and Dao, Tri and Rudra, Atri and R{\'e}, Christopher},
  journal={Advances in neural information processing systems},
  volume={34},
  pages={572--585},
  year={2021}
}

@article{gu2021efficiently,
  title={Efficiently modeling long sequences with structured state spaces},
  author={Gu, Albert and Goel, Karan and R{\'e}, Christopher},
  journal={arXiv preprint arXiv:2111.00396},
  year={2021}
}

@article{gupta2022diagonal,
  title={Diagonal state spaces are as effective as structured state spaces},
  author={Gupta, Ankit and Gu, Albert and Berant, Jonathan},
  journal={Advances in neural information processing systems},
  volume={35},
  pages={22982--22994},
  year={2022}
}

@inproceedings{orvieto2023resurrecting,
  title={Resurrecting recurrent neural networks for long sequences},
  author={Orvieto, Antonio and Smith, Samuel L and Gu, Albert and Fernando, Anushan and Gulcehre, Caglar and Pascanu, Razvan and De, Soham},
  booktitle={International conference on machine learning},
  pages={26670--26698},
  year={2023},
  organization={PMLR}
}

@article{gu2023mamba,
  title={Mamba: Linear-time sequence modeling with selective state spaces},
  author={Gu, Albert and Dao, Tri},
  journal={arXiv preprint arXiv:2312.00752},
  year={2023}
}

@article{dao2024transformers,
  title={Transformers are SSMs: Generalized models and efficient algorithms through structured state space duality},
  author={Dao, Tri and Gu, Albert},
  journal={arXiv preprint arXiv:2405.21060},
  year={2024}
}

@inproceedings{lahoti2026mamba,
  title={Mamba-3: Improved sequence modeling using state space principles},
  author={Lahoti, Aakash Sunil and Li, Kevin and Chen, Berlin and Wang, Caitlin and Bick, Aviv and Kolter, Zico and Dao, Tri and Gu, Albert},
  booktitle={International Conference on Learning Representations},
  volume={2026},
  pages={86173--86202},
  year={2026}
}

@article{smith2022simplified,
  title={Simplified state space layers for sequence modeling},
  author={Smith, Jimmy TH and Warrington, Andrew and Linderman, Scott W},
  journal={arXiv preprint arXiv:2208.04933},
  year={2022}
}

@article{fu2022hungry,
  title={Hungry hungry hippos: Towards language modeling with state space models},
  author={Fu, Daniel Y and Dao, Tri and Saab, Khaled K and Thomas, Armin W and Rudra, Atri and R{\'e}, Christopher},
  journal={arXiv preprint arXiv:2212.14052},
  year={2022}
}

@article{schmidhuber1992learning,
  title={Learning to control fast-weight memories: An alternative to dynamic recurrent networks},
  author={Schmidhuber, J{\"u}rgen},
  journal={Neural Computation},
  volume={4},
  number={1},
  pages={131--139},
  year={1992},
  publisher={MIT Press, One Rogers Street, Cambridge, MA 02142-1209, USA journals-info~…}
}

@inproceedings{hinton1987using,
  title={Using fast weights to deblur old memories},
  author={Hinton, Geoffrey E and Plaut, David C},
  booktitle={Proceedings of the ninth annual conference of the Cognitive Science Society},
  pages={177--186},
  year={1987}
}

@article{ba2016using,
  title={Using fast weights to attend to the recent past},
  author={Ba, Jimmy and Hinton, Geoffrey E and Mnih, Volodymyr and Leibo, Joel Z and Ionescu, Catalin},
  journal={Advances in neural information processing systems},
  volume={29},
  year={2016}
}

@inproceedings{sun2020test,
  title={Test-time training with self-supervision for generalization under distribution shifts},
  author={Sun, Yu and Wang, Xiaolong and Liu, Zhuang and Miller, John and Efros, Alexei and Hardt, Moritz},
  booktitle={International conference on machine learning},
  pages={9229--9248},
  year={2020},
  organization={PMLR}
}

@inproceedings{zhang2026test,
  title={Test-time training done right},
  author={Zhang, Tianyuan and Bi, Sai and Hong, Yicong and Zhang, Kai and Luan, Fujun and Yang, Songlin and Sunkavalli, Kalyan and Freeman, William and Tan, Hao},
  booktitle={International Conference on Learning Representations},
  volume={2026},
  pages={157604--157638},
  year={2026}
}

@inproceedings{behrouz2026s,
  title={It's all connected: A journey through test-time memorization, attentional bias, retention, and online optimization},
  author={Behrouz, Ali and Razaviyayn, Meisam and Zhong, Peilin and Mirrokni, Vahab},
  booktitle={International Conference on Learning Representations},
  volume={2026},
  pages={131306--131333},
  year={2026}
}

@article{zhong2026rethinking,
  title={Rethinking Expressivity and Efficiency in Test-Time Training},
  author={Zhong, Zeyun and Chen, Joya and Martin, Manuel and Diederichs, Frederik and Gall, Juergen and Beyerer, Juergen},
  journal={arXiv preprint arXiv:2608.21308},
  year={2026}
}

@inproceedings{cohen2022implicit,
  title={On the implicit bias of gradient descent for temporal extrapolation},
  author={Cohen-Karlik, Edo and David, Avichai Ben and Cohen, Nadav and Globerson, Amir},
  booktitle={International Conference on Artificial Intelligence and Statistics},
  pages={10966--10981},
  year={2022},
  organization={PMLR}
}

@article{jastrzkebski2017three,
  title={Three factors influencing minima in sgd},
  author={Jastrz{\k{e}}bski, Stanis{\l}aw and Kenton, Zachary and Arpit, Devansh and Ballas, Nicolas and Fischer, Asja and Bengio, Yoshua and Storkey, Amos},
  journal={arXiv preprint arXiv:1711.04623},
  year={2017}
}

@article{smith2021origin,
  title={On the origin of implicit regularization in stochastic gradient descent},
  author={Smith, Samuel L and Dherin, Benoit and Barrett, David GT and De, Soham},
  journal={arXiv preprint arXiv:2101.12176},
  year={2021}
}

@article{li2019towards,
  title={Towards explaining the regularization effect of initial large learning rate in training neural networks},
  author={Li, Yuanzhi and Wei, Colin and Ma, Tengyu},
  journal={Advances in neural information processing systems},
  volume={32},
  year={2019}
}

@article{barrett2020implicit,
  title={Implicit gradient regularization},
  author={Barrett, David GT and Dherin, Benoit},
  journal={arXiv preprint arXiv:2009.11162},
  year={2020}
}

@inproceedings{andriushchenko2023sgd,
  title={Sgd with large step sizes learns sparse features},
  author={Andriushchenko, Maksym and Varre, Aditya Vardhan and Pillaud-Vivien, Loucas and Flammarion, Nicolas},
  booktitle={International Conference on Machine Learning},
  pages={903--925},
  year={2023},
  organization={PMLR}
}

@inproceedings{barsbey2025large,
  title={Large learning rates simultaneously achieve robustness to spurious correlations and compressibility},
  author={Barsbey, Melih and Prieto, Lucas and Zafeiriou, Stefanos and Birdal, Tolga},
  booktitle={2025 IEEE/CVF International Conference on Computer Vision (ICCV)},
  pages={2055--2066},
  year={2025},
  organization={IEEE}
}

@article{anil2022exploring,
  title={Exploring length generalization in large language models},
  author={Anil, Cem and Wu, Yuhuai and Andreassen, Anders and Lewkowycz, Aitor and Misra, Vedant and Ramasesh, Vinay and Slone, Ambrose and Gur-Ari, Guy and Dyer, Ethan and Neyshabur, Behnam},
  journal={Advances in Neural Information Processing Systems},
  volume={35},
  pages={38546--38556},
  year={2022}
}

@article{lobacheva2023large,
  title={Large Learning Rates Improve Generalization: But How Large Are We Talking About?},
  author={Lobacheva, Ekaterina and Pockonechnyy, Eduard and Kodryan, Maxim and Vetrov, Dmitry},
  journal={arXiv preprint arXiv:2311.11303},
  year={2023}
}

@article{galli2026flatland,
  title={Flatland: The Adventures of Gradient Descent with Large Step Sizes},
  author={Galli, Leonardo and Fox, Curtis and Bartolomaeus, Wiebke and Schmidt, Mark and Rauhut, Holger},
  journal={arXiv preprint arXiv:2606.06722},
  year={2026}
}

@article{andriushchenko2023modern,
  title={A modern look at the relationship between sharpness and generalization},
  author={Andriushchenko, Maksym and Croce, Francesco and M{\"u}ller, Maximilian and Hein, Matthias and Flammarion, Nicolas},
  journal={arXiv preprint arXiv:2302.07011},
  year={2023}
}

@article{sadrtdinov2024large,
  title={Where do large learning rates lead us?},
  author={Sadrtdinov, Ildus and Kodryan, Maxim and Pokonechny, Eduard and Lobacheva, Ekaterina and Vetrov, Dmitry},
  journal={Advances in Neural Information Processing Systems},
  volume={37},
  pages={58445--58479},
  year={2024}
}

@article{smith2017don,
  title={Don't decay the learning rate, increase the batch size},
  author={Smith, Samuel L and Kindermans, Pieter-Jan and Ying, Chris and Le, Quoc V},
  journal={arXiv preprint arXiv:1711.00489},
  year={2017}
}

@article{lewkowycz2020large,
  title={The large learning rate phase of deep learning: the catapult mechanism},
  author={Lewkowycz, Aitor and Bahri, Yasaman and Dyer, Ethan and Sohl-Dickstein, Jascha and Gur-Ari, Guy},
  journal={arXiv preprint arXiv:2003.02218},
  year={2020}
}

@article{wang2025training,
  title={Training Instabilities Induce Flatness Bias in Gradient Descent},
  author={Wang, Lawrence and Roberts, Stephen J},
  journal={arXiv preprint arXiv:2511.12558},
  year={2025}
}

@inproceedings{mohtashami2023special,
  title={Special properties of gradient descent with large learning rates},
  author={Mohtashami, Amirkeivan and Jaggi, Martin and Stich, Sebastian U},
  booktitle={International conference on machine learning},
  pages={25082--25104},
  year={2023},
  organization={PMLR}
}

@article{cohen2021gradient,
  title={Gradient descent on neural networks typically occurs at the edge of stability},
  author={Cohen, Jeremy M and Kaur, Simran and Li, Yuanzhi and Kolter, J Zico and Talwalkar, Ameet},
  journal={arXiv preprint arXiv:2103.00065},
  year={2021}
}

@article{kalra2023phase,
  title={Phase diagram of early training dynamics in deep neural networks: effect of the learning rate, depth, and width},
  author={Kalra, Dayal Singh and Barkeshli, Maissam},
  journal={Advances in Neural Information Processing Systems},
  volume={36},
  pages={51621--51662},
  year={2023}
}

@inproceedings{kalra2025universal,
  title     = {Universal Sharpness Dynamics in Neural Network Training: Fixed Point Analysis, Edge of Stability, and Route to Chaos},
  author    = {Kalra, Dayal Singh and He, Tianyu and Barkeshli, Maissam},
  booktitle = {International Conference on Learning Representations (ICLR)},
  year      = {2025}
}

@article{wang2022analyzing,
  title={Analyzing sharpness along gd trajectory: Progressive sharpening and edge of stability},
  author={Wang, Zixuan and Li, Zhouzi and Li, Jian},
  journal={Advances in Neural Information Processing Systems},
  volume={35},
  pages={9983--9994},
  year={2022}
}

@article{zhu2022understanding,
  title={Understanding edge-of-stability training dynamics with a minimalist example},
  author={Zhu, Xingyu and Wang, Zixuan and Wang, Xiang and Zhou, Mo and Ge, Rong},
  journal={arXiv preprint arXiv:2210.03294},
  year={2022}
}

@article{jastrzebski2020break,
  title={The break-even point on optimization trajectories of deep neural networks},
  author={Jastrzebski, Stanislaw and Szymczak, Maciej and Fort, Stanislav and Arpit, Devansh and Tabor, Jacek and Cho, Kyunghyun and Geras, Krzysztof},
  journal={arXiv preprint arXiv:2002.09572},
  year={2020}
}

@article{yoo2025understanding,
  title={Understanding sharpness dynamics in NN training with a minimalist example: The effects of dataset difficulty, depth, stochasticity, and more},
  author={Yoo, Geonhui and Song, Minhak and Yun, Chulhee},
  journal={arXiv preprint arXiv:2506.06940},
  year={2025}
}

@inproceedings{hautus1969controllability,
  title={Controllability and observability conditions of linear autonomous systems},
  author={Hautus, Malo LJ},
  booktitle={Indagationes mathematicae (proceedings)},
  year={1969}
}

@article{gao2020pile,
  title={The pile: An 800gb dataset of diverse text for language modeling},
  author={Gao, Leo and Biderman, Stella and Black, Sid and Golding, Laurence and Hoppe, Travis and Foster, Charles and Phang, Jason and He, Horace and Thite, Anish and Nabeshima, Noa and others},
  journal={arXiv preprint arXiv:2101.00027},
  year={2020}
}

@article{penedo2024fineweb,
  title={The fineweb datasets: Decanting the web for the finest text data at scale},
  author={Penedo, Guilherme and Kydl{\'\i}{\v{c}}ek, Hynek and Lozhkov, Anton and Mitchell, Margaret and Raffel, Colin and Von Werra, Leandro and Wolf, Thomas and others},
  journal={Advances in Neural Information Processing Systems},
  volume={37},
  pages={30811--30849},
  year={2024}
}

@article{lanczos1950iteration,
  title={An iteration method for the solution of the eigenvalue problem of linear differential and integral operators},
  author={Lanczos, Cornelius},
  journal={Journal of research of the National Bureau of Standards},
  volume={45},
  number={4},
  pages={255--282},
  year={1950}
}

@article{jumper2021highly,
  title={Highly accurate protein structure prediction with AlphaFold},
  author={Jumper, John and Evans, Richard and Pritzel, Alexander and Green, Tim and Figurnov, Michael and Ronneberger, Olaf and Tunyasuvunakool, Kathryn and Bates, Russ and {\v{Z}}{\'\i}dek, Augustin and Potapenko, Anna and others},
  journal={nature},
  volume={596},
  number={7873},
  pages={583--589},
  year={2021},
  publisher={Nature Publishing Group UK, London}
}

@article{dill2012protein,
  title={The protein-folding problem, 50 years on},
  author={Dill, Ken A and MacCallum, Justin L},
  journal={science},
  volume={338},
  number={6110},
  pages={1042--1046},
  year={2012},
  publisher={American Association for the Advancement of Science}
}

@article{lake2016building,
  title={Building machines that learn and think like people},
  author={Lake, Brenden M and Ullman, Tomer D and Tenenbaum, Joshua B and Gershman, Samuel J},
  journal={arXiv preprint arXiv:1604.00289},
  year={2016}
}

@article{frobenius1878lineare,
  title={{\"U}ber lineare Substitutionen und bilineare Formen},
  author={Frobenius, Herrn},
  journal={Journal f{\"u}r die reine und angewandte Mathematik (Crelles Journal)},
  volume={1878},
  number={84},
  pages={1--63},
  year={1878},
  publisher={De Gruyter}
}

@book{chen1984linear,
  title={Linear system theory and design},
  author={Chen, Chi-Tsong and Chen, Chi-Tsong},
  volume={301},
  year={1984},
  publisher={Holt, Rinehart and Winston, New York}
}

@book{dattorro2005convex,
  title={Convex optimization \& Euclidean distance geometry},
  author={Dattorro, Jon},
  volume={105},
  year={2005},
  publisher={Meboo Publishing, USA}
}

\appendix

\section{Technical Preliminaries}
\begin{lemma}
\label[lemma]{columnspace}
    Let $\mathbf{Y} \in \mathbb{R}^{m \times n}$ and
    $\mathbf{A} \in \mathbb{R}^{p \times n}$.
    Then, $\mathbf{Y} = \mathbf{X}\mathbf{A}$
    admits a solution $\mathbf{X} \in \mathbb{R}^{m \times p}$
    if and only if
    \begin{equation}
        \operatorname{ran}(\mathbf{Y}^{\top})
        \subseteq
        \operatorname{ran}(\mathbf{A}^{\top}).
    \end{equation}
\end{lemma}

\begin{proof}
    Suppose there exists $\mathbf{X} \in \mathbb{R}^{m \times p}$
    such that $\mathbf{Y} = \mathbf{X}\mathbf{A}$.
    Taking transposes yields
    \begin{equation}
        \mathbf{Y}^{\top}
        = \mathbf{A}^{\top}\mathbf{X}^{\top}.
    \end{equation}
    Thus, every column of $\mathbf{Y}^{\top}$ is a linear
    combination of the columns of $\mathbf{A}^{\top}$, which implies
    \begin{equation}
        \operatorname{ran}(\mathbf{Y}^{\top})
        \subseteq
        \operatorname{ran}(\mathbf{A}^{\top}).
    \end{equation}
    Conversely, suppose that
    $\operatorname{ran}(\mathbf{Y}^{\top})
    \subseteq \operatorname{ran}(\mathbf{A}^{\top})$.
    Then, every column $\mathbf{y}_i$ of $\mathbf{Y}^{\top}$
    belongs to the column space of $\mathbf{A}^{\top}$.
    Consequently, for each $i \in [m]$, there exists
    $\mathbf{x}_i \in \mathbb{R}^{p}$ such that
    \begin{equation}
        \mathbf{y}_i = \mathbf{A}^{\top}\mathbf{x}_i.
    \end{equation}
    Stacking these $\mathbf{x}_i$'s horizontally, we define
    \begin{equation}
        \mathbf{X}^{\top}
        =
        \begin{bmatrix}
            \mathbf{x}_1 & \cdots & \mathbf{x}_m
        \end{bmatrix}
    \end{equation}
    and obtain
    \begin{equation}
        \mathbf{Y}^{\top}
        = \mathbf{A}^{\top}\mathbf{X}^{\top}.
    \end{equation}
    Therefore, $\mathbf{X}$ is a solution to $\mathbf{Y} = \mathbf{X}\mathbf{A}$.
\end{proof}

\begin{lemma}
\label[lemma]{almostsurely}
Suppose that we partially observe $N$ independent trajectories of the linear dynamical system given in \eqref{LDS} up to the time horizon $T$. Recall the zero-padded history matrix $\mathbf{Z}$, defined in \Cref{remark:representation} and formed from the $N$ independent trajectories $\{(\mathbf{y}_0^{(i)}, \mathbf{y}_1^{(i)}, \ldots, \mathbf{y}_T^{(i)})\}_{i=1}^{N}$. If $N \geq m$, then $\mathbf{Z}\mathbf{Z}^\top$ is nonsingular almost surely.
\end{lemma}
\begin{proof}
Note that $\mathbf{Z} \in \mathbb{R}^{mT \times NT}$; therefore, $\mathbf{Z}\mathbf{Z}^\top \in \mathbb{R}^{mT \times mT}$. By using the identity $\operatorname{rank}(\mathbf{A}\mathbf{B}) \leq \min \{\operatorname{rank}(\mathbf{A}), \operatorname{rank}(\mathbf{B})\}$,
\begin{equation}
    \operatorname{rank}(\mathbf{Z}\mathbf{Z}^\top) \leq \operatorname{rank}(\mathbf{Z}) \leq \min \{mT, NT\}. 
\end{equation}
If $N < m$, then $\operatorname{rank}(\mathbf{Z}\mathbf{Z}^\top) < mT$. Therefore, we need $N \geq m$ for the nonsingularity of $\mathbf{Z}\mathbf{Z}^\top$ with 
\begin{equation}
    \operatorname{rank}(\mathbf{Z}\mathbf{Z}^\top) = \operatorname{rank}(\mathbf{Z}) = mT. 
\end{equation}
This implies that $\mathbf{Z}$ must have full rank, in particular full row rank.
 
Recall that $\mathbf{Z}$ is a block upper-triangular matrix such that
\begin{equation}
    \mathbf{Z}
    := \begin{bmatrix}
        \mathbf{Y}_0 & \mathbf{Y}_1 & \cdots & \mathbf{Y}_{T-1} \\
        \mathbf{0}   & \mathbf{Y}_0 & \cdots & \mathbf{Y}_{T-2} \\
        \vdots       & \ddots       & \ddots & \vdots \\
        \mathbf{0}   & \cdots       & \mathbf{0} & \mathbf{Y}_0
    \end{bmatrix} \in \mathbb{R}^{mT \times NT},
\end{equation}
where $\mathbf{Y}_k := [\mathbf{y}_k^{(1)} \ \cdots \ \mathbf{y}_k^{(N)}]
\in \mathbb{R}^{m \times N}$. Suppose that $\mathbf{Y}_0$ has full rank, i.e., $\operatorname{rank}(\mathbf{Y}_0) = m$. Define $\mathbf{u}
    := [\mathbf{u}_0^{\top} \ \cdots \
       \mathbf{u}_{T-1}^{\top}]^{\top}
    \in \mathbb{R}^{mT}$ such that $\mathbf{u}^{\top}\mathbf{Z}=\mathbf{0}$, where $\mathbf{u}_j \in \mathbb{R}^m$ for all $j \in [T-1]_0$. By the block
structure of $\mathbf{Z}$, its $(k+1)$st block column gives
\begin{equation}
    \sum_{j=0}^{k}
        \mathbf{u}_j^{\top}\mathbf{Y}_{k-j} = \mathbf{0},
\end{equation}
for all $k \in [T-1]_0$.
For $k=0$, since $\mathbf{Y}_0$ is full rank and $N \geq m$, $\operatorname{ker}(\mathbf{Y}_0^\top) = \{\mathbf{0}\}$. This implies that $\mathbf{u}_0=\mathbf{0}$. For $k=1$, 
\begin{equation}
    \mathbf{u}_0^\top \mathbf{Y}_1 +  \mathbf{u}_1^\top \mathbf{Y}_0 = \mathbf{u}_1^\top \mathbf{Y}_0 = \mathbf{0},
\end{equation}
which implies that $\mathbf{u}_1 = \mathbf{0}$. By induction, $\mathbf{u}=\mathbf{0}$. Therefore, $\operatorname{ker}(\mathbf{Z}^\top) = \{\mathbf{0}\}$, and $\mathbf{Z}$ has full rank.  

Conversely, if
$\mathbf{Y}_0$ does not have full row rank, choose a nonzero vector
$\mathbf{a}\in\mathbb{R}^m$ such that
$\mathbf{a}^{\top}\mathbf{Y}_0=\mathbf{0}$. Since the last block
row of $\mathbf{Z}$ is
$[\mathbf{0}\ \cdots\ \mathbf{0}\ \mathbf{Y}_0]$, the vector
$\mathbf{u}=[\mathbf{0}^{\top}\ \cdots\ \mathbf{0}^{\top}\
\mathbf{a}^{\top}]^{\top}$ is a nonzero left-null vector of
$\mathbf{Z}$. Therefore, $\mathbf{Z}$ is rank-deficient. By contrapositive, this proves that $\mathbf{Z}$ has full rank \emph{if and only if} $\operatorname{rank}(\mathbf{Y}_0) = m$.


Under the assumptions in \Cref{sec:ProblemSetup},
\begin{equation}
  \boldsymbol{\mathcal{Y}}_0
    \sim \mathcal{N}\!\left(
        \mathbf{C}\bar{\mathbf{x}}_0,
        \mathbf{C}\mathbf{P}_0\mathbf{C}^{\top}+\mathbf{R}
    \right) \: \: \text{and} \: \:
\mathbf{C}\mathbf{P}_0\mathbf{C}^{\top}+\mathbf{R}\succ\mathbf{0}.  
\end{equation}
The columns of $\mathbf{Y}_0$ are independent samples from a nondegenerate multivariate Gaussian distribution. If $N\geq m$, the first $m$ columns
therefore have a joint density on $\mathbb{R}^{m\times m}$.
The set of singular $m\times m$ matrices has Lebesgue measure zero. Hence, these columns form a nonsingular matrix almost surely, and
$\operatorname{rank}(\mathbf{Y}_0)=m$ almost surely. Therefore, $\mathbf{Z}$ has full row rank
almost surely, and thus $\mathbf{Z}\mathbf{Z}^{\top}$ is
nonsingular almost surely.    
\end{proof}

\subsection{Proof of \Cref{prop:similarity}}
\label[appendix]{appendix:proofofproposition}
\begin{proof}
For every $i \geq 0$, we have
\begin{align}
\mathbf{M}_i(\boldsymbol{\theta}_{\mathbf{S}})
&=
(\mathbf{W}_3\mathbf{S}^{-1})
(\mathbf{S}\mathbf{W}_1\mathbf{S}^{-1})^i
(\mathbf{S}\mathbf{W}_2)\\
&=
\mathbf{W}_3\mathbf{S}^{-1}
\mathbf{S}\mathbf{W}_1^i\mathbf{S}^{-1}
\mathbf{S}\mathbf{W}_2\\
&=
\mathbf{W}_3\mathbf{W}_1^i\mathbf{W}_2\\
&=
\mathbf{M}_i(\boldsymbol{\theta}).
\end{align}
Since $\hat{\mathcal{L}}$ depends on the parameters
only through the Markov parameters
$\{\mathbf{M}_i\}_{i=0}^{T-1}$, $\hat{\mathcal{L}}(\boldsymbol{\theta}_{\mathbf{S}})
= \hat{\mathcal{L}}(\boldsymbol{\theta})$.
Consequently, if $\boldsymbol{\theta}\in\Omega$,
then $\boldsymbol{\theta}_{\mathbf{S}}\in\Omega$
for every $\mathbf{S}\in\mathrm{GL}_d(\mathbb{R})$.

We now characterize when the orbit
$\{\boldsymbol{\theta}_{\mathbf{S}} :
\mathbf{S} \in \mathrm{GL}_d(\mathbb{R})\}$ is a singleton.
Fix a parameterization $\boldsymbol{\theta} \in \Theta$ and suppose that
$\boldsymbol{\theta}_{\mathbf{S}}
= \boldsymbol{\theta}$
for every $\mathbf{S}\in\mathrm{GL}_d(\mathbb{R})$. Then
\begin{equation}
\mathbf{S}\mathbf{W}_1\mathbf{S}^{-1}
= \mathbf{W}_1, \quad 
\mathbf{S}\mathbf{W}_2 = \mathbf{W}_2, \quad 
\mathbf{W}_3\mathbf{S}^{-1} = \mathbf{W}_3.
\end{equation}
Taking $\mathbf{S}=2\mathbf{I}_d$ in the
last two identities yields
$\mathbf{W}_2=\mathbf{0}$ and
$\mathbf{W}_3=\mathbf{0}$. Moreover, the first identity implies that
$\mathbf{W}_1$ commutes with every invertible
matrix. Let $\mathbf{S}_1 = \operatorname{diag}(\gamma_1, \gamma_2, \cdots, \gamma_d) \in \mathrm{GL}_d(\mathbb{R})$ be an invertible diagonal matrix with distinct diagonal entries, i.e., $\gamma_i \neq \gamma_j$ for $i \neq j$. Since $\mathbf{W}_1\mathbf{S}_1 = \mathbf{S}_1 \mathbf{W}_1$,
\begin{equation}
    (\mathbf{W}_1\mathbf{S}_1)_{ij} = w_{ij}\gamma_j = (\mathbf{S}_1\mathbf{W}_1)_{ij} = \gamma_i w_{ij},
\end{equation}
for every $(i,j) \in [d] \times [d]$, where the $(i,j)$th entry of $\mathbf{W}_1$ is denoted by $w_{ij}$. Consequently, for every $(i,j) \in [d] \times [d]$, 
\begin{equation}
    (\gamma_i - \gamma_j)w_{ij} = 0.
\end{equation}
Note that $\gamma_i \neq \gamma _j$ for $i \neq j$. Therefore,
\begin{equation}
    w_{ij} = 0
\end{equation}
for all off-diagonal entries. This implies that $\mathbf{W}_1$ must be a diagonal matrix. Furthermore, let $\mathbf{S}_2 \in\mathrm{GL}_d(\mathbb{R})$ be a permutation matrix that swaps $i$th and $j$th coordinates. By definition, $\mathbf{S}_2^{-1} =\mathbf{S}_2$. Therefore, the fact that $\mathbf{W}_1$ commutes with every permutation matrix implies that 
\begin{equation}
    \mathbf{S}_2\mathbf{W}_1\mathbf{S}_2^{-1} = \mathbf{S}_2\mathbf{W}_1\mathbf{S}_2 = \mathbf{W}_1.
\end{equation}
This holds if and only if $w_{ii} = w_{jj}$. Hence, $\mathbf{W}_1=\lambda\mathbf{I}_d$
for some $\lambda\in\mathbb{R}$.

Conversely, suppose that
$\mathbf{W}_1=\lambda\mathbf{I}_d$,
$\mathbf{W}_2=\mathbf{0}$, and
$\mathbf{W}_3=\mathbf{0}$.
Then, for every
$\mathbf{S}\in\mathrm{GL}_d(\mathbb{R})$,
\begin{equation}
    \boldsymbol{\theta}_{\mathbf{S}}
=
(\lambda\mathbf{I}_d,\mathbf{0},\mathbf{0})
=
\boldsymbol{\theta}.
\end{equation}
\end{proof}
\subsection{Matrix Calculus}

\label[appendix]{appendix:dir_derivatives}
Let $\mathcal{L}:\mathbb{R}^{K \times L} \to \mathbb{R}$ be a differentiable function with continuous first and second order derivatives on $\mathbb{R}^{K \times L}$. Our objective is to derive closed-form expressions for the first and second order directional derivatives of $\mathcal{L}$ in the direction of $\mathbf{U} \in \mathbb{R}^{K \times L}$, where $\norm{\mathbf{U}}_F < \infty$, denoted by $D_{\mathbf{U}} \mathcal{L}\left(\mathbf{X}\right)$ and $D_{\mathbf{U}}^2 \mathcal{L}\left(\mathbf{X}\right)$, respectively. By the limit definition of the derivative, the first derivative of $\mathcal{L}\left(\mathbf{X}\right)$ with respect to each entry of $\mathbf{X}$ can be expressed as follows:

\begin{equation}
    \frac{\partial \mathcal{L}\left(\mathbf{X}\right)}{\partial X_{ij}} = \lim _{\Delta t \rightarrow 0} \frac{\mathcal{L}\left(\mathbf{X}+ \Delta t \mathbf{e}_i \mathbf{e}_j^\top\right) - \mathcal{L}\left(\mathbf{X}\right)}{\Delta t}, 
    \label{lim}
\end{equation}
where $\mathbf{e}_i$ is the $i$th standard basis vector of $\mathbb{R}^K$ and $\mathbf{e}_j$ is the $j$th standard basis vector of $\mathbb{R}^L$. If the limit in (\ref{lim}) exists, then by substitution of variables
\begin{equation}
    \frac{\partial \mathcal{L}\left(\mathbf{X}\right)}{\partial X_{ij}} U_{ij}= \lim _{\Delta t \rightarrow 0} \frac{\mathcal{L}\left(\mathbf{X}+ \Delta t U_{ij}\mathbf{e}_i \mathbf{e}_j^\top\right) - \mathcal{L}\left(\mathbf{X}\right)}{\Delta t}.
    \label{lemma1proofreq}
\end{equation}
By definition, the total change in $\mathcal{L}(\mathbf{X})$ in the direction of $\mathbf{U}$ is the sum of changes due to each entry of $\mathbf{X}$. Then 
\begin{align}
D_{\mathbf{U}} \mathcal{L}(\mathbf{X}) &= \sum_{i,j \in [K] \times [L]}\frac{\partial \mathcal{L}(\mathbf{X})}{\partial X_{ij}} U_{ij} \label{first-order-form-frob}\\
&= \sum_{i,j \in [K] \times [L]}\lim _{\Delta t \rightarrow 0} \frac{\mathcal{L}(\mathbf{X}+ \Delta t U_{ij}\mathbf{e}_i \mathbf{e}_j^\top) - \mathcal{L}\left(\mathbf{X}\right)}{\Delta t} \\
&= \lim _{\Delta t \rightarrow 0} \frac{\mathcal{L}\left(\mathbf{X}+ \Delta t \mathbf{U}\right) - \mathcal{L}\left(\mathbf{X}\right)}{\Delta t} \\
    &= \lim _{\Delta t \rightarrow 0} \frac{\mathcal{L}\left(\mathbf{X}+ \left(\Delta t+t\right) \mathbf{U}\right) - \mathcal{L}\left(\mathbf{X}+t\mathbf{U}\right)}{\Delta t} \Biggl|_{t=0} \\
    &=\frac{\partial \mathcal{L}\left(\mathbf{X}+t\mathbf{U}\right)}{\partial t} \Biggl |_{t=0}. \label{Gateaux}
\end{align}
This is known as the \emph{Gâteaux derivative}, which represents the change in $\mathcal{L}\left(\mathbf{X}\right)$ under a perturbation in the direction of $\mathbf{U}$ \citep{dattorro2005convex}. By the same reasoning, we obtain the following result.
\begin{lemma}[\citealp{dattorro2005convex}]
\label[lemma]{second-directional}
The second directional derivative of $\mathcal{L}$ at $\mathbf{X}$ in the direction $\mathbf{U} \in \mathbb{R}^{K \times L}$ is given by
\begin{equation}
    D_{\mathbf{U}}^2 \mathcal{L}\left(\mathbf{X}\right) = \frac{\partial^2}{\partial t^2} \mathcal{L}\left(\mathbf{X} + t\mathbf{U}\right) \Big |_{t = 0}.
    \label{Gateaux-2}
\end{equation}
\end{lemma}

\begin{proof}
We can express the derivative of $\frac{\partial \mathcal{L}(\mathbf{X})}{\partial X_{ij}}$ with respect to each entry of $\mathbf{X}$ using the limit definition of the derivative as follows: 
\begin{equation}
    \frac{\partial^2 \mathcal{L}(\mathbf{X})}{\partial X_{kl}\partial X_{ij}} = \frac{\partial}{\partial X_{kl}} \left( \frac{\partial \mathcal{L}(\mathbf{X})}{\partial X_{ij}} \right) = \lim_{\Delta t \rightarrow 0 } \frac{\partial \mathcal{L}(\mathbf{X}+\Delta t \mathbf{e}_i \mathbf{e}_j^\top)- \partial \mathcal{L}(\mathbf{X})}{\partial X_{kl} \Delta t}
\end{equation}
for all $(k,l) \in [K] \times [L]$. Then, we can rewrite it as 
\begin{equation}
\lim_{\Delta h, \Delta t \rightarrow 0} \frac{\mathcal{L}(\mathbf{X}+\Delta t \mathbf{e}_i \mathbf{e}_j ^\top + \Delta h \mathbf{e}_k \mathbf{e}_l^\top) - \mathcal{L}(\mathbf{X}+\Delta t \mathbf{e}_i \mathbf{e}_j ^\top)  - \mathcal{L}(\mathbf{X}+\Delta h \mathbf{e}_k \mathbf{e}_l^\top)+\mathcal{L}(\mathbf{X})}{\Delta h \Delta t}.   
\end{equation}
By using the substitution of variables as in (\ref{lemma1proofreq}),
\begin{equation}
   \frac{\partial^2 \mathcal{L}(\mathbf{X})}{\partial X_{kl}\partial X_{ij}} U_{ij} U_{kl} =\frac{\partial}{\partial X_{kl}} \left( \frac{\partial \mathcal{L}(\mathbf{X})}{\partial X_{ij}} U_{ij}\right) U_{kl} = \lim_{\Delta t \rightarrow 0 } \frac{\partial \mathcal{L}(\mathbf{X}+\Delta t U_{ij} \mathbf{e}_i \mathbf{e}_j^\top)- \partial \mathcal{L}(\mathbf{X})}{\partial X_{kl} \Delta t} U_{kl}.
   \label{42}
\end{equation}
Equivalently, 
\begin{equation}
\begin{aligned}
\lim_{\Delta h,\Delta t\to 0}
\frac{1}{\Delta h\,\Delta t}\Bigl[
& \mathcal{L}\bigl(\mathbf{X}
  + \Delta t U_{ij}\mathbf{e}_i\mathbf{e}_j^\top
  + \Delta h U_{kl}\mathbf{e}_k\mathbf{e}_l^\top\bigr) \\
&- \mathcal{L}\bigl(\mathbf{X}
  + \Delta t U_{ij}\mathbf{e}_i\mathbf{e}_j^\top\bigr)
 - \mathcal{L}\bigl(\mathbf{X}
  + \Delta h U_{kl}\mathbf{e}_k\mathbf{e}_l^\top\bigr)
 + \mathcal{L}(\mathbf{X})
\Bigr],
\end{aligned}
\end{equation}
which can be proved by substitution of variables in (\ref{42}). In turn, the second-order differential along any $\mathbf{U} \in \mathbb{R}^{K \times L}$ is
\begin{equation}
    D_{\mathbf{U}}^2 \mathcal{L}(\mathbf{X}) =\sum_{i,j} \sum_{k,l}  \frac{\partial^2 \mathcal{L}(\mathbf{X})}{\partial X_{kl}\partial X_{ij}} U_{ij} U_{kl} = \left<\nabla \left< \nabla \mathcal{L}(\mathbf{X}),\mathbf{U} \right>, \mathbf{U} \right>,
    \label{second-order-form-frob}
\end{equation}
where 
\begin{equation}
    \nabla \mathcal{L}(\mathbf{X}) = \begin{bmatrix}
        \frac{\partial \mathcal{L}(\mathbf{X})}{\partial X_{11}} &  \frac{\partial \mathcal{L}(\mathbf{X})}{\partial X_{12}} & \dots & \frac{\partial \mathcal{L}(\mathbf{X})}{\partial X_{1L}} \\ 
        \frac{\partial \mathcal{L}(\mathbf{X})}{\partial X_{21}} &  \frac{\partial \mathcal{L}(\mathbf{X})}{\partial X_{22}} & \dots & \frac{\partial \mathcal{L}(\mathbf{X})}{\partial X_{2L}} \\
        \vdots &  \ddots   & \dots & \vdots \\
        \frac{\partial \mathcal{L}(\mathbf{X})}{\partial X_{K1}} & \frac{\partial \mathcal{L}(\mathbf{X})}{\partial X_{K2}}   & \dots & \frac{\partial \mathcal{L}(\mathbf{X})}{\partial X_{KL}} \\
    \end{bmatrix} \in \mathbb{R}^{K \times L}.
\end{equation}
Equivalently,
\begin{align}
D_{\mathbf{U}}^2 \mathcal{L}(\mathbf{X}) =&\sum_{k,l} \lim_{\Delta t \rightarrow 0 } \frac{\partial \mathcal{L}(\mathbf{X}+\Delta t \mathbf{U})- \partial \mathcal{L}(\mathbf{X})}{\partial X_{kl} \Delta t} U_{kl} \\
=& \lim_{\Delta t \rightarrow 0} \frac{\mathcal{L}(\mathbf{X}+2\Delta t \mathbf{U}) - 2\mathcal{L}(\mathbf{X}+\Delta t \mathbf{U}) + \mathcal{L}(\mathbf{X})}{\Delta t^2} \\ 
=& \frac{\partial^2}{\partial t^2}
   \mathcal{L}(\mathbf{X} + t\mathbf{U})\biggr|_{t=0}\!.
\end{align}
\end{proof}

\begin{remark}
\label[remark]{remark:directional-taylor}
As we showed in (\ref{first-order-form-frob}) and (\ref{second-order-form-frob}),
\begin{align}
    D_{\mathbf{U}}\mathcal{L}(\mathbf{X})
    &= \sum_{i,j \in [K] \times [L]}\frac{\partial \mathcal{L}(\mathbf{X})}{\partial X_{ij}} U_{ij} = 
    \left\langle
        \nabla\mathcal{L}(\mathbf{X}),
        \mathbf{U}
    \right\rangle_F,
    \label{eq:first-directional-gradient}
    \\
    D_{\mathbf{U}}^2\mathcal{L}(\mathbf{X})
    &= \left<\nabla \left< \nabla \mathcal{L}(\mathbf{X}),\mathbf{U} \right>, \mathbf{U} \right> = 
    \operatorname{vec}(\mathbf{U})^\top
    \nabla_{\operatorname{vec}(\mathbf{X})}^2
    \mathcal{L}(\mathbf{X})
    \operatorname{vec}(\mathbf{U})
    \label{eq:second-directional-hessian},
\end{align}
respectively. Consequently, by the definition of the Rayleigh quotient\footnote{For a real symmetric matrix $\mathbf{A}$, $\lambda_{\max}(\mathbf{A}) = \max_{\|\mathbf{u}\|_2 = 1} \mathbf{u}^\top \mathbf{A} \mathbf{u}$.}, the maximum eigenvalue of the Hessian can be obtained as follows:
\begin{equation}
    \lambda_{\max}\!\left(
        \nabla^2
        \mathcal{L}(\mathbf{X})
    \right)
    =
    \max_{\|\mathbf{U}\|_F=1}
    D_{\mathbf{U}}^2\mathcal{L}(\mathbf{X}).
\label{eq:hessian-rayleigh-directional}
\end{equation}
\end{remark}

\begin{remark}
\label[remark]{remark:matrix-valued-taylor}
Let $
    \mathbf{F}:
    \mathbb{R}^{K\times L}
    \longrightarrow
    \mathbb{R}^{M\times N}$ 
be a twice continuously differentiable matrix-valued function of a matrix variable. Unlike the scalar-valued case,
the first derivative of $\mathbf{F}$ at $\mathbf{X}$ is a linear map
$ 
    D\mathbf{F}(\mathbf{X})
    :
    \mathbb{R}^{K\times L}
    \longrightarrow
    \mathbb{R}^{M\times N},
$ 
whereas the second derivative is a symmetric bilinear map
$ 
    D^2\mathbf{F}(\mathbf{X})
    :
    \mathbb{R}^{K\times L}
    \times
    \mathbb{R}^{K\times L}
    \longrightarrow
    \mathbb{R}^{M\times N}.
$ 
Accordingly, for any perturbation
$\boldsymbol{\Delta}\in\mathbb{R}^{K\times L}$,
the second-order Taylor expansion of $\mathbf{F}$ around $\mathbf{X}$
is
\begin{equation}
    \mathbf{F}(\mathbf{X}+\boldsymbol{\Delta})
    =
    \mathbf{F}(\mathbf{X})
    +
    D\mathbf{F}(\mathbf{X})[\boldsymbol{\Delta}]
    +
    \frac{1}{2}
    D^2\mathbf{F}(\mathbf{X})
    [\boldsymbol{\Delta},\boldsymbol{\Delta}]
    +
    o\!\left(
        \|\boldsymbol{\Delta}\|_F^2
    \right).
\label{eq:matrix-valued-taylor}
\end{equation}

For a fixed perturbation direction
$\mathbf{U}\in\mathbb{R}^{K\times L}$, our derivations for first and second directional derivatives in (\ref{Gateaux}) and (\ref{Gateaux-2}) generalize to a matrix-valued function $\mathbf{F}$ as follows: 
\begin{align}
D_{\mathbf{U}}\mathbf{F}(\mathbf{X})
:= \frac{\partial}{\partial t}\mathbf{F}(\mathbf{X}+t\mathbf{U})\biggr|_{t=0}
&= \scalebox{0.8}{$\begin{bmatrix}
\frac{\partial F_{11}(\mathbf{X}+t\mathbf{U})}{\partial t}\big|_{t=0} & \frac{\partial F_{12}(\mathbf{X}+t\mathbf{U})}{\partial t}\big|_{t=0} & \cdots & \frac{\partial F_{1N}(\mathbf{X}+t\mathbf{U})}{\partial t}\big|_{t=0} \\
\frac{\partial F_{21}(\mathbf{X}+t\mathbf{U})}{\partial t}\big|_{t=0} & \frac{\partial F_{22}(\mathbf{X}+t\mathbf{U})}{\partial t}\big|_{t=0} & \cdots & \frac{\partial F_{2N}(\mathbf{X}+t\mathbf{U})}{\partial t}\big|_{t=0} \\
\vdots & \vdots & \ddots & \vdots \\
\frac{\partial F_{M1}(\mathbf{X}+t\mathbf{U})}{\partial t}\big|_{t=0} & \frac{\partial F_{M2}(\mathbf{X}+t\mathbf{U})}{\partial t}\big|_{t=0} & \cdots & \frac{\partial F_{MN}(\mathbf{X}+t\mathbf{U})}{\partial t}\big|_{t=0}
\end{bmatrix}$}, \label{firstorderdirecformulaformatrix} \\
D_{\mathbf{U}}^2\mathbf{F}(\mathbf{X})
:= \frac{\partial^2}{\partial t^2}\mathbf{F}(\mathbf{X}+t\mathbf{U})\biggr|_{t=0}
&= \scalebox{0.8}{$\begin{bmatrix}
\frac{\partial^2 F_{11}(\mathbf{X}+t\mathbf{U})}{\partial t^2}\big|_{t=0} & \frac{\partial^2 F_{12}(\mathbf{X}+t\mathbf{U})}{\partial t^2}\big|_{t=0} & \cdots & \frac{\partial^2 F_{1N}(\mathbf{X}+t\mathbf{U})}{\partial t^2}\big|_{t=0} \\
\frac{\partial^2 F_{21}(\mathbf{X}+t\mathbf{U})}{\partial t^2}\big|_{t=0} & \frac{\partial^2 F_{22}(\mathbf{X}+t\mathbf{U})}{\partial t^2}\big|_{t=0} & \cdots & \frac{\partial^2 F_{2N}(\mathbf{X}+t\mathbf{U})}{\partial t^2}\big|_{t=0} \\
\vdots & \vdots & \ddots & \vdots \\
\frac{\partial^2 F_{M1}(\mathbf{X}+t\mathbf{U})}{\partial t^2}\big|_{t=0} & \frac{\partial^2 F_{M2}(\mathbf{X}+t\mathbf{U})}{\partial t^2}\big|_{t=0} & \cdots & \frac{\partial^2 F_{MN}(\mathbf{X}+t\mathbf{U})}{\partial t^2}\big|_{t=0}
\end{bmatrix}$}.
\end{align}
Then, the second-order Taylor expansion of $\mathbf{F}$ around $\mathbf{X}$ along the line $\mathbf{X}+t\mathbf{U}$ reduces to
\begin{equation}
\begin{aligned}
    \mathbf{F}(\mathbf{X}+t\mathbf{U})
    =
    \mathbf{F}(\mathbf{X})+
    tD_{\mathbf{U}}\mathbf{F}(\mathbf{X}) +
    \frac{t^2}{2}
    D_{\mathbf{U}}^2\mathbf{F}(\mathbf{X})
    +
    o(t^2).
\end{aligned}
\label{eq:matrix-valued-directional-taylor}
\end{equation}
Equivalently, define $\mathbf{f}(\mathbf{x}) := \operatorname{vec} \left(
        \mathbf{F}(\mathbf{X})
    \right) \in \mathbb{R}^{MN},
    \mathbf{x}
    :=
    \operatorname{vec}(\mathbf{X}) \in \mathbb{R}^{KL}$, and $\mathbf{h}
    :=
    \operatorname{vec}(\boldsymbol{\Delta}) \in \mathbb{R}^{KL}$. Then, the vectorized second-order Taylor expansion is
\begin{equation}
\begin{aligned}
    \mathbf{f}(\mathbf{x}+\mathbf{h})
    =
    \mathbf{f}(\mathbf{x})+
    \mathbf{J}_{\mathbf{F}}(\mathbf{X})\mathbf{h}+
    \frac{1}{2}
    \begin{bmatrix}
        \mathbf{h}^{\top}
        \nabla^2 f_1(\mathbf{x})
        \mathbf{h}
        \\
        \vdots
        \\
        \mathbf{h}^{\top}
        \nabla^2 f_{MN}(\mathbf{x})
        \mathbf{h}
    \end{bmatrix}
    +
    o\!\left(\|\mathbf{h}\|_2^2\right),
\end{aligned}
\label{eq:vectorized-matrix-valued-taylor}
\end{equation}
where
\begin{equation}
    \mathbf{J}_{\mathbf{F}}(\mathbf{X})
    :=
    \frac{
        \partial
        \operatorname{vec}(\mathbf{F}(\mathbf{X}))
    }{
        \partial
        \operatorname{vec}(\mathbf{X})
    }
    \in
    \mathbb{R}^{MN\times KL}
\end{equation}
is the Jacobian of the vectorized map, and
$\nabla ^2f_i(\mathbf{x}) \in \mathbb{R}^{KL \times KL}$ for all $i \in [MN]$. Hence,
\begin{equation}
        \operatorname{vec}
    \left(
        D_{\mathbf{U}}\mathbf{F}(\mathbf{X})
    \right)
    =
    \mathbf{J}_{\mathbf{F}}(\mathbf{X})
    \operatorname{vec}(\mathbf{U}).
    \label{eq:vec-first-directional}
\end{equation}
\end{remark}

\begin{lemma}
    Let $\mathbf{A}:\mathbb{R} \rightarrow \mathbb{R}^{m \times n}$
    be a continuously differentiable matrix-valued function.
    For any constant matrices $\mathbf{B} \in \mathbb{R}^{p \times m}$
    and $\mathbf{C} \in \mathbb{R}^{n \times q}$, we have
    \begin{equation}
        \frac{d}{dt}\left(\mathbf{B}\mathbf{A}(t)\mathbf{C}\right)
        = \mathbf{B}\frac{d\mathbf{A}(t)}{dt}\mathbf{C}.
    \end{equation}
    \label[lemma]{diffrule}
\end{lemma}

\begin{proof}
    By the limit definition of the derivative, we have
    \begin{align}
        \frac{d}{dt}\left(\mathbf{B}\mathbf{A}(t)\mathbf{C}\right)
        &= \lim_{h \to 0}
        \frac{\mathbf{B}\mathbf{A}(t+h)\mathbf{C}
        - \mathbf{B}\mathbf{A}(t)\mathbf{C}}{h} \nonumber\\
        &= \lim_{h \to 0}
        \mathbf{B}\left(
        \frac{\mathbf{A}(t+h)-\mathbf{A}(t)}{h}
        \right)\mathbf{C} \nonumber\\
        &= \mathbf{B}\left(
        \lim_{h \to 0}
        \frac{\mathbf{A}(t+h)-\mathbf{A}(t)}{h}
        \right)\mathbf{C} \nonumber\\
        &= \mathbf{B}\frac{d\mathbf{A}(t)}{dt}\mathbf{C}.
    \end{align}
\end{proof}

\begin{remark}
    Let $\mathbf{A}:\mathbb{R}_+ \rightarrow \mathbb{R}^{m \times n}$
    be a continuously differentiable matrix-valued function, and define
    $\mathbf{f}(t) := \mathbf{A}^\top(t)\mathbf{A}(t)$.
    By the product rule and the fact that differentiation commutes
    with transposition, we have
    \begin{equation}
        \frac{d\mathbf{f}(t)}{dt}
        =
        \left(\frac{d\mathbf{A}(t)}{dt}\right)^\top \mathbf{A}(t)
        +
        \mathbf{A}^\top(t)\frac{d\mathbf{A}(t)}{dt}.
    \end{equation}
    \label[remark]{diffchainrule}
\end{remark}

\subsection{Lower Shift Matrices}
\begin{definition}
    A lower shift matrix $\mathbf{L}_n \in \mathbb{R}^{n \times n}$ is a binary matrix with ones on the subdiagonal and zeros elsewhere. For instance, an $n\times n$ lower shift matrix $\mathbf{L}_n$ takes the form 
    \begin{equation}
        \mathbf{L}_n := \begin{bmatrix}
            0 & 0 & 0 &\cdots & 0 \\ 
            1 & 0 & 0 & \cdots & 0 \\ 
            0 & 1 & 0 & \cdots & 0 \\ 
            \vdots & \cdots & \ddots &\cdots & 0 \\ 
            0 & 0 & \cdots & 1 & 0 
        \end{bmatrix} \in \mathbb{R}^{n \times n}.
    \end{equation}
\end{definition}

\begin{remark}
\label[remark]{lowershiftmatrixremark1}
    Note that multiplying an $n$-dimensional vector $\mathbf{x} \in \mathbb{R}^n$ by the lower shift matrix $\mathbf{L}_n \in \mathbb{R}^{n \times n}$ shifts its entries down by one position, inserts a zero in the first position, and discards the last entry, i.e.,
    \begin{equation}
        \mathbf{L}_n \mathbf{x} = \begin{bmatrix}
            0 & x_1 & x_2 & \cdots &x_{n-1}
        \end{bmatrix}^\top.
    \end{equation}
    Furthermore, to shift the entries of $\mathbf{x}$ down by $m$ positions, we can apply the lower shift matrix $\mathbf{L}_n$ repeatedly $m$ times, i.e.,
\begin{equation}
    \mathbf{L}_n^m \mathbf{x} =
    \begin{bmatrix}
        \mathbf{0}_m^\top &
        x_1 &
        \cdots &
        x_{n-m}
    \end{bmatrix}^\top.
    \label{blockshifteq}
\end{equation}
When $n$ is divisible by $m$, this operation can equivalently be expressed using the Kronecker product as
\begin{equation}
    (\mathbf{L}_{n/m} \otimes \mathbf{I}_m)\mathbf{x}
    =
    \begin{bmatrix}
        \mathbf{0}_m^\top &
        x_1 &
        \cdots &
        x_{n-m}
    \end{bmatrix}^\top,
\end{equation}
where $\mathbf{L}_{n/m} \in \mathbb{R}^{\frac{n}{m} \times \frac{n}{m}}$.
When $n$ is not divisible by $m$, we can define
$\mathbf{P} := \begin{bmatrix}
    \mathbf{I}_n & \mathbf{0}_{n \times (qm-n)}
\end{bmatrix}$
such that
\begin{equation}
    \mathbf{L}_n^m
    = \mathbf{P}(\mathbf{L}_q \otimes \mathbf{I}_m)\mathbf{P}^\top,
\end{equation}
where $q := \lceil n/m \rceil$.
\end{remark}

\begin{remark}
    \label[remark]{remark:lower-shift-nilpotent}
    For any $n \times n$ lower shift matrix $\mathbf{L}_n \in \mathbb{R}^{n \times n}$, 
    \begin{equation}
        \mathbf{L}_n^n = \mathbf{0}.
    \end{equation}
\end{remark}

\begin{definition}
\label[definition]{def:nilpotent}
    A square matrix $\mathbf{N} \in \mathbb{R}^{n \times n}$ is \emph{nilpotent} if there exists $k \in \mathbb{N}$ such that $\mathbf{N}^k = \mathbf{0}$.
\end{definition}
By \Cref{remark:lower-shift-nilpotent}, every lower shift matrix is nilpotent.

\begin{lemma}
\label[lemma]{lemma:nilpotent}
    Let $\mathbf{N} \in \mathbb{R}^{n \times n}$ be a nilpotent matrix. Then, $\operatorname{spec}(\mathbf{N}) = \{0\}$.
\end{lemma}

\begin{proof}
    Suppose $\mathbf{N}^k = \mathbf{0}$ for some $k \in \mathbb{N}$. Let $\lambda \in \operatorname{spec}(\mathbf{N})$, and let $\mathbf{v} \in \mathbb{C}^n \setminus \{\mathbf{0}\}$ satisfy $\mathbf{N}\mathbf{v} = \lambda\mathbf{v}$. Then, for any $j \in \mathbb{N}$, we have
    \begin{equation}
        \mathbf{N}^{j}\mathbf{v}
        = \lambda^{j}\mathbf{v}.
    \end{equation}
    Taking $j = k$ yields
    \begin{equation}
        \mathbf{0} = \mathbf{N}^k\mathbf{v} = \lambda^k\mathbf{v}.
    \end{equation}
    Since $\mathbf{v} \neq \mathbf{0}$, we have $\lambda^k = 0$, and hence $\lambda = 0$.
\end{proof}

\subsection{Companion Matrix}

\begin{definition}[{\citealp[Equation~3.3.12]{horn2012matrix}}]
    \label[definition]{companionmatrix}
    For a degree-$n$ monic polynomial $p(x) = c_0 + c_1x + \ldots + c_{n-1}x^{n-1} + x^n$, the companion matrix is defined as a square matrix
    \begin{equation}
        H(p) := \begin{bmatrix}
0 & 0 & \cdots & 0 & -c_0 \\
1 & 0 & \cdots & 0 & -c_1 \\
0 & 1 & \cdots & 0 & -c_2 \\
\vdots & \vdots & \ddots & \vdots & \vdots \\
0 & 0 & \cdots & 1 & -c_{n-1}
\end{bmatrix} \in \mathbb{R}^{n \times n}.
    \end{equation}
\end{definition}

\section{Proof of \Cref{lemma:uniqueness}}
\label[appendix]{appendix:proof-uniqueness}

\begin{proof}
We first show that any prescribed sequence of Markov parameters within the training horizon can be realized by $\mathbf{\Phi}_T(\boldsymbol{\theta})$ when $d \geq mT$. This means that $\boldsymbol{\theta} \mapsto \mathbf{\Phi}_T(\boldsymbol{\theta})$ is surjective. 


Let $\mathbf{\Phi}
    = [\mathbf{M}_0 \ \cdots \ \mathbf{M}_{T-1}]
    \in \mathbb{R}^{m \times mT}$ be arbitrary, where $\mathbf{M}_k \in \mathbb{R}^{m \times m}$ for
every $k \in [T-1]_0$. Let $\mathbf{e}_1,\ldots,\mathbf{e}_T$ denote
the standard basis vectors of $\mathbb{R}^T$, and define a lower shift
matrix $\mathbf{L}_T \in \mathbb{R}^{T \times T}$. As we showed in \Cref{lowershiftmatrixremark1},  $\mathbf{L}_T\mathbf{e}_i = \mathbf{e}_{i+1}$ for all $i \in [T-1]$. 

For $d = mT$, we can choose a particular parameterization $\bar{\boldsymbol{\theta}} = (\overline{\mathbf{W}}_1,\overline{\mathbf{W}}_2, \overline{\mathbf{W}}_3) \in \Theta$ such that
\begin{equation}
    \overline{\mathbf{W}}_1 = \mathbf{L}_T \otimes \mathbf{I}_m,
    \qquad
    \overline{\mathbf{W}}_2 = \mathbf{e}_1 \otimes \mathbf{I}_m,
    \qquad
    \overline{\mathbf{W}}_3 = \mathbf{\Phi}.
    \label{eq:lemma3-shift-realization}
\end{equation}
For this particular parameterization $\bar{\boldsymbol{\theta}}$, we have
\begin{equation}
    \overline{\mathbf{W}}_1^{\,k}\overline{\mathbf{W}}_2
    = (\mathbf{L}_T^k\mathbf{e}_1) \otimes \mathbf{I}_m
    = \mathbf{e}_{k+1} \otimes \mathbf{I}_m,
\end{equation}
for every $k \in [T-1]_0$, where the last equality follows from \eqref{blockshifteq}. Note that 
\begin{equation}
    \mathbf{e}_{k+1} \otimes \mathbf{I}_m = \begin{bmatrix}
        \mathbf{0}_{mk \times m} \\ 
        \mathbf{I}_m \\ 
        \mathbf{0}_{(T-k-1)m \times m}
    \end{bmatrix}.
\end{equation}
Hence, 
\begin{equation}
        \overline{\mathbf{W}}_3
    \overline{\mathbf{W}}_1^{\,k}
    \overline{\mathbf{W}}_2
    = \mathbf{\Phi}(\mathbf{e}_{k+1} \otimes \mathbf{I}_m) 
     = \begin{bmatrix}
         \mathbf{M}_0 & \mathbf{M}_1 & \cdots & \mathbf{M}_{T-1}
     \end{bmatrix}\begin{bmatrix}
        \mathbf{0}_{mk \times m} \\ 
        \mathbf{I}_m \\ 
        \mathbf{0}_{(T-k-1)m \times m}
    \end{bmatrix} = \mathbf{M}_k.
    \label{eq:lemma3-prescribed-markov}
\end{equation}
Since $\mathbf{M}_k \in \mathbb{R}^{m \times m}$ was arbitrary, $\mathbf{\Phi}_T(\boldsymbol{\theta})$ can represent any $\mathbf{\Phi} \in \mathbb{R}^{m \times mT}$. Therefore, the map
$\boldsymbol{\theta} \mapsto \mathbf{\Phi}_T(\boldsymbol{\theta})$
is surjective.

If $d>mT$, let $r=d-mT$ and append $r$ unused hidden
coordinates by setting
\begin{equation}
    \mathbf{W}_1 =
    \begin{bmatrix}
        \overline{\mathbf{W}}_1 & \mathbf{0} \\
        \mathbf{0} & \mathbf{0}_{r \times r}
    \end{bmatrix},
    \qquad
    \mathbf{W}_2 =
    \begin{bmatrix}
        \overline{\mathbf{W}}_2 \\
        \mathbf{0}_{r \times m}
    \end{bmatrix},
    \qquad
    \mathbf{W}_3 =
    \begin{bmatrix}
        \overline{\mathbf{W}}_3 & \mathbf{0}_{m \times r}
    \end{bmatrix}.
    \label{eq:lemma3-padded-realization}
\end{equation}
The additional coordinates do not change the Markov parameters, so
$\mathbf{W}_3\mathbf{W}_1^k\mathbf{W}_2=\mathbf{M}_k$ for every
$k \in [T-1]_0$. Consequently, in either case there exists
$\boldsymbol{\theta} \in \Theta$ such that
$\mathbf{\Phi}_T(\boldsymbol{\theta})=\mathbf{\Phi}$. Since
$\mathbf{\Phi}$ was arbitrary, the map
$\boldsymbol{\theta} \mapsto \mathbf{\Phi}_T(\boldsymbol{\theta})$
is surjective onto $\mathbb{R}^{m \times mT}$. This lower-shift matrix realization was also leveraged by \citet{cohen2022implicit} in linear RNNs.

Note that 
\begin{equation}
    \hat{\mathcal{L}}(\boldsymbol{\theta})
    = \frac{1}{NT}
    \left\|
        \mathbf{Y}_+
        - \mathbf{\Phi}_{T}(\boldsymbol{\theta})\mathbf{Z}
    \right\|_F^2 
    = \frac{1}{NT}
    \left\|
        \operatorname{vec}(\mathbf{Y}_+)
        - (\mathbf{Z}^{\top} \otimes \mathbf{I}_m)
        \operatorname{vec}(\mathbf{\Phi}_{T}(\boldsymbol{\theta}))
    \right\|_2^2,
\end{equation}
by \eqref{equivalentrepresentation}. Since $\boldsymbol{\theta} \mapsto \mathbf{\Phi}_T(\boldsymbol{\theta})$
is surjective when $d \geq mT$, the optimization objective $\min_{\boldsymbol{\theta}\in \Theta} \hat{\mathcal{L}}(\boldsymbol{\theta})$ is equivalent to
\begin{equation}
    \min_{\mathbf{\Phi} \in \mathbb{R}^{m \times mT}}\left \{f(\mathbf{\Phi}):=
        \frac{1}{NT}
        \|\mathbf{Y}_+ - \mathbf{\Phi}\mathbf{Z}\|_F^2\right\},
    \label{eq:lemma3-ls-reduction}
\end{equation}
in the sense that for every $\boldsymbol{\theta}^* \in \Omega$, $\mathbf{\Phi}_T(\boldsymbol{\theta}^*)$ minimizes $f$. Furthermore, for every $\mathbf{\Phi}^*$ that minimizes $f$, there exists $\boldsymbol{\theta}^*$ such that $\mathbf{\Phi}_T(\boldsymbol{\theta}^*) = \mathbf{\Phi}^*$ due to the surjectivity of $\boldsymbol{\theta} \mapsto \mathbf{\Phi}_T(\boldsymbol{\theta})$.

Note that when $d \geq mT$, $f$ is a differentiable convex quadratic with gradient
\begin{equation}
   \nabla_{\mathbf{\Phi}} f(\mathbf{\Phi})
    = \frac{2}{NT}\bigl(\mathbf{\Phi}\mathbf{Z}\mathbf{Z}^{\top}-\mathbf{Y}_+\mathbf{Z}^{\top}\bigr). 
\end{equation}
Hence, $\mathbf{\Phi}^*$ minimizes $f$ if and only if it satisfies
the normal equation
\begin{equation}
\mathbf{\Phi}^*\mathbf{Z}\mathbf{Z}^{\top}
    = \mathbf{Y}_+\mathbf{Z}^{\top}.
    \label{eq:lemma3-normal-equation}
\end{equation}
The equation $\mathbf{Y}_+=\mathbf{\Phi}\mathbf{Z}$ has an exact
solution precisely when
$\operatorname{ran}(\mathbf{Y}_+^{\top})
\subseteq\operatorname{ran}(\mathbf{Z}^{\top})$, as shown in \Cref{columnspace}. In that case, the optimal training loss is zero. Otherwise,
\eqref{eq:lemma3-normal-equation} gives the least-squares solutions. When $\mathbf{Z}\mathbf{Z}^{\top}$ is nonsingular, 
the right multiplication of \eqref{eq:lemma3-normal-equation} by its inverse
shows that the unique minimizer of $f$ is
\begin{equation}
    \mathbf{\Phi}^*
    = \mathbf{Y}_+\mathbf{Z}^{\top}
      (\mathbf{Z}\mathbf{Z}^{\top})^{-1}.
\end{equation}
Therefore, every $\boldsymbol{\theta}^* \in \Omega$ must satisfy
\begin{equation}
    \mathbf{\Phi}_T(\boldsymbol{\theta}^*)
    = \mathbf{Y}_+\mathbf{Z}^{\top}
      (\mathbf{Z}\mathbf{Z}^{\top})^{-1}.
    \label{eq:lemma3-unique-markov}
\end{equation}
\end{proof}
\section{Proof of \Cref{thm:arbitrary-charpoly}}
\label[appendix]{appendix:proof-arbitrary-charpoly}

\begin{proof}
Let $\{a_0, a_1, \ldots, a_{d-1}\}$ be an arbitrary set of real coefficients. Moreover, let $\mathbf{\Phi}^* = [\mathbf{M}_0^* \ \cdots \ \mathbf{M}_{T-1}^*] \in \mathbb{R}^{m \times mT}$ be a minimizer of $f$ defined in \eqref{eq:lemma3-ls-reduction}, where $\mathbf{M}_k^* \in \mathbb{R}^{m \times m}$ for every $k \in [T-1]_0$. As we showed in the proof of \Cref{lemma:uniqueness}, when $d \geq mT$, every $\boldsymbol{\theta} \in \Theta$ that satisfies $\mathbf{\Phi}_T(\boldsymbol{\theta}) = \mathbf{\Phi}^*$ is a global minimizer of $\hat{\mathcal{L}}$. Therefore, to show that there exists a global minimizer whose recurrent weight matrix has a characteristic polynomial with arbitrary coefficients, it suffices to construct a parameterization $\Tilde{\boldsymbol{\theta}} = (\Tilde{\mathbf{W}}_1, \Tilde{\mathbf{W}}_2, \Tilde{\mathbf{W}}_3) \in \Theta$ such that $\mathbf{\Phi}_T(\Tilde{\boldsymbol{\theta}}) = \mathbf{\Phi}^*$ and the coefficients of the characteristic polynomial of $\Tilde{\mathbf{W}}_1$ are $\{a_0, a_1, \ldots, a_{d-1}\}$.

First, we denote the sets of standard basis vectors of $\mathbb{R}^T$, $\mathbb{R}^m$, and $\mathbb{R}^d$ by $\{\mathbf{e}_1,\ldots,\mathbf{e}_T\}$, $\{\mathbf{v}_1,\ldots,\mathbf{v}_m\}$, and $\{\mathbf{u}_1,\ldots,\mathbf{u}_d\}$, respectively. Now, for every $k \in [T]$, we define
\begin{equation}
    \mathbf{E}_k := \begin{bmatrix}
        \mathbf{I}_m \otimes \mathbf{e}_k \\
        \mathbf{0}_{r \times m}
    \end{bmatrix} \in \mathbb{R}^{d \times m},
    \label{eq:thm-charpoly-Ek}
\end{equation}
where $r := d - mT$, and choose a particular parameterization $\Tilde{\boldsymbol{\theta}} \in \Theta$ such that
\begin{equation}
    \Tilde{\mathbf{W}}_1 = \mathbf{L}_d - \mathbf{a}\mathbf{u}_d^\top
    = \begin{bmatrix}
        0 & 0 & \cdots & 0 & -a_0 \\
        1 & 0 & \cdots & 0 & -a_1 \\
        0 & 1 & \cdots & 0 & -a_2 \\
        \vdots & \vdots & \ddots & \vdots & \vdots \\
        0 & 0 & \cdots & 1 & -a_{d-1}
    \end{bmatrix},
    \qquad
    \Tilde{\mathbf{W}}_2 = \mathbf{E}_1,
    \qquad
    \Tilde{\mathbf{W}}_3 = \sum_{k=0}^{T-1} \mathbf{M}_k^* \mathbf{E}_{k+1}^\top,
    \label{eq:thm-charpoly-realization}
\end{equation}
where $\mathbf{L}_d \in \mathbb{R}^{d \times d}$ is a lower shift matrix and $\mathbf{a} := [a_0 \ a_1 \ \cdots \ a_{d-1}]^\top \in \mathbb{R}^d$. Note that $\Tilde{\mathbf{W}}_1$ is a companion matrix. Therefore, its characteristic polynomial is given by
\begin{equation}
    \det(s\mathbf{I}_d - \Tilde{\mathbf{W}}_1) = s^d + \sum_{j=0}^{d-1} a_j s^j
\end{equation}
\citep{kailath1980linear,horn2012matrix}. In particular, when $\mathbf{a} = \mathbf{0}$, $\Tilde{\mathbf{W}}_1$ reduces to a lower shift matrix.

Now, we show that $\mathbf{\Phi}_T(\Tilde{\boldsymbol{\theta}}) = \mathbf{\Phi}^*$. As we showed in \Cref{lowershiftmatrixremark1}, $\mathbf{L}_d\mathbf{u}_i = \mathbf{u}_{i+1}$ for all $i \in [d-1]$. Since $\mathbf{u}_d^\top\mathbf{u}_i = 0$ for all $i \in [d-1]$, we have
\begin{equation}
    \Tilde{\mathbf{W}}_1\mathbf{u}_i = \mathbf{u}_{i+1}
    \label{eq:thm-charpoly-shift}
\end{equation}
for all $i \in [d-1]$. This means that the last column of $\Tilde{\mathbf{W}}_1$ is not involved unless $\Tilde{\mathbf{W}}_1$ is applied to $\mathbf{u}_d$. By \eqref{eq:thm-charpoly-Ek}, the $j$th column of $\mathbf{E}_k$ is
\begin{equation}
\label{shiftingpr}
    \mathbf{E}_k\mathbf{v}_j
    = \begin{bmatrix}
        \mathbf{v}_j \otimes \mathbf{e}_k \\
        \mathbf{0}_{r}
    \end{bmatrix}
    = \mathbf{u}_{(j-1)T+k}
\end{equation}
for every $j \in [m]$ and $k \in [T]$. Furthermore, for every $k \in [T-1]$, we have $(j-1)T + k \leq mT - 1 \leq d-1$. Hence, by \eqref{eq:thm-charpoly-shift}, we have
\begin{equation}
    \Tilde{\mathbf{W}}_1\mathbf{E}_k\mathbf{v}_j
    = \mathbf{u}_{(j-1)T+k+1}
\end{equation}
for every $j \in [m]$. Moreover, by leveraging \eqref{shiftingpr}, we obtain 
\begin{equation}
     \mathbf{E}_{k+1}\mathbf{v}_j = \mathbf{u}_{(j-1)T+k+1} = \Tilde{\mathbf{W}}_1\mathbf{E}_k\mathbf{v}_j
\end{equation}
for every $j \in [m]$. This means that $\Tilde{\mathbf{W}}_1\mathbf{E}_k = \mathbf{E}_{k+1}$ for every $k \in [T-1]$. Recall that $\Tilde{\mathbf{W}}_2  := \mathbf{E}_1$. Then, by induction,
\begin{equation}
    \Tilde{\mathbf{W}}_1^{k}\Tilde{\mathbf{W}}_2 = \mathbf{E}_{k+1}
\end{equation}
for every $k \in [T-1]_0$. Using the mixed-product property of the Kronecker product, we show that
\begin{equation}
    \mathbf{E}_k^\top \mathbf{E}_\ell
    = (\mathbf{I}_m \otimes \mathbf{e}_k^\top)(\mathbf{I}_m \otimes \mathbf{e}_\ell)
    = \mathbf{I}_m \otimes (\mathbf{e}_k^\top\mathbf{e}_\ell)
    =
    \begin{cases}
        \mathbf{I}_m, & k = \ell, \\
        \mathbf{0}, & k \neq \ell,
    \end{cases}
    \label{eq:thm-charpoly-orthogonality}
\end{equation}
for every $k, \ell \in [T]$. Consequently,
\begin{equation}
    \Tilde{\mathbf{W}}_3 \Tilde{\mathbf{W}}_1^{k} \Tilde{\mathbf{W}}_2
    = \sum_{\ell=0}^{T-1} \mathbf{M}_\ell^* \mathbf{E}_{\ell+1}^\top\mathbf{E}_{k+1}
    = \mathbf{M}_k^*
\end{equation}
for every $k \in [T-1]_0$. Therefore, $\mathbf{\Phi}_T(\Tilde{\boldsymbol{\theta}}) = \mathbf{\Phi}^*$, and hence $\Tilde{\boldsymbol{\theta}} \in \Omega$.

\end{proof}

\section{Proof of \Cref{thm:exact-sharpness}}
\label[appendix]{appendix:proof-exact-sharpness}

\begin{proof}
To simplify subsequent derivations, we define the vector of all trainable parameters as $\boldsymbol{w}:=[\operatorname{vec}(\mathbf{W}_1)^\top \: \: \operatorname{vec}(\mathbf{W}_2)^\top \: \: \operatorname{vec}(\mathbf{W}_3)^\top]^\top \in \mathbb{R}^{d^2+2dm}$.
Furthermore, we showed in \Cref{lemma:uniqueness} that $\boldsymbol{\theta} \mapsto \mathbf{\Phi}_{T}(\boldsymbol{\theta})$ is surjective when $d \geq mT$. Therefore, the optimization objective $\min_{\boldsymbol{\theta}} \hat{\mathcal{L}}(\boldsymbol{\theta})$ is equivalent to 
\begin{equation}
    \min_{\mathbf{\Phi}\in \mathbb{R}^{m \times mT}} \frac{1}{NT}\|\mathbf{Y}_{+} - \mathbf{\Phi}\mathbf{Z}\|_F^2
\end{equation}
in the sense that for any $\boldsymbol{\theta}^* \in \Omega$, there exists a minimizer $\mathbf{\Phi}^* \in \argmin_{\mathbf{\Phi}\in \mathbb{R}^{m \times mT}} \|\mathbf{Y}_{+} - \mathbf{\Phi}\mathbf{Z}\|_F^2$ such that $\mathbf{\Phi}^* = \mathbf{\Phi}_T(\boldsymbol{\theta^*})$; and for any $\mathbf{\Phi}^* \in \argmin_{\mathbf{\Phi}\in \mathbb{R}^{m \times mT}} \|\mathbf{Y}_{+} - \mathbf{\Phi}\mathbf{Z}\|_F^2$, there exists a $\boldsymbol{\theta}^* \in \Omega$ such that $\mathbf{\Phi}^* = \mathbf{\Phi}_T(\boldsymbol{\theta}^*)$. Therefore, for any $\boldsymbol{\theta}^* \in \Omega$

\begin{equation}
\mathbf{\Phi}_T(\boldsymbol{\theta}^*)\mathbf{Z}\mathbf{Z}^\top-\mathbf{Y}_+\mathbf{Z}^\top = \mathbf{0},
\label{eq:markov-normal-equation}
\end{equation}
by the convexity of the Frobenius norm.  

As we showed in \Cref{remark:directional-taylor},
\begin{equation}
        \lambda_{\max}\!\left(
        \nabla^2
        \mathcal{L}(\mathbf{X})
    \right)
    =
    \max_{\|\mathbf{U}\|_F=1}
    D_{\mathbf{U}}^2\mathcal{L}(\mathbf{X}).
\end{equation}
Hence, let us define an arbitrary perturbation $\mathbf{U}
    :=
    (\mathbf{U}_1,\mathbf{U}_2,\mathbf{U}_3)$ in the parameter space $\Theta$, where $
    \mathbf{U}_1\in\mathbb{R}^{d\times d},
    \mathbf{U}_2\in\mathbb{R}^{d\times m}$ and $
    \mathbf{U}_3\in\mathbb{R}^{m\times d}$. Furthermore, we also define its vectorized form as $\mathbf{u}:= [\operatorname{vec}(\mathbf{U}_1)^\top \: \: \operatorname{vec}(\mathbf{U}_2)^\top \: \: \operatorname{vec}(\mathbf{U}_3)^\top]^\top \in \mathbb{R}^{d^2+2md}$. For any $\boldsymbol{\theta}^* \in \Omega$, we can define the parameterized path along the perturbation direction $\mathbf{U}$ as $\boldsymbol{\theta}^*(t)
    :=
    \left(
        \mathbf{W}_1^*+t\mathbf{U}_1,\,
        \mathbf{W}_2^*+t\mathbf{U}_2,\,
        \mathbf{W}_3^*+t\mathbf{U}_3
    \right).$
Then, as we showed in (\ref{Gateaux}) and (\ref{Gateaux-2}), the first and second directional derivatives are defined by 
\begin{equation}
    \dot{\mathbf{\Phi}}_{\mathbf{U}}
    :=
    \frac{\partial}{\partial t}
    \mathbf{\Phi}_T(\boldsymbol{\theta}^*(t))
    \biggr|_{t=0} \quad \text{and} \quad
    \ddot{\mathbf{\Phi}}_{\mathbf{U}}
    :=
    \frac{\partial^2}{\partial t^2}
    \mathbf{\Phi}_T(\boldsymbol{\theta}^*(t))
    \biggr|_{t=0},
    \label{eq:phi-second-path-derivative}
\end{equation}
respectively. Leveraging \Cref{second-directional}, for any $\boldsymbol{\theta}^* \in \Omega$, 
\begin{align}
    D_{\mathbf{U}}^2
    \hat{\mathcal{L}}(\boldsymbol{\theta}^*)
    &= \frac{\partial^2}{\partial t^2} \hat{\mathcal{L}}\left(\boldsymbol{\theta}^{*} (t)\right) \Big |_{t = 0} \\ 
    &= \frac{1}{NT}\frac{\partial^2}{\partial t^2} \left \|\mathbf{Y}_+ - \mathbf{\Phi}_T(\boldsymbol{\theta}^*(t))\mathbf{Z} \right \|_F^2 \Big |_{t = 0} \\
    &= \frac{1}{NT}\frac{\partial^2}{\partial t^2} \operatorname{tr}\left(\left(\mathbf{Y}_+ - \mathbf{\Phi}_T\left(\boldsymbol{\theta}^*\left(t\right)\right)\mathbf{Z}\right)^\top\left(\mathbf{Y}_+ - \mathbf{\Phi}_T\left(\boldsymbol{\theta}^*\left(t\right)\right)\mathbf{Z}\right)\right) \Big |_{t = 0} \\ 
    &=\frac{1}{NT}\frac{\partial^2}{\partial t^2} \left(\| \mathbf{Y}_+\|_F^2 - 2\operatorname{tr}\left(\mathbf{Y}_+^\top\mathbf{\Phi}_T\left(\boldsymbol{\theta}^*\left(t\right)\right)\mathbf{Z}\right) + \operatorname{tr}\left(\mathbf{Z}^\top\mathbf{\Phi}^\top_T\left(\boldsymbol{\theta}^*(t)\right)\mathbf{\Phi}_T\left(\boldsymbol{\theta}^*(t)\right)\mathbf{Z}\right) \right)\Big |_{t = 0} \\
    &= -\frac{2}{NT}\operatorname{tr}\left(\mathbf{Y}_+^\top\ddot{\mathbf{\Phi}}_{\mathbf{U}}\mathbf{Z}\right) +\frac{1}{NT}\operatorname{tr}\left(\mathbf{Z}^\top\frac{\partial^2}{\partial t^2}\left (\mathbf{\Phi}^\top_T\left(\boldsymbol{\theta}^*(t)\right)\mathbf{\Phi}_T\left(\boldsymbol{\theta}^*(t)\right)\right)\Big |_{t = 0}\mathbf{Z}\right),
\end{align}
by using the identity $\|\mathbf{A}\|_F^2 = \operatorname{tr}(\mathbf{A}^\top \mathbf{A})$, \Cref{diffrule}, and  the fact that differentiation and trace operations commute. Then, by \Cref{diffchainrule}, we have
\begin{align}
    \frac{\partial}{\partial t}\left \{\mathbf{\Phi}^\top_T\left(\boldsymbol{\theta}^*(t)\right)\mathbf{\Phi}_T\left(\boldsymbol{\theta}^*(t)\right)\right\} &= \left (\frac{\partial\mathbf{\Phi}_T\left(\boldsymbol{\theta}^*(t)\right)}{\partial t} \right)^\top \mathbf{\Phi}_T\left(\boldsymbol{\theta}^*(t)\right) + \mathbf{\Phi}^\top_T\left(\boldsymbol{\theta}^*(t)\right)\frac{\partial\mathbf{\Phi}_T\left(\boldsymbol{\theta}^*(t)\right)}{\partial t}, \\
    \frac{\partial^2}{\partial t^2}\left \{\mathbf{\Phi}^\top_T\left(\boldsymbol{\theta}^*(t)\right)\mathbf{\Phi}_T\left(\boldsymbol{\theta}^*(t)\right)\right\} &= \left (\frac{\partial^2\mathbf{\Phi}_T\left(\boldsymbol{\theta}^*(t)\right)}{\partial t^2} \right)^\top \mathbf{\Phi}_T\left(\boldsymbol{\theta}^*(t)\right) \\ &+ 2\left (\frac{\partial \mathbf{\Phi}_T\left(\boldsymbol{\theta}^*(t)\right)}{\partial t} \right)^\top \frac{\partial \mathbf{\Phi}_T\left(\boldsymbol{\theta}^*(t)\right)}{\partial t} +  \mathbf{\Phi}^\top_T\left(\boldsymbol{\theta}^*(t)\right)\frac{\partial^2\mathbf{\Phi}_T\left(\boldsymbol{\theta}^*(t)\right)}{\partial t^2}.
\end{align}
Therefore, 
\begin{equation}
    \frac{\partial^2}{\partial t^2}\left \{\mathbf{\Phi}^\top_T\left(\boldsymbol{\theta}^*(t)\right)\mathbf{\Phi}_T\left(\boldsymbol{\theta}^*(t)\right)\right\}\Big |_{t = 0} = \ddot{\mathbf{\Phi}}_{\mathbf{U}}^\top \mathbf{\Phi}_T(\boldsymbol{\theta^*})+ 2\dot{\mathbf{\Phi}}_{\mathbf{U}}^\top\dot{\mathbf{\Phi}}_{\mathbf{U}} + \mathbf{\Phi}^\top_T(\boldsymbol{\theta^*})\ddot{\mathbf{\Phi}}_{\mathbf{U}}. 
\end{equation}
Consequently, 
\begin{align}
    D_{\mathbf{U}}^2
    \hat{\mathcal{L}}(\boldsymbol{\theta}^*) &= -\frac{2}{NT}\operatorname{tr}\left(\mathbf{Y}_+^\top\ddot{\mathbf{\Phi}}_{\mathbf{U}}\mathbf{Z}\right) +\frac{1}{NT}\operatorname{tr}\left(\mathbf{Z}^\top(\ddot{\mathbf{\Phi}}_{\mathbf{U}}^\top \mathbf{\Phi}_T(\boldsymbol{\theta^*})+ 2\dot{\mathbf{\Phi}}_{\mathbf{U}}^\top\dot{\mathbf{\Phi}}_{\mathbf{U}} + \mathbf{\Phi}^\top_T(\boldsymbol{\theta^*})\ddot{\mathbf{\Phi}}_{\mathbf{U}})\mathbf{Z}\right) \\ 
    &= -\frac{2}{NT}\operatorname{tr}\left(\mathbf{Y}_+^\top\ddot{\mathbf{\Phi}}_{\mathbf{U}}\mathbf{Z}\right) + \frac{2}{NT}\operatorname{tr}\left(\mathbf{Z}^\top\ddot{\mathbf{\Phi}}_{\mathbf{U}}^\top \mathbf{\Phi}_T(\boldsymbol{\theta^*})\mathbf{Z}\right) + \frac{2}{NT}\operatorname{tr}\left(\mathbf{Z}^\top(\dot{\mathbf{\Phi}}_{\mathbf{U}}^\top\dot{\mathbf{\Phi}}_{\mathbf{U}})\mathbf{Z}\right) \\ 
    &= \frac{2}{NT}\|\dot{\mathbf{\Phi}}_{\mathbf{U}}\mathbf{Z}\|_F^2 - \frac{2}{NT}\operatorname{tr}\left(\mathbf{Z}(\mathbf{Y}_+^\top - \mathbf{Z}^\top \mathbf{\Phi}_T^\top(\boldsymbol{\theta}^*))\ddot{\mathbf{\Phi}}_{\mathbf{U}}\right), \label{secondordersimp}
\end{align}
by using the identities $\operatorname{tr}(\mathbf{A}^\top \mathbf{B}) = \operatorname{tr}(\mathbf{B}^\top \mathbf{A})$ and the fact that the trace operator is linear and invariant under cyclic shifts. As we showed in (\ref{eq:markov-normal-equation}), $\mathbf{Z}(\mathbf{Y}_+^\top - \mathbf{Z}^\top \mathbf{\Phi}_T^\top(\boldsymbol{\theta}^*)) = \mathbf{0}$ due to the convexity of Frobenius norm. Therefore, the second term on the right-hand side of \eqref{secondordersimp} vanishes
exactly. Consequently, the second directional derivative of $\hat{\mathcal{L}}$ at any minimizer $\boldsymbol{\theta}^* \in \Omega$ along $\mathbf{U}$ simplifies to 
\begin{equation}
     D_{\mathbf{U}}^2
    \hat{\mathcal{L}}(\boldsymbol{\theta}^*) = \frac{2}{NT}\|\dot{\mathbf{\Phi}}_{\mathbf{U}}\mathbf{Z}\|_F^2.
\end{equation}
Moreover,
\begin{equation}
    D_{\mathbf{U}}^2
    \hat{\mathcal{L}}(\boldsymbol{\theta}^*) = \frac{2}{NT} \|\operatorname{vec}(\dot{\mathbf{\Phi}}_{\mathbf{U}}\mathbf{Z})\|_2^2 = \frac{2}{NT} \|(\mathbf{Z}^\top \otimes \mathbf{I})\operatorname{vec}
        (\dot{\mathbf{\Phi}}_{\mathbf{U}})\|_2^2,
\label{eq:phi-directional-jacobian}
\end{equation}
by using the identity $\operatorname{vec}(\dot{\mathbf{\Phi}}_{\mathbf{U}}\mathbf{Z})
    =(\mathbf{Z}^\top\otimes\mathbf{I})
    \operatorname{vec}(
        \dot{\mathbf{\Phi}}_{\mathbf{U}})$.

As shown in \eqref{eq:vec-first-directional}, the vectorized, first-order directional derivative of a continuously differentiable matrix-valued function of a matrix variable, which is taken in the direction $\operatorname{vec}(\mathbf{U})$, can be expressed in terms of its Jacobian. Consequently,
\begin{equation}
       \operatorname{vec}
        (\dot{\mathbf{\Phi}}_{\mathbf{U}}) =
    \mathbf{K}^*\mathbf{u},
\end{equation}
where $\mathbf{K}^*$ is the Jacobian of $\operatorname{vec}(\mathbf{\Phi}_T(\boldsymbol{\theta}))$ with respect to $\operatorname{vec}(\boldsymbol{\theta})$ calculated at $\boldsymbol{\theta}^* \in \Omega$, i.e., $\mathbf{K}^* = \frac{\partial}{\partial \operatorname{vec}(\boldsymbol{\theta})}\operatorname{vec}(\mathbf{\Phi}_T(\boldsymbol{\theta}))\Big |_{\boldsymbol{\theta} = \boldsymbol{\theta^*}}$. Hence, 
\begin{align}
    D_{\mathbf{U}}^2
    \hat{\mathcal{L}}(\boldsymbol{\theta}^*)
    &=
   \frac{2}{NT} \mathbf{u}^\top
    \mathbf{K}^{*\top}
        (\mathbf{Z}\otimes\mathbf{I})
        (\mathbf{Z}^\top\otimes\mathbf{I})
    \mathbf{K}^*
    \mathbf{u} \\
    &=
    \frac{2}{NT}\mathbf{u}^\top
    \mathbf{K}^{*\top}
    (\mathbf{Z}\mathbf{Z}^\top
        \otimes
        \mathbf{I})
    \mathbf{K}^*
    \mathbf{u} \\ 
    &= 2\mathbf{u}^\top
    \mathbf{K}^{*\top}
    (\widehat{\boldsymbol{\Sigma}}_T
        \otimes
        \mathbf{I})
    \mathbf{K}^*
    \mathbf{u}.
\label{eq:directional-output-norm}
\end{align}
Note that, by
\eqref{eq:second-directional-hessian}, $D_{\mathbf{U}}^2
    \hat{\mathcal{L}}(\boldsymbol{\theta}^*)
    =
    \mathbf{u}^\top
    \nabla_{\operatorname{vec}(\boldsymbol{\theta})}^2
    \hat{\mathcal{L}}(\boldsymbol{\theta}^*)
    \mathbf{u}$ for every $\mathbf{u} \in \mathbb{R}^{d^2+2md}$. Since \eqref{eq:directional-output-norm} holds for every $\mathbf{u} \in \mathbb{R}^{d^2+2md}$, we conclude that
\begin{equation}
    \nabla_{\operatorname{vec}(\boldsymbol{\theta})}^2
    \hat{\mathcal{L}}(\boldsymbol{\theta}^*)
    =
    2\mathbf{K}^{*\top}
    \left(
        \widehat{\boldsymbol{\Sigma}}_T
        \otimes
        \mathbf{I}
    \right)
    \mathbf{K}^*,
\label{eq:exact-rnn-hessian-proof}
\end{equation}
where $\lambda_{\max}(\nabla_{\operatorname{vec}(\boldsymbol{\theta})}^2
    \hat{\mathcal{L}}(\boldsymbol{\theta})) = \lambda_{\max}(\nabla_{\boldsymbol{\theta}}^2
    \hat{\mathcal{L}}(\boldsymbol{\theta}))$ for all $\boldsymbol{\theta} \in \Theta$. This implies that for any $\boldsymbol{\theta}^* \in \Omega$, 
\begin{equation}
    \lambda_{\max}\!\left(
        \nabla_{\operatorname{vec}(\boldsymbol{\theta})}^2
        \hat{\mathcal{L}}(\boldsymbol{\theta}^*)
    \right)
    =
    2\lambda_{\max}\!\left(
        \mathbf{K}^{*\top}
        \left(
            \widehat{\boldsymbol{\Sigma}}_T
            \otimes
            \mathbf{I}
        \right)
        \mathbf{K}^*
    \right) 
    =
    2\sigma_{\max}^2\!\left(
        \left(
            \widehat{\boldsymbol{\Sigma}}_T^{1/2}
            \otimes
            \mathbf{I}
        \right)
        \mathbf{K}^*
    \right).
\label{eq:sharpness-final-proof}
\end{equation}

As we defined in \Cref{remark:representation}, $\mathbf{M}_k(\boldsymbol{\theta}) :=
    \mathbf{W}_3\mathbf{W}_1^k\mathbf{W}_2$ for all $k \in [T-1]_0$. By using \eqref{firstorderdirecformulaformatrix}, the first directional derivative of $\mathbf{M}_k(\boldsymbol{\theta})$ along $\mathbf{U}$ is \begin{equation}
    D_{\mathbf{U}}\mathbf{M}_k(\boldsymbol{\theta})
    =
    \mathbf{U}_3\mathbf{W}_1^k\mathbf{W}_2 + \mathbf{W}_3\mathbf{W}_1^k\mathbf{U}_2 +  \sum_{\ell=0}^{k-1}
    \mathbf{W}_3\mathbf{W}_1^\ell
    \mathbf{U}_1
    \mathbf{W}_1^{k-1-\ell}\mathbf{W}_2.
\label{eq:markov-directional-derivative}
\end{equation}
If we vectorize the first directional derivative of $\mathbf{M}_k(\boldsymbol{\theta})$ along $\mathbf{U}$, we obtain
\begin{equation}
\begin{aligned}
    \operatorname{vec}
    \left(
        D_{\mathbf{U}}\mathbf{M}_k(\boldsymbol{\theta})
    \right)
    ={}&
    \mathbf{K}_{1,k}
    \operatorname{vec}(\mathbf{U}_1)
    +
    \mathbf{K}_{2,k}
    \operatorname{vec}(\mathbf{U}_2)
    +
    \mathbf{K}_{3,k}
    \operatorname{vec}(\mathbf{U}_3), 
\end{aligned}
\label{eq:markov-vectorized-directional}
\end{equation}
by using the identity $\operatorname{vec}(\mathbf{A}\mathbf{X}\mathbf{B})
    = (\mathbf{B}^\top\otimes\mathbf{A})
    \operatorname{vec}(\mathbf{X})$, where
\begin{equation}
    \mathbf{K}_{1,k}
    =
    \sum_{\ell=0}^{k-1}
    \left[
        \left(
            \mathbf{W}_1^{k-1-\ell}\mathbf{W}_2
        \right)^\top
        \otimes
        \mathbf{W}_3\mathbf{W}_1^\ell
    \right],
    \mathbf{K}_{2,k}
    =
    \mathbf{I}
    \otimes
    \mathbf{W}_3\mathbf{W}_1^k,
    \mathbf{K}_{3,k}
    =
    \left(
        \mathbf{W}_1^k\mathbf{W}_2
    \right)^\top
    \otimes
    \mathbf{I}.
    \end{equation}
Note that $\mathbf{K}^* = \frac{\partial}{\partial \operatorname{vec}(\boldsymbol{\theta})}\operatorname{vec}(\mathbf{\Phi}_T(\boldsymbol{\theta}))\Big |_{\boldsymbol{\theta} = \boldsymbol{\theta^*}}$ such that

\begin{equation}
    \mathbf{K} = \frac{\partial\operatorname{vec}(\mathbf{\Phi}_T(\boldsymbol{\theta}))}{\partial \operatorname{vec}(\boldsymbol{\theta})}= \begin{bmatrix}
        \frac{\partial\operatorname{vec}(\mathbf{M}_0(\boldsymbol{\theta}))}{\partial \operatorname{vec}(\boldsymbol{\theta})} \\ \frac{\partial\operatorname{vec}(\mathbf{M}_1(\boldsymbol{\theta}))}{\partial \operatorname{vec}(\boldsymbol{\theta})} \\ \vdots \\\frac{\partial\operatorname{vec}(\mathbf{M}_{T-1}(\boldsymbol{\theta}))}{\partial \operatorname{vec}(\boldsymbol{\theta})}
    \end{bmatrix}.
\end{equation}
As we showed in \Cref{remark:matrix-valued-taylor}, 
\begin{equation}
     \operatorname{vec}
    \left(
        D_{\mathbf{U}}\mathbf{M}_k(\boldsymbol{\theta})
    \right) = \frac{\partial\operatorname{vec}(\mathbf{M}_k(\boldsymbol{\theta}))}{\partial \operatorname{vec}(\boldsymbol{\theta})} \mathbf{u}.
\end{equation}
Therefore, if we stack $\mathbf{K}_{i,k}$ vertically 
for every $k \in [T-1]_0$ and repeat it for every $i \in \{1,2,3\}$, we can form $\mathbf{K}$. Then, evaluating at $\boldsymbol{\theta}=\boldsymbol{\theta}^*$ gives
\begin{equation}
    \mathbf{K}^*
    =
    \begin{bmatrix}
        \mathbf{K}_{1,0}^* &
        \mathbf{K}_{2,0}^* &
        \mathbf{K}_{3,0}^*\\
        \mathbf{K}_{1,1}^* &
        \mathbf{K}_{2,1}^* &
        \mathbf{K}_{3,1}^*\\
        \vdots & \vdots & \vdots\\
        \mathbf{K}_{1,T-1}^* &
        \mathbf{K}_{2,T-1}^* &
        \mathbf{K}_{3,T-1}^*
    \end{bmatrix},
\label{eq:Kstar-proof}
\end{equation}
with $\mathbf{K}_{1,0}^*=\mathbf{0}$.
\end{proof}

\section{Proof of \Cref{prop:rescaling}}
\label[appendix]{appendix:proof-rescaling}

\begin{proof}
Let $\boldsymbol{\theta}^* = (\mathbf{W}_1^*, \mathbf{W}_2^*, \mathbf{W}_3^*) \in \Omega$ and $c \in \mathbb{R}\setminus\{0\}$. Note that $\boldsymbol{\theta}^*_c = \boldsymbol{\theta}^*_{\mathbf{S}}$ with $\mathbf{S} = c\mathbf{I}_d \in \mathrm{GL}_d(\mathbb{R})$. Therefore, by \Cref{prop:similarity}, $\mathbf{M}_i(\boldsymbol{\theta}^*_c) = \mathbf{M}_i(\boldsymbol{\theta}^*)$ for every $i \geq 0$, and $\boldsymbol{\theta}^*_c \in \Omega$.

We now show that $\rho_{\text{co}}(\boldsymbol{\theta}^*_c) = \rho_{\text{co}}(\boldsymbol{\theta}^*)$. Since the recurrent weight matrix of $\boldsymbol{\theta}^*_c$ is $\mathbf{W}_1^*$, the two parameterizations have the same spectrum. Furthermore, for every $\lambda \in \operatorname{spec}(\mathbf{W}_1^*)$,
\begin{align}
    \begin{bmatrix} \lambda\mathbf{I}_d - \mathbf{W}_1^* & c\mathbf{W}_2^* \end{bmatrix}
    &=
    \begin{bmatrix} \lambda\mathbf{I}_d - \mathbf{W}_1^* & \mathbf{W}_2^* \end{bmatrix}
    \begin{bmatrix} \mathbf{I}_d & \mathbf{0} \\ \mathbf{0} & c\mathbf{I}_m \end{bmatrix}, \\
    \begin{bmatrix} \lambda\mathbf{I}_d - \mathbf{W}_1^* \\ c^{-1}\mathbf{W}_3^* \end{bmatrix}
    &=
    \begin{bmatrix} \mathbf{I}_d & \mathbf{0} \\ \mathbf{0} & c^{-1}\mathbf{I}_m \end{bmatrix}
    \begin{bmatrix} \lambda\mathbf{I}_d - \mathbf{W}_1^* \\ \mathbf{W}_3^* \end{bmatrix}.
\end{align}
Since the block-diagonal matrices on the right-hand sides have full rank, 

\begin{equation}
    \operatorname{rank}\left(\begin{bmatrix} \lambda\mathbf{I}_d - \mathbf{W}_1^* & c\mathbf{W}_2^* \end{bmatrix}\right)
    =
    \operatorname{rank}\left(\begin{bmatrix} \lambda\mathbf{I}_d - \mathbf{W}_1^* & \mathbf{W}_2^* \end{bmatrix}\right)
\end{equation}
and 
\begin{equation}
        \operatorname{rank}\left(\begin{bmatrix} \lambda\mathbf{I}_d - \mathbf{W}_1^* \\ c^{-1}\mathbf{W}_3^* \end{bmatrix}\right)
    =
    \operatorname{rank}\left(\begin{bmatrix} \lambda\mathbf{I}_d - \mathbf{W}_1^* \\ \mathbf{W}_3^* \end{bmatrix}\right).
\end{equation}
Hence, $\rho_{\text{co}}(\boldsymbol{\theta}^*_c) = \rho_{\text{co}}(\boldsymbol{\theta}^*)$.

Next, we compute the Jacobian of $\operatorname{vec}(\mathbf{\Phi}_T(\boldsymbol{\theta}))$ at $\boldsymbol{\theta}^*_c$. Since $\boldsymbol{\theta}^*_c \in \Omega$, \Cref{thm:exact-sharpness} gives its blocks as
\begin{alignat}{2}
    \sum_{\ell=0}^{k-1}
    \Bigl[
        \left(\mathbf{W}_1^{*\,k-1-\ell}\, c\mathbf{W}_2^*\right)^{\top}
        &\otimes
        c^{-1}\mathbf{W}_3^*\mathbf{W}_1^{*\,\ell}
    \Bigr]
    = \mathbf{K}^*_{1,k}, \\
    \mathbf{I}_m
    &\otimes c^{-1}\mathbf{W}_3^*\mathbf{W}_1^{*\,k}
    = c^{-1}\mathbf{K}^*_{2,k}, \\
    \left(\mathbf{W}_1^{*\,k}\, c\mathbf{W}_2^*\right)^{\top}
    &\otimes \mathbf{I}_m
    = c\,\mathbf{K}^*_{3,k}
\end{alignat}
for every $k \in [T-1]_0$. Hence, the Jacobian at $\boldsymbol{\theta}^*_c$ is $\mathbf{K}^*\mathbf{D}_c$, where $\mathbf{D}_c := \operatorname{diag}(\mathbf{I}_{d^2}, c^{-1}\mathbf{I}_{dm}, c\mathbf{I}_{dm})$. By \eqref{eq:exact-rnn-hessian-proof},
\begin{equation}
    \nabla^2_{\operatorname{vec}(\boldsymbol{\theta})}\hat{\mathcal{L}}(\boldsymbol{\theta}^*_c)
    =
    2\mathbf{D}_c\mathbf{K}^{*\top}
    \left(\widehat{\boldsymbol{\Sigma}}_T \otimes \mathbf{I}_m\right)
    \mathbf{K}^*\mathbf{D}_c.
    \label{eq:rescaled-hessian}
\end{equation}

Let $\mathbf{U}_3 \in \mathbb{R}^{m \times d}$ satisfy $\|\mathbf{U}_3\|_F = 1$, and define $\mathbf{u} := [\mathbf{0}^\top \ \ \mathbf{0}^\top \ \ \operatorname{vec}(\mathbf{U}_3)^\top]^\top$. Then, $\mathbf{D}_c\mathbf{u} = c\mathbf{u}$. Moreover, $\mathbf{K}^*_{3,k}\operatorname{vec}(\mathbf{U}_3) = \operatorname{vec}(\mathbf{U}_3\mathbf{W}_1^{*\,k}\mathbf{W}_2^*)$ for every $k \in [T-1]_0$ by the identity $\operatorname{vec}(\mathbf{A}\mathbf{X}\mathbf{B}) = (\mathbf{B}^\top \otimes \mathbf{A})\operatorname{vec}(\mathbf{X})$. Stacking these vectors vertically gives $\mathbf{K}^*\mathbf{u} = \operatorname{vec}(\mathbf{U}_3\mathcal{C}_T(\boldsymbol{\theta}^*))$. Therefore, by \eqref{eq:rescaled-hessian},
\begin{align}
    \mathbf{u}^\top
    \nabla^2_{\operatorname{vec}(\boldsymbol{\theta})}\hat{\mathcal{L}}(\boldsymbol{\theta}^*_c)
    \mathbf{u}
    &=
    2c^2
    \operatorname{vec}(\mathbf{U}_3\mathcal{C}_T(\boldsymbol{\theta}^*))^\top
    \left(\widehat{\boldsymbol{\Sigma}}_T \otimes \mathbf{I}_m\right)
    \operatorname{vec}(\mathbf{U}_3\mathcal{C}_T(\boldsymbol{\theta}^*)) \\
    &=
    2c^2
    \operatorname{tr}\left(
        \mathbf{U}_3\mathcal{C}_T(\boldsymbol{\theta}^*)
        \widehat{\boldsymbol{\Sigma}}_T
        \mathcal{C}_T(\boldsymbol{\theta}^*)^\top\mathbf{U}_3^\top
    \right) \\
    &\geq
    2c^2\lambda_{\min}(\widehat{\boldsymbol{\Sigma}}_T)
    \|\mathbf{U}_3\mathcal{C}_T(\boldsymbol{\theta}^*)\|_F^2,
\end{align}
where we use the identity $\operatorname{vec}(\mathbf{A})^\top(\mathbf{B} \otimes \mathbf{I}_m)\operatorname{vec}(\mathbf{A}) = \operatorname{tr}(\mathbf{A}\mathbf{B}\mathbf{A}^\top)$ for every symmetric matrix $\mathbf{B}$. Now, let $\mathbf{x} \in \mathbb{R}^d$ be a unit left singular vector of $\mathcal{C}_T(\boldsymbol{\theta}^*)$ associated with its largest singular value, let $\mathbf{a} \in \mathbb{R}^m$ be a unit vector, and choose $\mathbf{U}_3 = \mathbf{a}\mathbf{x}^\top$. Note that for any vectors $\mathbf{a}$ and $\mathbf{b}$, 
\begin{equation}
    \|\mathbf{a}\mathbf{b}^\top\|_F^2 = \operatorname{tr}(\mathbf{a}\mathbf{b}^\top\mathbf{b}\mathbf{a}^\top) = \operatorname{tr}(\mathbf{a}^\top\mathbf{a}\mathbf{b}^\top\mathbf{b}) = \|\mathbf{a}\|_2^2\|\mathbf{b}\|_2^2.
\end{equation}
Then, $\|\mathbf{U}_3\|_F = 1$ and $\|\mathbf{U}_3\mathcal{C}_T(\boldsymbol{\theta}^*)\|_F = \|\mathcal{C}_T(\boldsymbol{\theta}^*)^\top\mathbf{x}\|_2 = \sigma_{\max}(\mathcal{C}_T(\boldsymbol{\theta}^*))$. By the Rayleigh-quotient characterization in \Cref{remark:directional-taylor},
\begin{equation}
    \lambda_{\max}(\nabla^2\hat{\mathcal{L}}(\boldsymbol{\theta}^*_c))
    \geq
    2c^2\lambda_{\min}(\widehat{\boldsymbol{\Sigma}}_T)\,
    \sigma_{\max}^2(\mathcal{C}_T(\boldsymbol{\theta}^*)).
\end{equation}

Similarly, let $\mathbf{U}_2 \in \mathbb{R}^{d \times m}$ satisfy $\|\mathbf{U}_2\|_F = 1$, and define $\mathbf{u} := [\mathbf{0}^\top \ \ \operatorname{vec}(\mathbf{U}_2)^\top \ \ \mathbf{0}^\top]^\top$. Then, $\mathbf{D}_c\mathbf{u} = c^{-1}\mathbf{u}$, and $\mathbf{K}^*\mathbf{u} = \operatorname{vec}(\mathbf{\Gamma})$, where
\begin{equation}
    \mathbf{\Gamma} := \begin{bmatrix}
        \mathbf{W}_3^*\mathbf{U}_2 & \mathbf{W}_3^*\mathbf{W}_1^*\mathbf{U}_2 & \cdots & \mathbf{W}_3^*\mathbf{W}_1^{*\,T-1}\mathbf{U}_2
    \end{bmatrix} \in \mathbb{R}^{m \times mT}.
\end{equation}
By the same argument,
\begin{align}
    \mathbf{u}^\top
    \nabla^2_{\operatorname{vec}(\boldsymbol{\theta})}\hat{\mathcal{L}}(\boldsymbol{\theta}^*_c)
    \mathbf{u}
    &\geq
    2c^{-2}\lambda_{\min}(\widehat{\boldsymbol{\Sigma}}_T)\|\mathbf{\Gamma}\|_F^2 \\
    &=
    2c^{-2}\lambda_{\min}(\widehat{\boldsymbol{\Sigma}}_T)
    \sum_{k=0}^{T-1}\|\mathbf{W}_3^*\mathbf{W}_1^{*\,k}\mathbf{U}_2\|_F^2 \\
    &=
    2c^{-2}\lambda_{\min}(\widehat{\boldsymbol{\Sigma}}_T)
    \|\mathcal{O}_T(\boldsymbol{\theta}^*)\mathbf{U}_2\|_F^2.
\end{align}
Again, by the same argument, let $\mathbf{y} \in \mathbb{R}^d$ be a unit right singular vector of $\mathcal{O}_T(\boldsymbol{\theta}^*)$ associated with its largest singular value, and choose $\mathbf{U}_2 = \mathbf{y}\mathbf{a}^\top$. Then, $\|\mathbf{U}_2\|_F = 1$ and $\|\mathcal{O}_T(\boldsymbol{\theta}^*)\mathbf{U}_2\|_F = \sigma_{\max}(\mathcal{O}_T(\boldsymbol{\theta}^*))$. Therefore, 

\begin{equation}
    \lambda_{\max}(\nabla^2\hat{\mathcal{L}}(\boldsymbol{\theta}^*_c))
    \geq
    2\lambda_{\min}(\widehat{\boldsymbol{\Sigma}}_T)
    \max\left\{
        c^2\sigma_{\max}^2(\mathcal{C}_T(\boldsymbol{\theta}^*)),\,
        c^{-2}\sigma_{\max}^2(\mathcal{O}_T(\boldsymbol{\theta}^*))
    \right\}.
\end{equation}

\end{proof}

\section{Proof of \Cref{prop:unbounded-rho}}
\label[appendix]{appendix:proof-unbounded-rho}

\begin{proof}
Let $\lambda \in \mathbb{R}\setminus\{0\}$. Since $\widehat{\boldsymbol{\Sigma}}_T$ has full rank, $\mathbf{\Phi}^* = [\mathbf{M}_0^* \ \cdots \ \mathbf{M}_{T-1}^*] \in \mathbb{R}^{m \times mT}$ is the unique minimizer of $f$ defined in \eqref{eq:lemma3-ls-reduction}. As we showed in the proof of \Cref{lemma:uniqueness}, when $d \geq mT$, every $\boldsymbol{\theta} \in \Theta$ that satisfies $\mathbf{\Phi}_T(\boldsymbol{\theta}) = \mathbf{\Phi}^*$ is a global minimizer of $\hat{\mathcal{L}}$. Furthermore, we denote the standard basis vectors of $\mathbb{R}^T$ and $\mathbb{R}^m$ by $\{\mathbf{e}_1, \ldots, \mathbf{e}_T\}$ and $\{\mathbf{v}_1, \ldots, \mathbf{v}_m\}$, respectively. We consider the cases $d > mT$ and $d = mT$ separately.

\paragraph{Case $d > mT$.} Let $r := d - mT \geq 1$. Let us define
\begin{equation}
    \mathbf{\Psi} := \begin{bmatrix}
        \mathbf{v}_1\mathbf{v}_1^\top & \lambda\mathbf{v}_1\mathbf{v}_1^\top & \cdots & \lambda^{T-1}\mathbf{v}_1\mathbf{v}_1^\top
    \end{bmatrix} \in \mathbb{R}^{m \times mT},
\end{equation}
to choose a particular parameterization $\hat{\boldsymbol{\theta}} = (\widehat{\mathbf{W}}_1, \widehat{\mathbf{W}}_2, \widehat{\mathbf{W}}_3) \in \Theta$ such that
\begin{equation}
    \widehat{\mathbf{W}}_1 = \begin{bmatrix}
        \mathbf{L}_T \otimes \mathbf{I}_m & \mathbf{0} & \mathbf{0} \\
        \mathbf{0} & \lambda & \mathbf{0} \\
        \mathbf{0} & \mathbf{0} & \mathbf{0}_{(r-1)\times(r-1)}
    \end{bmatrix},
    \quad
    \widehat{\mathbf{W}}_2 = \begin{bmatrix}
        \mathbf{e}_1 \otimes \mathbf{I}_m \\
        \mathbf{v}_1^\top \\
        \mathbf{0}_{(r-1)\times m}
    \end{bmatrix},
    \quad
    \widehat{\mathbf{W}}_3 = \begin{bmatrix}
        \mathbf{\Phi}^* - \mathbf{\Psi} & \mathbf{v}_1 & \mathbf{0}_{m\times(r-1)}
    \end{bmatrix},
\end{equation}
where the last block row and block column are omitted if $r = 1$.

We first show that $\hat{\boldsymbol{\theta}} \in \Omega$. Since $\widehat{\mathbf{W}}_1$ is block diagonal, $(\mathbf{L}_T \otimes \mathbf{I}_m)^k = \mathbf{L}_T^k \otimes \mathbf{I}_m$ by the mixed-product property of the Kronecker product. Furthermore, $\mathbf{L}_T^k\mathbf{e}_1 = \mathbf{e}_{k+1}$ by \Cref{lowershiftmatrixremark1}, and we have
\begin{equation}
    \widehat{\mathbf{W}}_1^k\widehat{\mathbf{W}}_2
    =
    \begin{bmatrix}
        \mathbf{e}_{k+1} \otimes \mathbf{I}_m \\
        \lambda^k\mathbf{v}_1^\top \\
        \mathbf{0}_{(r-1)\times m}
    \end{bmatrix}
\end{equation}
for every $k \in [T-1]_0$. Hence,
\begin{equation}
    \widehat{\mathbf{W}}_3\widehat{\mathbf{W}}_1^k\widehat{\mathbf{W}}_2
    =
    (\mathbf{\Phi}^* - \mathbf{\Psi})(\mathbf{e}_{k+1} \otimes \mathbf{I}_m) + \lambda^k\mathbf{v}_1\mathbf{v}_1^\top
    =
    \mathbf{M}_k^* - \lambda^k\mathbf{v}_1\mathbf{v}_1^\top + \lambda^k\mathbf{v}_1\mathbf{v}_1^\top
    =
    \mathbf{M}_k^*
\end{equation}
for every $k \in [T-1]_0$. Therefore, $\mathbf{\Phi}_T(\hat{\boldsymbol{\theta}}) = \mathbf{\Phi}^*$, and hence $\hat{\boldsymbol{\theta}} \in \Omega$.

We now show that $\rho_{\text{co}}(\hat{\boldsymbol{\theta}}) = |\lambda|$. Since the $T$th power of $\mathbf{L}_T \otimes \mathbf{I}_m$ is $\mathbf{L}_T^{T} \otimes \mathbf{I}_m = \mathbf{0}$ by \Cref{remark:lower-shift-nilpotent}, the matrix $\mathbf{L}_T \otimes \mathbf{I}_m$ is nilpotent. Hence, by \Cref{lemma:nilpotent}, its only eigenvalue is $0$. Consequently, since $\widehat{\mathbf{W}}_1$ is a block-diagonal matrix, $\operatorname{spec}(\widehat{\mathbf{W}}_1) = \{0, \lambda\}$.

Let $\mathbf{q} = [\mathbf{q}_1^\top \ \ q_2 \ \ \mathbf{q}_3^\top]^\top \in \mathbb{R}^d$ be a vector, where $\mathbf{q}_1 \in \mathbb{R}^{mT}$, $q_2 \in \mathbb{R}$, and $\mathbf{q}_3 \in \mathbb{R}^{r-1}$, that satisfies $\mathbf{q}^\top[\lambda\mathbf{I}_d - \widehat{\mathbf{W}}_1 \ \ \widehat{\mathbf{W}}_2] = \mathbf{0}$. Note that $\lambda\mathbf{I}_{mT} - \mathbf{L}_T \otimes \mathbf{I}_m$ is nonsingular for any $\lambda \neq 0$. Then, $\mathbf{q}_1^\top(\lambda\mathbf{I}_{mT} - \mathbf{L}_T \otimes \mathbf{I}_m) = \mathbf{0}$ and $\lambda\mathbf{q}_3^\top = \mathbf{0}$ imply that $\mathbf{q}_1 = \mathbf{0}$ and $\mathbf{q}_3 = \mathbf{0}$. Consequently, $\mathbf{q}^\top\widehat{\mathbf{W}}_2 = q_2\mathbf{v}_1^\top = \mathbf{0}$, which implies $q_2 = 0$. Therefore, $\operatorname{rank}([\lambda\mathbf{I}_d - \widehat{\mathbf{W}}_1 \ \ \widehat{\mathbf{W}}_2]) = d$. Similarly, let $\mathbf{p} = [\mathbf{p}_1^\top \ \ p_2 \ \ \mathbf{p}_3^\top]^\top \in \mathbb{R}^d$ satisfy $(\lambda\mathbf{I}_d - \widehat{\mathbf{W}}_1)\mathbf{p} = \mathbf{0}$ and $\widehat{\mathbf{W}}_3\mathbf{p} = \mathbf{0}$. Then, $\mathbf{p}_1 = \mathbf{0}$ and $\mathbf{p}_3 = \mathbf{0}$ by the same argument, and $\widehat{\mathbf{W}}_3\mathbf{p} = p_2\mathbf{v}_1 = \mathbf{0}$ implies $p_2 = 0$. Therefore, $\operatorname{rank}([(\lambda\mathbf{I}_d - \widehat{\mathbf{W}}_1)^\top \ \ \widehat{\mathbf{W}}_3^\top]^\top) = d$. This means that $\lambda$ is a controllable and observable mode of $\widehat{\mathbf{W}}_1$ by the Hautus lemma \citep{hautus1969controllability}. Since the only other eigenvalue of $\widehat{\mathbf{W}}_1$ is $0$, we have $\rho_{\text{co}}(\hat{\boldsymbol{\theta}}) = |\lambda|$.

\paragraph{Case $d = mT$.} Since $\mathbf{\Phi}^* \neq \mathbf{0}$, there exists $(k, j) \in [T-1]_0 \times [m]$ such that $\mathbf{M}_k^*\mathbf{v}_j \neq \mathbf{0}$. We denote the standard basis vectors of $\mathbb{R}^d$ by $\{\mathbf{u}_1, \ldots, \mathbf{u}_d\}$ and define $\iota(k,j) := (j-1)T + k$ for every $k \in [T-1]_0$ and $j \in [m]$. Since $d = mT$, the map $(k,j) \mapsto \iota(k,j)$ is a bijection from $[T-1]_0 \times [m]$ onto $[d-1]_0$, and $\iota(T-1, m) = d-1$. 

Let us define a monic polynomial $g(s) := \sum_{i=0}^{d-1} t_i s^i$ with $t_{d-1} = 1$ as follows:
\begin{enumerate}[label=(\alph*)]
    \item \label{item:aaaa} If $\mathbf{M}_{T-1}^*\mathbf{v}_m \neq \mathbf{0}$, then $g(s) := s^{d-1}$, i.e., $t_i = 0$ for every $i \in [d-2]_0$.
    \item \label{item:bbbb} Otherwise, we choose $(k_0, j_0) \in [T-1]_0 \times [m]$ such that $\mathbf{M}_{k_0}^*\mathbf{v}_{j_0} \neq \mathbf{0}$, and let $i_0 := \iota(k_0, j_0)$. Note that since $\mathbf{M}_{T-1}^*\mathbf{v}_m = \mathbf{0}$, we have $(k_0, j_0) \neq (T-1, m)$, and hence $i_0 \leq d-2$. Then, $g(s) := s^{d-1} + \delta s^{i_0}$ with $\delta := (|\lambda|/2)^{d-1-i_0} > 0$, i.e., $t_{i_0} = \delta$, $t_{d-1} = 1$, and $t_i = 0$ otherwise.
\end{enumerate}
Now, let $\chi(s) := (s - \lambda)g(s) = s^d + \sum_{\ell=0}^{d-1} a_\ell s^\ell$ and $\mathbf{a} := [a_0 \ \cdots \ a_{d-1}]^\top \in \mathbb{R}^d$. We choose the particular parameterization $\Tilde{\boldsymbol{\theta}} = (\Tilde{\mathbf{W}}_1, \Tilde{\mathbf{W}}_2, \Tilde{\mathbf{W}}_3) \in \Theta$ as in \eqref{eq:thm-charpoly-realization}:
\begin{equation}
    \Tilde{\mathbf{W}}_1 = \mathbf{L}_d - \mathbf{a}\mathbf{u}_d^\top,
    \qquad
    \Tilde{\mathbf{W}}_2 = \mathbf{E}_1,
    \qquad
    \Tilde{\mathbf{W}}_3 = \sum_{k=0}^{T-1} \mathbf{M}_k^*\mathbf{E}_{k+1}^\top,
\end{equation}
where $\mathbf{E}_k = \mathbf{I}_m \otimes \mathbf{e}_k$ is defined in \eqref{eq:thm-charpoly-Ek} with $r = 0$. As we showed in the proof of \Cref{thm:arbitrary-charpoly}, $\mathbf{\Phi}_T(\Tilde{\boldsymbol{\theta}}) = \mathbf{\Phi}^*$ for every $\mathbf{a} \in \mathbb{R}^d$. Hence, $\Tilde{\boldsymbol{\theta}} \in \Omega$.

We now show that $\rho_{\text{co}}(\Tilde{\boldsymbol{\theta}}) = |\lambda|$. Since $\Tilde{\mathbf{W}}_1$ is the companion matrix of $\chi$, its characteristic polynomial is $\chi$ as in the proof of \Cref{thm:arbitrary-charpoly}. Hence, $\operatorname{spec}(\Tilde{\mathbf{W}}_1)$ consists of $\lambda$ and the roots of $g$. 

In \ref{item:aaaa}, every root of $g$ is $0$. In \ref{item:bbbb}, $g(s) = s^{i_0}(s^{d-1-i_0} + \delta)$; hence, every root of $g$ is either $0$ or a root of $s^{d-1-i_0} = -\delta$, whose modulus is $\delta^{1/(d-1-i_0)} = |\lambda|/2$. Therefore, every eigenvalue of $\Tilde{\mathbf{W}}_1$ other than $\lambda$ has modulus smaller than $|\lambda|$.

To show that $\lambda$ is controllable, let $\mathbf{q} \in \mathbb{R}^d$ satisfy $\mathbf{q}^\top[\lambda\mathbf{I}_d - \Tilde{\mathbf{W}}_1 \ \ \Tilde{\mathbf{W}}_2] = \mathbf{0}$. Then, $\mathbf{q}^\top\Tilde{\mathbf{W}}_1 = \lambda\mathbf{q}^\top$ and $\mathbf{q}^\top\Tilde{\mathbf{W}}_2 = \mathbf{0}$, which imply  that $\mathbf{q}^\top\Tilde{\mathbf{W}}_1^k\Tilde{\mathbf{W}}_2 = \lambda^k\mathbf{q}^\top\Tilde{\mathbf{W}}_2 = \mathbf{0}$ for every $k \geq 0$. Furthermore, note that $\Tilde{\mathbf{W}}_2\mathbf{v}_1 = \mathbf{E}_1\mathbf{v}_1 = \mathbf{u}_1$ by \eqref{shiftingpr}, and $\Tilde{\mathbf{W}}_1\mathbf{u}_i = \mathbf{u}_{i+1}$ for every $i \in [d-1]$ by \eqref{eq:thm-charpoly-shift}. Hence, $\Tilde{\mathbf{W}}_1^k\Tilde{\mathbf{W}}_2\mathbf{v}_1 = \mathbf{u}_{k+1}$, and thus $\mathbf{q}^\top\mathbf{u}_{k+1} = 0$ for every $k \in [d-1]_0$. This implies $\mathbf{q} = \mathbf{0}$. Therefore, $\operatorname{rank}([\lambda\mathbf{I}_d - \Tilde{\mathbf{W}}_1 \ \ \Tilde{\mathbf{W}}_2]) = d$. This means that $\lambda$ is a controllable mode of $\Tilde{\mathbf{W}}_1$ by the Hautus lemma \citep{hautus1969controllability}.

Now, let us define $\mathbf{w} := [t_0 \ \ t_1 \ \ \cdots \ \ t_{d-1}]^\top \in \mathbb{R}^d$, which is the vector of the coefficients of $g(s)$. Comparing them with the coefficients of $\chi(s) = (s - \lambda)g(s)$ gives $a_0 = -\lambda t_0$ and $a_i = t_{i-1} - \lambda t_i$ for every $i \in [d-1]$. Then, we have
\begin{equation}
    \Tilde{\mathbf{W}}_1\mathbf{w}
    = \mathbf{L}_d\mathbf{w} - \mathbf{a}
    = \begin{bmatrix} -a_0 \\ t_0 - a_1 \\ \vdots \\ t_{d-2} - a_{d-1} \end{bmatrix}
    = \begin{bmatrix} \lambda t_0 \\ \lambda t_1 \\ \vdots \\ \lambda t_{d-1} \end{bmatrix}
    = \lambda\mathbf{w},
\end{equation}
since $\mathbf{u}_d^\top\mathbf{w} = t_{d-1} = 1$. Furthermore, the term $\mathbf{a}\mathbf{u}_d^\top$ changes only the last column of $\lambda\mathbf{I}_d - \Tilde{\mathbf{W}}_1$. Hence, deleting the first row and the last column of $\lambda\mathbf{I}_d - \Tilde{\mathbf{W}}_1$ yields a $(d-1) \times (d-1)$ upper triangular matrix whose diagonal entries are $-1$ and whose superdiagonal entries are $\lambda$. Consequently, $\operatorname{rank}(\lambda\mathbf{I}_d - \Tilde{\mathbf{W}}_1) \geq d-1$, and since $\lambda\mathbf{I}_d - \Tilde{\mathbf{W}}_1$ is singular, $\ker(\lambda\mathbf{I}_d - \Tilde{\mathbf{W}}_1) = \operatorname{span}\{\mathbf{w}\}$. Now, let $\mathbf{p} \in \mathbb{R}^d$ satisfy $(\lambda\mathbf{I}_d - \Tilde{\mathbf{W}}_1)\mathbf{p} = \mathbf{0}$ and $\Tilde{\mathbf{W}}_3\mathbf{p} = \mathbf{0}$. Then, this holds if and only if $\mathbf{p} = \alpha\mathbf{w}$ for some $\alpha \in \mathbb{R}$, and $\alpha\Tilde{\mathbf{W}}_3\mathbf{w} = \mathbf{0}$. Since $\mathbf{v}_j^\top\mathbf{E}_{k+1}^\top\mathbf{w} = \mathbf{u}_{\iota(k,j)+1}^\top\mathbf{w} = t_{\iota(k,j)}$ by \eqref{shiftingpr}, we have
\begin{equation}
    \Tilde{\mathbf{W}}_3\mathbf{w}
    = \sum_{k=0}^{T-1}\mathbf{M}_k^*\mathbf{E}_{k+1}^\top\mathbf{w}
    = \sum_{k=0}^{T-1}\sum_{j=1}^{m} t_{\iota(k,j)}\mathbf{M}_k^*\mathbf{v}_j.
\end{equation}
In \ref{item:aaaa}, the only nonzero coefficient of $g$ is $t_{d-1} = 1$, and $d - 1 = \iota(T-1, m)$; hence, $\Tilde{\mathbf{W}}_3\mathbf{w} = \mathbf{M}_{T-1}^*\mathbf{v}_m \neq \mathbf{0}$. In \ref{item:bbbb}, the only nonzero coefficients of $g$ are $t_{i_0} = \delta$ and $t_{d-1} = 1$; hence, $\Tilde{\mathbf{W}}_3\mathbf{w} = \delta\mathbf{M}_{k_0}^*\mathbf{v}_{j_0} + \mathbf{M}_{T-1}^*\mathbf{v}_m = \delta\mathbf{M}_{k_0}^*\mathbf{v}_{j_0} \neq \mathbf{0}$. In both cases, $\alpha = 0$, and hence $\mathbf{p} = \mathbf{0}$. Therefore, $\operatorname{rank}([(\lambda\mathbf{I}_d - \Tilde{\mathbf{W}}_1)^\top \ \ \Tilde{\mathbf{W}}_3^\top]^\top) = d$. This means that $\lambda$ is a controllable and observable mode of $\Tilde{\mathbf{W}}_1$ by the Hautus lemma \citep{hautus1969controllability}. Since every other eigenvalue of $\Tilde{\mathbf{W}}_1$ has modulus smaller than $|\lambda|$, we have $\rho_{\text{co}}(\Tilde{\boldsymbol{\theta}}) = |\lambda|$.
\end{proof}

\section{Details of the Experiment in \Cref{fig:mamba}}
\label[appendix]{appendix:mamba-experiment}
In this section, we explain the experiment we ran on Mamba architectures in \Cref{fig:mamba}. We note that \Cref{fig:mamba-1} and \Cref{fig:mamba-2} are reproductions of \citet[Fig.~1a and~1b]{ruiz2025understanding}, respectively, without any training intervention that improves length generalization. On the other hand, to the best of our knowledge, the experiment for Mamba-3 in \Cref{fig:mamba-3} has not been reported previously.

\paragraph{Models.} We use the pretrained checkpoints released by the authors of the corresponding architectures\footnote{The checkpoints are available at \url{https://huggingface.co/state-spaces}.} without any fine-tuning. For Mamba-1 \citep{gu2023mamba}, we use the checkpoints with 130M and 370M parameters. For Mamba-2 \citep{dao2024transformers}, we use the checkpoints with 130M, 370M, 780M, and 1.3B parameters. These models were trained on 300B tokens of the Pile \citep{gao2020pile} with the GPT-NeoX-20B tokenizer. For Mamba-3 \citep{lahoti2026mamba}, we use the checkpoints of its multi-input multi-output (MIMO) variant with 187M, 444M, 894M, and 1.5B parameters. These models were trained on 100B tokens of FineWeb-Edu \citep{penedo2024fineweb} with the Llama-3.1 tokenizer. Note that all of these models were trained on sequences of $T = 2048$ tokens.

\paragraph{Test Sequences.} We test all models on the validation data of the Pile dataset, which consists of 214,670 documents. The training dataset Mamba-1 and Mamba-2 encompasses the training dataset of the Pile dataset. On the other hand, Mamba-3 is not trained on the Pile. Note that this mismatch makes the prediction harder at every position for Mamba-3. Furthermore, the perplexity values in \Cref{fig:mamba-3} are not comparable with those in \Cref{fig:mamba-1} and \Cref{fig:mamba-2}, since both the training data and the tokenizer differ.

Most documents in the Pile are far shorter than the context lengths we are interested in. The median document has only 438 tokens. Therefore, we construct long test sequences as follows: We tokenize each document with the tokenizer of the model under test, shuffle the documents with a fixed seed, and concatenate them into a single stream in which each document is followed by an end-of-sequence token. Then, we cut the stream into non-overlapping sequences of fixed length. In this way, we obtain 256 sequences of 131072 tokens, which is $64$ times the training horizon. Furthermore, to reduce the noise at the early positions, we obtain 4096 additional sequences of 8192 tokens from the part of the stream that follows the long sequences. Hence, the two sets of sequences do not share any tokens. 

\paragraph{Evaluation.} We process each test sequence in a single forward pass from its first token to its last token. As in training, the hidden state of each model is initialized to zero at the beginning of the sequence.

Let $m$ denote the size of the vocabulary and $\Delta_m := \{\mathbf{p} \in \mathbb{R}_+^m : \mathbf{p}^\top \mathbf{1} = 1\}$ denote the probability simplex. We denote by $\mathbf{y}_k^{(i)} \in \Delta_m$ the one-hot encoding of the token at position $k$ of the $i$th test sequence and by $\hat{\mathbf{y}}_k^{(i)} \in \Delta_m$ the distribution that the model predicts for this token after having observed $(\mathbf{y}_0^{(i)}, \ldots, \mathbf{y}_{k-1}^{(i)})$. Then, the cross-entropy loss incurred at position $k$ of the $i$th test sequence is $-\log(\mathbf{y}_k^{(i)\top} \hat{\mathbf{y}}_k^{(i)})$.

\begin{definition}[\citealp{ruiz2025understanding}]
Let $N_k$ denote the number of test sequences that are longer than $k$ tokens. For every $k \geq 1$, we define the \emph{position-wise perplexity} at position $k$ by
\begin{equation}
    \mathrm{PPL}_k := \exp\left(-\frac{1}{N_k}\sum_{i=1}^{N_k} \log\left(\mathbf{y}_k^{(i)\top} \hat{\mathbf{y}}_k^{(i)}\right)\right).
    \label{eq:positionwise-perplexity}
\end{equation}
\end{definition}

In other words, the position-wise perplexity is the exponential of the empirical per-step prediction risk under the cross-entropy loss, where the average is taken over the test sequences rather than over the positions. Ideally, the position-wise perplexity should not increase beyond the training horizon, since a longer context can only provide more information about the next token. We plot $\mathrm{PPL}_k$ for every $k \in [131071]$ in \Cref{fig:mamba}, where $N_k = 4352$ if $k < 8192$ and $N_k = 256$ otherwise.

\begin{remark}
Unlike \citet{ruiz2025understanding}, we do not smooth the position-wise perplexity over neighboring positions. Therefore, each point on a curve in \Cref{fig:mamba} is an average over only $N_k$ tokens. This is the reason for the excessive noise for $k \geq 8192$. Beyond this position, only the 256 long test sequences remain, whereas the earlier positions are averaged over 4352 test sequences. 
\end{remark}

For Mamba-1 and Mamba-2, the position-wise perplexity increases beyond the training horizon for every model size, and it grows by orders of magnitude for both Mamba-1 models and for the Mamba-2 model with 780M parameters, which agrees with the observations of \citet{ruiz2025understanding}. On the other hand, the degradation of Mamba-3 is considerably milder: Over the 256 long test sequences, the perplexity averaged over the positions $k \in [65536, 131071]$ is $8\%$ to $12\%$ higher than the perplexity averaged over the positions $k \in [1024, 2047]$, depending on the model size. We note that, with only 256 long test sequences, this increase is not large enough to be distinguished from the noise with confidence. 
\end{document}